\PassOptionsToPackage{table}{xcolor}
\documentclass[]{bytedance_seed}

\usepackage[toc,page,header]{appendix}
\usepackage{longtable}

\usepackage{minitoc}

\usepackage[utf8]{inputenc}  
\usepackage[T1]{fontenc}     
\usepackage{hyperref}        
\usepackage{url}             
\usepackage{booktabs}        
\usepackage{amsfonts}        
\usepackage{nicefrac}        
\usepackage{microtype}       
\usepackage{xcolor}          
\usepackage{amsmath}         
\usepackage{amsthm}
\usepackage{amssymb}         
\usepackage{colortbl}        
\usepackage{graphicx}        
\usepackage{cleveref}        
\usepackage{multirow}        
\usepackage{array}           
\usepackage{float}           
\usepackage{enumitem}        
\usepackage{subcaption}
\usepackage{wrapfig}
\usepackage{algorithm}
\usepackage[noend]{algpseudocode}

\usepackage[most]{tcolorbox} 
\usepackage{multicol}

\newcommand{\R}{\mathbb{R}}

\newtheorem{theorem}{Theorem}

\newtheorem{definition}{Definition}

\newtcbtheorem{theobox}{Example}%
{enhanced, breakable,
  colback=seedblue!3,                 
  colframe=seedblue!60!black,         
  boxrule=0pt,                    
  borderline={0.6pt}{0pt}{seedblue!60!black}, 
  arc=3pt,
  fonttitle=\bfseries,
  coltitle=white,
  attach boxed title to top left={yshift=-2mm,xshift=2mm},
  boxed title style={colback=seedblue!80!black, boxrule=0pt, arc=3pt},
}{th}
\usepackage{enumitem}

\definecolor{lightergray}{gray}{0.9} 
\newcommand{\ut}{LoopLM}

\title{Scaling Latent Reasoning via Looped Language Models}

\author[1,2,\dagger]{Rui-Jie Zhu*}
\author[1,3]{Zixuan Wang*}
\author[1]{Kai Hua*}
\author[4,5]{Tianyu Zhang*}
\author[1]{Ziniu Li*}
\author[1,6]{Haoran Que*}
\author[3]{Boyi Wei*}
\author[1,7]{Zixin Wen*}
\author[1]{Fan Yin*}
\author[11]{He Xing*}

\author[8]{Lu Li}
\author[1]{Jiajun Shi}
\author[1]{Kaijing Ma}
\author[1,7]{Shanda Li}
\author[2,9]{Taylor Kergan}
\author[2,9]{Andrew Smith}
\author[1,10]{Xingwei Qu}
\author[2]{Mude Hui}
\author[1]{Bohong Wu}
\author[1]{Qiyang Min}
\author[1]{Hongzhi Huang}
\author[1]{Xun Zhou}
\author[6]{Wei Ye}
\author[11]{Jiaheng Liu}
\author[11]{Jian Yang}
\author[11]{Yunfeng Shi}
\author[10]{Chenghua Lin}
\author[11]{Enduo Zhao}
\author[1]{Tianle Cai}

\author[1,\dagger]{\\Ge Zhang*}
\author[1,\dagger]{Wenhao Huang}
\author[4,5]{Yoshua Bengio}
\author[2,\dagger]{Jason Eshraghian}

\affiliation[1]{ByteDance Seed}
\affiliation[2]{UC Santa Cruz}
\affiliation[3]{Princeton University}
\affiliation[4]{Mila - Quebec AI Institute}
\affiliation[5]{University of Montreal}
\affiliation[6]{Peking University}
\affiliation[7]{Carnegie Mellon University}
\affiliation[8]{University of Pennsylvania}
\affiliation[9]{Conscium}
\affiliation[10]{University of Manchester}
\affiliation[11]{M-A-P}
\contribution[*]{Core Contributors}
\contribution[\dagger]{Corresponding authors}

\abstract{
Modern LLMs are trained to ``think'' primarily via explicit text generation, such as chain-of-thought (CoT), which defers reasoning to post-training and under-leverages pre-training data. We present and \textbf{open-source} Ouro, named after the recursive Ouroboros, a family of pre-trained Looped Language Models (LoopLM) that instead build reasoning into the pre-training phase through (i) iterative computation in latent space, (ii) an entropy-regularized objective for learned depth allocation, and (iii) scaling to \textbf{7.7T} tokens. Ouro 1.4B and 2.6B models enjoy superior performance that match the results of up to 12B SOTA LLMs across a wide range of benchmarks. Through controlled experiments, we show this advantage stems not from increased knowledge capacity, but from superior knowledge manipulation capabilities. We also show that LoopLM yields reasoning traces more aligned with final outputs than explicit CoT.  We hope our results show the potential of LoopLM as a novel scaling direction in the reasoning era.
}

\date{\today}
\correspondence{\email{ridger@ucsc.edu},
\email{zhangge.eli@bytedance.com}, \email{huang.wenhao@bytedance.com}, \email{jsn@ucsc.edu}}

\checkdata[Project Page \& Base / Reasoning Models]{\url{http://ouro-llm.github.io}}

\begin{document}

\maketitle

\begin{figure}[htbp]
    \centering
    \includegraphics[width=\linewidth]{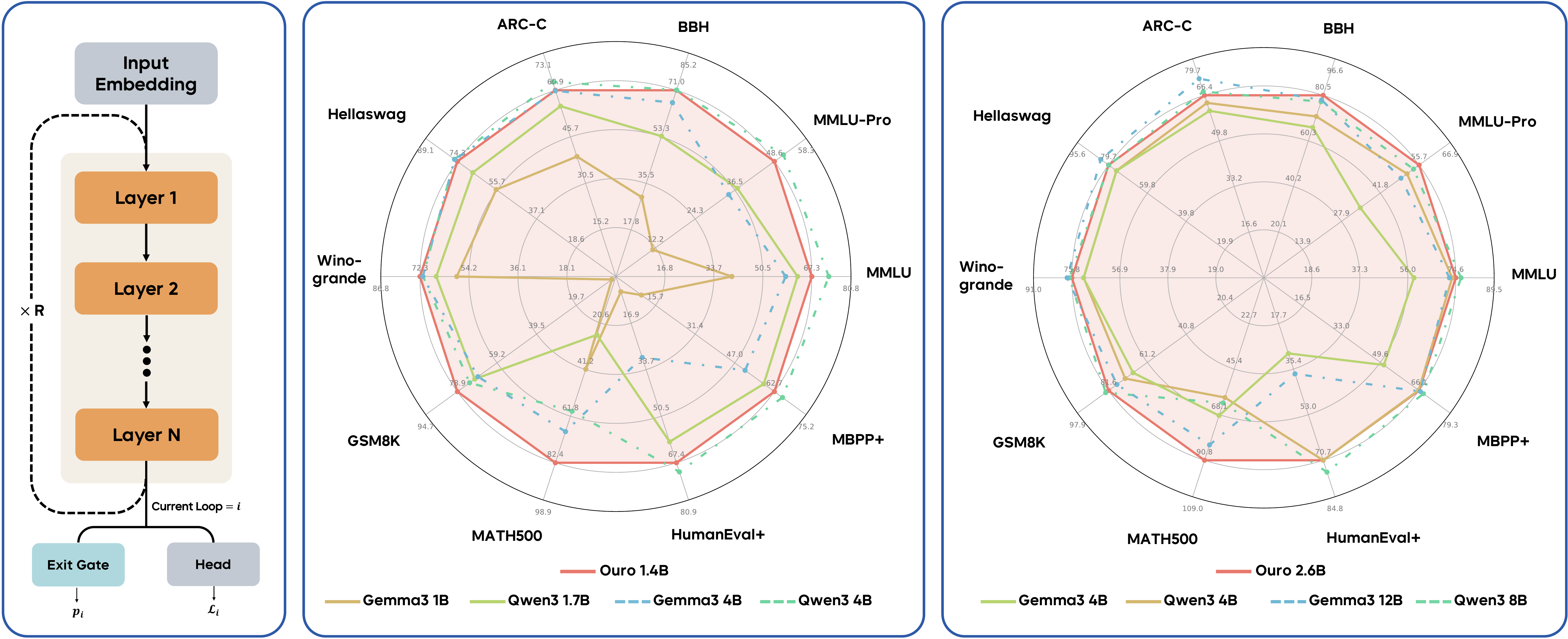}
    \caption{Ouro Looped Language Model performance. (Left) The parameter-shared looped architecture. (Middle \& Right) Radar plots comparing the Ouro 1.4B and 2.6B models, both with 4 recurrent steps (red), against individual transformer baselines. Our models demonstrate strong performance comparable to or exceeding much larger baselines.}
    \label{fig:radar_main}
\end{figure}

\section{Introduction}

The advancement of Large Language Models (LLMs) has historically relied on scaling up model size as the primary driver, accompanied by increases in data and compute~\cite{brown2020language,qwen2,qwen3,team2025gemma3}. However, deploying models with hundreds of billions of parameters requires extensive infrastructure, increasing latency and cost while limiting accessibility. These factors make \textit{parameter efficiency} critical: achieving better model capability within a fixed parameter budget. Such models not only mitigate overfitting on finite datasets with fewer trainable parameters, but also enable more practical deployment with lighter infrastructure. To achieve such parameter efficiency, two main avenues have been explored. The first expands the training corpus regardless of model size~\cite{dubey2024llama}, though data scarcity increasingly limits this path. The second leverages inference-time compute through Chain-of-Thought (CoT) reasoning~\cite{wei2022chain}, allowing models to spend more compute on complex problems via extended token generation.


\textbf{We explore a third pathway based on architectural innovation: achieving dynamic computation within a fixed parameter budget.} This is accomplished by recursively applying shared parameters, where a group of weight-tied layers are iteratively reused during the forward pass. We call this the Looped Language Model (\ut{}). The design yields several advantages. First, \ut{} enables adaptive computation via a learned early-exit mechanism: simple inputs can terminate after fewer recurrent steps, while complex ones allocate more iterations. This decouples the compute depth from parameter count. Second, unlike inference-time methods such as CoT, \ut{} scales by deepening its internal computational graph rather than extending the output sequence, avoiding context-length bloat. Finally, \ut{} can improve capacity per parameter and outperform standard transformers of larger sizes when trained on the same data. 

An extensive range of prior studies have explored \ut{} at modest scales~\cite{saunshi2025reasoning,gatmiry2024can,gatmiry2024role, huang2025transformers, merrill2025little, merrill2025exact,giannou2023looped,yang2023looped}, from the seminal Universal Transformer~\cite{dehghani2018universal} to recursive Transformers~\cite{bae2024relaxed} and latent reasoning approaches~\cite{geiping2025scaling,zeng2025pretraining,zhu2025survey}. Yet whether Looped Language Models translate into frontier-level gains at practically meaningful scales is unproven. To this end, we ask: 

\begin{center}
    \textbf{Does \ut{} exhibit more favorable scaling behavior (in capabilities, efficiency and safety),\\ compared to non-recursive transformer models?}
\end{center}

We show the answer is yes. We characterize \ut{}'s {} scaling trajectory and saturation behavior, demonstrating that \ut{} offers a more efficient path to higher performance. These claims are evaluated under multi-trillion-token training regimes typical of SoTA foundation models, extending well beyond prior work. Beyond the empirical gains, we analyze the mechanisms behind these improvements by asking the following questions:
\begin{itemize}
    \item[1.] Does the recursive reuse of weights yield the capability gains typically obtained by increasing depth with unshared weights?
    \item[2.] Are \ut{}'s gains monotonic in the number of loops? What are the factors that influence this?
\end{itemize}

\subsection*{Our Contribution}

\begin{figure}[t]
  \centering
  \includegraphics[width=0.9\linewidth]{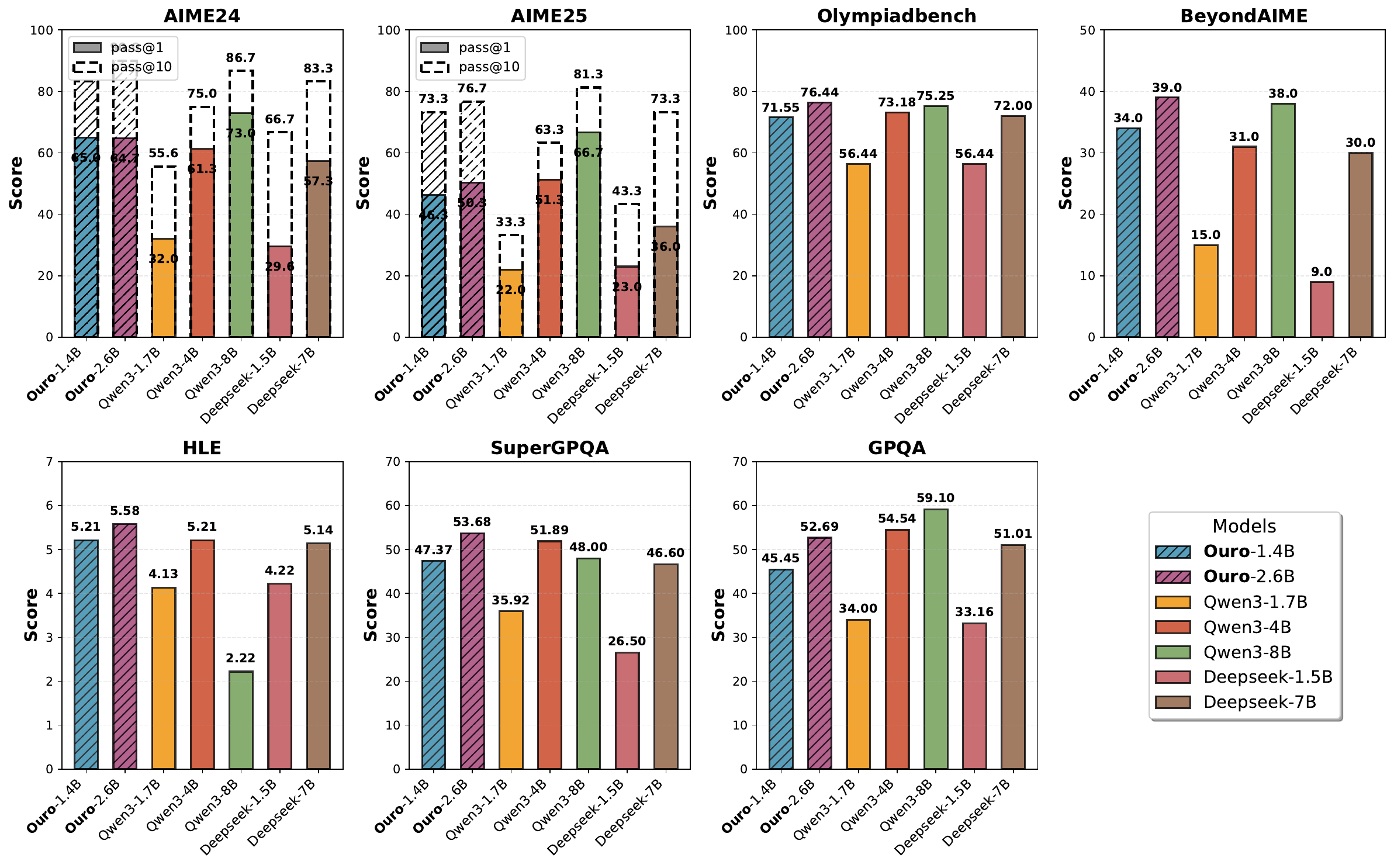}
  \caption{\textbf{Performance on advanced reasoning benchmarks.} Ouro-Thinking models compared with strong baselines such as Qwen3 and DeepSeek-Distill. \textbf{Ouro-1.4B-Thinking R4} is competitive with 4B models, and \textbf{Ouro-2.6B-Thinking R4} matches or exceeds 8B models across multiple math and science datasets.}
  \label{fig:reasoning_benchmark}
\end{figure}

We address the above questions with a multi-faceted study. We scale \ut{} pre-training to 7.7T tokens and thoroughly investigate its scaling behavior across multiple axes. To enable adaptive computation, we introduce training objectives that enable computationally efficient recurrence  while preserving peak performance. We also run controlled ablations to isolate the sources of \ut{}'s gains. Specifically, our contributions are:
\begin{itemize}
\item \textbf{Exceptional parameter efficiency at scale.} By pre-training on 7.7T tokens, we demonstrate that 1.4B and 2.6B parameter \ut{}s match 4B and 8B standard transformers on most benchmarks, yielding 2-3$\times$ parameter-efficiency gains that are critical for deployment in resource constrained environments (Figure~\ref{fig:radar_main} and Figure~\ref{fig:reasoning_benchmark}).
\item \textbf{Entropy-regularized adaptive computation.} Adaptive exits tend to collapse to shallow depths or overuse long loops. We avoid this with entropy-reguarization under a uniform prior over exit steps for unbiased depth exploration, followed by a focused training stage that tunes the compute-performance trade-off and allocates steps based on input difficulty.
\item \textbf{Mechanistic understanding of recurrence.} Using controlled experiments inspired by the physics-of-LMs framework, we find recurrence does not increase raw knowledge storage (approximately 2 bits per parameter for looped and non-looped models) but dramatically enhances knowledge manipulation capabilities on tasks requiring fact composition and multi-hop reasoning.
\item \textbf{Improved safety and faithfulness.} \ut{} reduces harmfulness on HEx\text{-}PHI~\citep{qifine}, with safety improving as recurrent steps increase (including extrapolated steps). Compared to CoT, our iterative latent updates produce reasoning traces that are better aligned with final outputs, indicating greater causal faithfulness rather than post-hoc rationalization.
\end{itemize}

Our study establishes loop depth as a third scaling axis beyond model size and data, and we publicly release the Ouro model family (1.4B and 2.6B parameters) to demonstrate the benefits of \ut{} at scale.

\section{Related Works}

The core ideas of this architecture have resurfaced in recent literature, with recurrent-depth structures used to improve the efficiency and reasoning capabilities of modern LLMs. For example, Geiping et al.~\cite{geiping2025scaling} adopts a ``recurrent depth'' to scale test-time computation in latent space. Similarly, Saunshi et al.~\cite{saunshi2025reasoning} demonstrates that ``looped transformers'' can match the performance of much deeper non-looped models on reasoning tasks, formally connecting looping to the generation of latent thoughts. The approach is refined by converting standard models into ``Relaxed Recursive Transformers'' with a common base block while injecting unique LoRA adapters across recursive steps~\cite{bae2024relaxed}. Similar concepts have emerged under different terms, such as ``pondering'' in continuous space~\cite{zeng2025pretraining} and ``inner thinking'' for adaptive computation~\cite{chen2025inner}. More advanced variants, such as Mixture-of-Recursions~\cite{bae2025mixture} combine recursive parameter efficiency with adaptive, token-level routing.

Across all these works, from the original Universal Transformer to its modern descendants, this emerging line of architectures can be understood in two complementary ways. From one perspective, it behaves like a deep Transformer where the weights of all layers are tied. From another, iteration functions as latent reasoning, where the hidden states form a latent chain of thought that progressively refines the representation to solve a task. Taken together, these results suggest that models can improve their ability to reason by reusing computation internally without having to increase parameter count, shifting scale to substance.

\paragraph{Perspective 1: Parameter Sharing for Model Efficiency.} This view treats \ut{} as parameter sharing: one or more Transformer blocks, or even submodules (e.g., attention, FFN), are reused across the depth of the model, reducing parameters without changing the computation. The most prominent example in the modern transformer era is ALBERT~\cite{lan2019albert}, which combines parameter re-use with embedding factorization to drastically reduce the total parameter count. Prior to the widespread adoption of LLMs, parameter sharing was explored extensively in machine translation~\cite{dabre2019recurrent}; Takase et al.~\cite{takase2021lessons} systematically studied sharing strategies to balance compression and accuracy. Interest in parameter reuse dropped as models grew larger, but it has resurged to shrink the memory footprint of LLMs. For example, Megrez2~\cite{li2025megrez2} reuses experts across layers in a standard Mixture-of-Experts (MoE) model, and shows a viable path forward for edge LLM deployment with limited memory.

\paragraph{Perspective 2: Latent Reasoning and Iterative Refinement.}
Here, the \ut{}'s iteration is viewed as latent reasoning where each step is a non-verbal ``thought'' that refines the model's internal representation. Empirically, increasing the number of recurrent steps improves performance on complex reaasoning tasks~\cite{geiping2025scaling, saunshi2025reasoning}. Some models make this process explicit by feeding hidden states back into the input. Coconut inserts a ``continuous thought'' token, which is derived from the previous steps's last-layer hidden state, so the model can ``ponder'' in a continuous latent space~\cite{hao2024training}. CoTFormer interleaves activations back into the input before reapplying this augmented sequence to the shared layers~\cite{mohtashami2023cotformer}. These explicit feedback loops contrast with implicit \ut{} variants, where the entire thought process is contained within the evolution of hidden states from previous recurrent step to current. Thus, both Perspective 1 (model compression) and Perspective 2 (latent reasoning) leverage shared-parameter iteration to improve parameter efficiency, and are being explored for enhance reasoning and efficient sequence-length expansion (e.g., PHD-Transformer~\cite{wu2025efficient}).

\section{Learning Adaptive Latent Reasoning with \ut{}}

\begin{figure}[t]
    \centering
    \includegraphics[width=\linewidth]{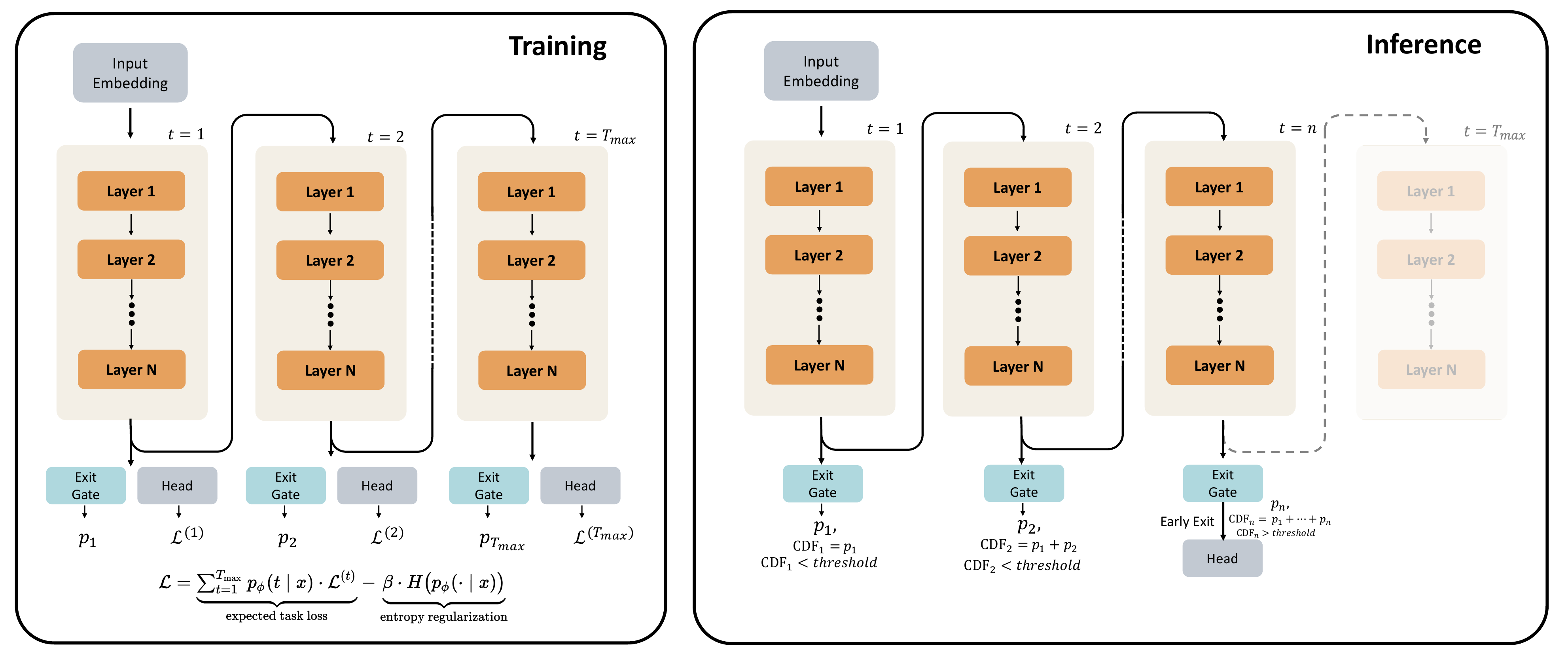}
    \caption{\textbf{Overview of Looped Language Model (\ut{}) architecture.} 
    \textbf{Left (Training):} During training, the model applies a stack of $N$ layers repeatedly for $T_{max}$ recurrent steps. At each recurrent step $\ell$, an exit gate predicts the probability $p_\ell$ of exiting, and a language modeling head $\mathcal{L}_\ell$ computes the lanugage modeling loss. 
    \textbf{Right (Inference):} At inference time, the model can exit early based on the accumulated exit probability.}
    \label{fig:rlm_architecture}
\end{figure}

In this section, we formally define the \ut{} architecture on causal Transformers and present our training scheme for adaptive latent reasoning. \Cref{fig:rlm_architecture} illustrates the architecture 
in training and inference. Our goal is to let the model choose the number of recurrent steps per token and per example, spending less compute on easy inputs and more on hard inputs, without sacrificing accuracy when many steps are available.

\subsection{\ut{} Architecture}

Let $\mathrm{emb}(\cdot):\mathbb{R}^{|V|}\to\mathbb{R}^d$ be the token embedding; $\mathcal{T}_{\theta}(\cdot): \mathbb{R}^{M\times d}\to\mathbb{R}^{M\times d}$ a causal transformer layer parameterized by $\theta$ with hidden size $d$ and input length $M$, and $\mathrm{lmhead}(\cdot):\mathbb{R}^{d} \to \mathbb{R}^{|V|}$ be the unembedding layer with vocabulary size $V$. A non-looped LM stacks $L$ layers, where $\circ$ denotes function composition:
$$
F(\cdot) := \mathrm{lmhead}\circ\mathcal{M}^L\circ\mathrm{emb(\cdot)},\quad \qquad \mathcal{M}^L(\cdot) := \mathcal{T}_{\theta_L}\circ \cdots \circ \mathcal{T}_{\theta_1}(\cdot)
$$

Let $t \in \{1, \dots, T_{\max}\}$ be the number of loop steps (the number of \textbf{recurrent steps} or \textbf{recurrent depth}). The looped model $F^{(t)}$ reuses the same depth-$L$ layer stack $t$ times:
\begin{equation}\label{eqdef:looped-lm}
F^{(t)}(\cdot) = \mathrm{lmhead}\circ\underbrace{\mathcal{M}^L\circ \mathcal{M}^L\circ\cdots\circ \mathcal{M}^L}_{t \text{ iterations}}\circ\ \mathrm{emb}(\cdot).
\end{equation}

Thus, $t=1$ yields the non-looped model $F^{(1)} \equiv F$.
As shown in Figure \ref{fig:rlm_architecture} (Left), at each recurrent step $t$, the model produces a language modeling head output. For a sequence $x=x_{1:M}$, define the token-level negative log-likelihood $\ell_i^{(t)}(x)$ and the aggregate single-step loss $\mathcal{L}^{(t)}$:
\begin{equation}\label{eqdef:loop-loss}
\begin{aligned}
\ell_i^{(t)}(x)&=-\log p^{(t)}_\theta\!\big(x_{i+1}\mid x_{1:i}\big),\\
\mathcal{L}^{(t)}&=\mathbb{E}_{x_{1:M}}\Bigg[\sum_{i=1}^{M-1}\ell_i^{(t)}(x)\Bigg].
\end{aligned}
\end{equation}

Here $p^{(t)}_\theta(\cdot\mid x_{1:i})=\mathrm{softmax}\!\big(\mathrm{lmhead}(h^{(t)}_i)\big)$, $x_{1:i}$ denotes the prefix containing tokens $1$ through $i$, and $h^{(t)}_i$ is the hidden state after $t$ loops at position $i$. The adaptive training objective below weights the token-level losses before aggregating over positions and sequences.

Prior literature \cite{merrill2025little,saunshi2025reasoning} has shown that scaling up $t$ is beneficial for reasoning tasks. However, this increases computation, and not all tokens require many steps \cite{ye2024physics,wang2025beyond}. Thus, it is crucial to spend the computation budget on the right tokens. This is achieved by the \textbf{gating mechanism} described in the next section.

\subsection{Adaptive Computation via Gating Mechanism} 
To enable adaptive computation, we add an exit gate that runs in parallel with the LM head at each step $t \leq T_{\max}$ (\Cref{fig:rlm_architecture}). At each loop $t$, the gate outputs an instantaneous (per-step) exit probability

$$
\lambda_t(x) = \sigma\left(\mathrm{Linear}_{\phi}\left(h^{(t)}\right)\right) \in(0,1)
$$
where $h^{(t)}$ is the final-layer hidden state at step $t$ and $\phi$ are the gate parameters. 
We define

$$
S_t(x)=\prod_{j=1}^t\bigl(1-\lambda_j(x)\bigr),\qquad S_0(x)\equiv 1,$$

as the survival, or the probability of not exiting in the first $t$ steps. The unnormalized probability of exiting first at step $t$ is then

$$
\tilde{p}_t(x)=\lambda_t(x)\,S_{t-1}(x),\qquad t=1,\dots,T_{\max}-1.
$$

To obtain a valid discrete distribution over exit steps, we assign the remaining mass to the final step:

\begin{equation}\label{eq:distribution}
p_\phi(t\mid x) =
\begin{cases}
    \tilde p_t(x), & t=1,\dots,T_{\max}-1,\\[3pt]
    S_{T_{\max}-1}(x), & t=T_{\max},
\end{cases}
\qquad\text{so that}\quad \sum_{t=1}^{T_{\max}}p_\phi(t\mid x)=1.
\end{equation}




\paragraph{Inference with Early Exit.}
As illustrated in Figure \ref{fig:rlm_architecture} (Right), we infer an exit step from the learned exit distribution $\{p_\phi(t\mid x)\}_{t=1}^{T_{\max}}$, enabling efficient inference. The cumulative exit probability up to step $n$ is:

$$
\mathrm{CDF}(n\mid x) = \sum_{t=1}^{n} p_\phi(t\mid x)
= 1 - \prod_{j=1}^{n}\bigl(1-\lambda_j(x)\bigr), \quad
n<T_{\max},\qquad \mathrm{CDF}(T_{\max}\mid x)=1.
$$

Given a threshold $q \in [0,1]$, we terminate at the first step where the cumulative probability crosses $q$:
$$
t_{\mathrm{exit}}(x) = \min\{\,m \in \{1,\dots,T_{\max}\}\,\;:\; \mathrm{CDF}(m\mid x) \ge q\, \}.
$$

The threshold $q$ controls the compute–accuracy tradeoff: smaller $q$ favors earlier exits (less compute), while larger $q$ allows deeper computation. In practice, $q$ may be chosen globally, calibrated per task, or scheduled with a floor/ceiling on steps. This deterministic, quantile-based policy follows the Q-exit criterion introduced in PALBERT~\cite{balagansky2022palbert}, selecting the first layer or recurrent step whose cumulative halting probability exceeds a threshold. It avoids sampling while remaining consistent with the learned distribution. 

The gating parameters $\phi$ (and thus $p_\phi$ via $\{\lambda_t\}$) are learned in  two stages:
\begin{itemize}
    \item \textbf{Stage I}: During pre-training, the gates are learned jointly with the LM by optimizing an entropy-regularized objective (\Cref{sec:entropy}).
    \item \textbf{Stage II}: We freeze the LM and fine-tune $\phi$ to sharpen $p_\phi$ (i.e., adjust depth allocation) without changing token-level predictions.
\end{itemize}

The complete training objective is described in the next section.

\subsection{Stage I: Learning an Entropy-Regularized Objective}\label{sec:entropy}

Under naive gradient descent on the next-token prediction loss, deeper loops typically reduce the single-step loss $\mathcal{L}^{(t)}$ from \Cref{eqdef:loop-loss} up to some depth; beyond that, gains diminish and the gradients shift probability mass toward later steps. As $p_{\phi}$ concentrates on late steps, those steps receive more training signal and their losses drop further, which in turn pulls even more mass to the end. This self-reinforcement collapses $p_{\phi}$ onto $t=T_{\rm max}$. An entropy term penalizes collapse to the deepest step, maintaining enough spread in $p_\phi$ to reflect input difficulty. 

Given the token-level losses from \Cref{eqdef:loop-loss} and the exit-step distribution from \Cref{eq:distribution}, our training objective combines next-token prediction with entropy regularization. The gate at position $i$ conditions on the same prefix $x_{1:i}$ used to predict $x_{i+1}$:

\begin{equation}\label{eqdef:entropy-reg-obj}
\mathcal{L}=\mathbb{E}_{x_{1:M}}\Bigg[\sum_{i=1}^{M-1}\Bigg(
\underbrace{\sum_{t=1}^{T_{\max}}p_\phi(t\mid x_{1:i})\,\ell_i^{(t)}(x)}_{\text{expected task loss}}
-\underbrace{\beta H\!\left(p_\phi(\cdot\mid x_{1:i})\right)}_{\text{entropy regularization}}
\Bigg)\Bigg].
\end{equation}
Here $H(p)=-\sum_t p_t\log p_t$. We use a sum over prediction positions; any token averaging or loss masking is applied to both the task and entropy terms. Each gate weights the losses for its own prediction, so their product remains inside the data expectation. The coefficient $\beta$ controls the exploration–exploitation trade-off: larger $\beta$ encourages higher-entropy (more exploratory) $p_\phi$, while smaller $\beta$ lets $p_\phi(t\mid x_{1:i})$ place most of its mass on a specific step when the model is confident about the optimal depth.

\paragraph{Alternative perspective: variational inference with a uniform prior.}
The objective in \Cref{eqdef:entropy-reg-obj} has a variational interpretation where each prediction has a latent exit step $z_i\in\{1,\dots,T_{\max}\}$, with variational distribution $p_\phi(z_i{=}t\mid x_{1:i})$ and prior $\pi(t)$. The corresponding $\beta$-weighted variational objective is:
\[
\mathcal{L}_{\beta\text{-VI}}=\mathbb{E}_{x_{1:M}}\Bigg[\sum_{i=1}^{M-1}\Bigg(
\sum_{t=1}^{T_{\max}}p_\phi(t\mid x_{1:i})\,\ell_i^{(t)}(x)
+\beta\,\mathrm{KL}\!\big(p_\phi(\cdot\mid x_{1:i})\,\|\,\pi\big)
\Bigg)\Bigg].
\]
With a uniform prior $\pi_t = 1/T_{\max}$, the KL becomes
\[
\mathrm{KL}\!\big(p_\phi(\cdot\mid x_{1:i})\,\|\,\pi\big)
=-H\!\left(p_\phi(\cdot\mid x_{1:i})\right)+\log T_{\max},
\]
so this objective differs from \Cref{eqdef:entropy-reg-obj} by the parameter-independent constant $\beta\log T_{\max}$ per prediction position, under the same aggregation. For $\beta=1$, it is the usual negative ELBO aggregated over predictions; other values of $\beta$ change the relative weight of the KL regularizer. The task term averages per-step losses under the exit distribution; it is not the negative log-likelihood obtained by marginalizing the predictive probabilities. This connects the objective to variational adaptive-computation methods such as PonderNet~\cite{banino2021pondernet}.

\paragraph{Why a uniform prior?}
Different priors encode different depth preferences. A geometric prior, as in Ref.~\cite{banino2021pondernet}, or Poisson-lognormal priors softly favor earlier halting~\cite{geiping2025scaling}, while a uniform prior is depth-unbiased. We adopt the uniform prior to decouple exit decisions driven by input difficulty from any global compute preference; the entropy term then prevents collapse to always using $T_{\rm max}$. Empirical comparisons with geometric priors are provided in Appendix \Cref{app:prior_empirical}.


\subsection{Stage II: Focused Adaptive Gate Training}\label{ut:adaptive_exit}


In this stage, we freeze the LM parameters and train only the exit gate to make termination decisions based on realized performance gains. We use a greedy signal that balances marginal improvement from an extra loop against additional compute. 

To ensure the gate does not alter LM representations, we compute a detached per-step loss $\mathcal{L}_{i,\mathrm{stop}}^{(t)}$ at each token $i$ and define the loss improvement from step $t\!-\!1$ to $t$ as

\begin{equation}\label{eqdef:improvement-per-step}
    I^{(t)}_i = \max\!\big(0,\ \mathcal{L}_{i,\mathrm{stop}}^{(t-1)} - \mathcal{L}_{i,\mathrm{stop}}^{(t)})
\end{equation}

where larger $I_i^{(t)}$ indicates ongoing improvement; a smaller value indicates that gains have stalled and \ut{} should opt for an early exit. We implement this by computing the \textbf{ideal continuation probability}, a training label that indicates whether to \textit{continue} (near 1) or \textit{exit} (near 0):

$$w^{(t)}_i = \sigma(k \cdot (I^{(t)}_i - \gamma))$$ 

with slope $k=50.0$ and threshold $\gamma=0.005$, so that $w^{(t)}_i\!\approx\!1$ recommends continuing and $w^{(t)}_i\!\approx\!0$ recommends exiting the loop.
The adaptive exit loss at step $t$ takes the binary cross-entropy between the gate's predicted continuation probability $1-\lambda^{(t)}_{i}$ and the ideal label $w^{(t)}_i$, averaged over the sequence length $M$:
\begin{equation}\label{eqdef:adaptive-loss}
\mathcal{L}^{(t)}_{\text{adaptive}}
= - \frac{1}{M}\sum_{i=1}^{M}\!\Big[
w^{(t)}_{i}\,\log\bigl(\underbrace{1-\lambda^{(t)}_{i}}_{\substack{\text{predicted}\\\text{continuation}}}\bigr)
+\bigl(1-w^{(t)}_{i}\bigr)\,\log\bigl(\underbrace{\lambda^{(t)}_{i}}_{\substack{\text{predicted}\\\text{exit}}}\bigr)
\Big].
\end{equation}

The total adaptive loss averages across recurrent steps: 
\begin{equation*}
    \mathcal{L}_{\text{adaptive}} = \frac{1}{T_{\max}} \sum_{t=2}^{T_{\max}} \mathcal{L}_{\text{adaptive}}^{(t)}
\end{equation*}

\paragraph{Significance of our adaptive loss.}
The adaptive loss in \Cref{eqdef:adaptive-loss} trains the gate at step $t$ to match its predictions to the ideal behavior derived from actual performance improvements:
\begin{itemize}
    \item \textbf{Predicted probabilities:} The gate generates $\lambda^{(t)}_{i}$ (exit probability) and $1-\lambda^{(t)}_{i}$ (continuation probability)
    \item \textbf{Target labels:} The ideal behavior is encoded as $w^{(t)}_{i}$ (target continuation probability) and $1-w^{(t)}_{i}$ (target exit probability)
\end{itemize}

This formulation penalizes two failure modes simultaneously:
\begin{itemize}
    \item \textbf{Underthinking:} the gate exits when it should continue (large label $w^{(t)}_{i}$, but large predicted exit $\lambda^{(t)}_{i}$)
    \item \textbf{Overthinking:} the gate continues when it should exit (small label $w^{(t)}_{i}$, but small predicted exit $1-\lambda^{(t)}_{i}$)
\end{itemize}

Optimizing \Cref{eqdef:adaptive-loss} trains the gate to choose a greedy exit step that trades additional compute for measured improvement. For empirical evaluations, see \Cref{sec:early_exit_strategy}.

\section{Training Looped Language Models}
\label{sec:training_looplm}
\begin{figure}[t]
    \centering
    \includegraphics[width=\linewidth]{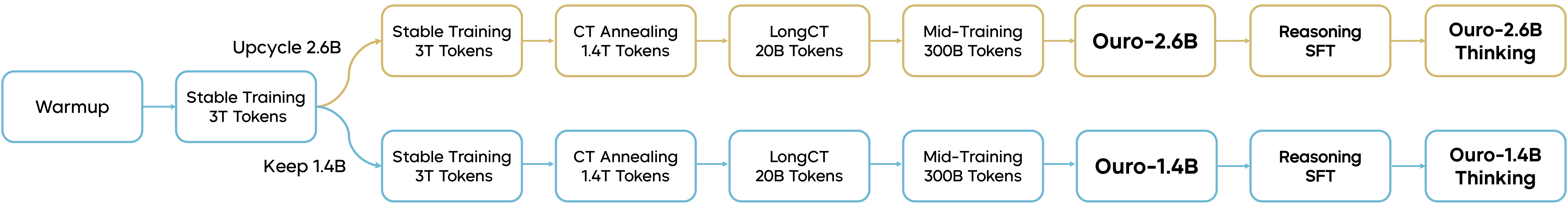}
    \caption{End-to-end Ouro training pipeline: shared warmup → Stable Training → forks into a 1.4B retained path and a 2.6B upcycled path → four shared stages → Reasoning SFT to produce Ouro-Thinking.}
    \label{fig:training_process}
\end{figure}

Our end-to-end training pipeline for the Ouro model family is shown in Figure~\ref{fig:training_process}. In total 7.7T tokens are used to train the base models \textbf{Ouro-1.4B} and \textbf{Ouro-2.6B}. A final Reasoning SFT (Supervised Fine-Tuning) yields the \textbf{Ouro-1.4B-Thinking} and \textbf{Ouro-2.6B-Thinking} variants. This section details the architecture, data composition, and specific configurations used in each of these training stages. A high-level overview of the training recipe of the first four stages is given in \Cref{tab:training_recipe}.


\begin{table}[h!]
\small
\centering
\caption{\textbf{Training recipe for Ouro 1.4B and 2.6B.}}
\label{tab:training_recipe}
\begin{tabular}{l|ccccc}
\toprule
 & \textbf{Stage 1a} & \textbf{Stage 1b} & \textbf{Stage 2} & \textbf{Stage 3} & \textbf{Stage 4} \\
 & \textbf{Pre-train I} & \textbf{Pre-train II} & \textbf{CT Annealing} & \textbf{LongCT} & \textbf{Mid-training} \\
\hline
\textbf{Hyperparameters} \\
Learning rate (Final) & $3.0\times10^{-4}$ & $3.0\times10^{-4}$ & $3.0\times10^{-5}$ & $3.0\times10^{-5}$ & $1.0\times10^{-5}$ \\
LR scheduler & Constant & Constant & Cosine Decay & Constant & Cosine Decay \\
Weight decay & \multicolumn{5}{c}{0.1} \\
Gradient norm clip & \multicolumn{5}{c}{1.0} \\
Optimizer & \multicolumn{5}{c}{AdamW ($\beta_1=0.9$, $\beta_2=0.95$)} \\
Batch size (tokens) &4M$\rightarrow$ 8M & \multicolumn{4}{c}{8M} \\
Sequence length & 4K & 4K & 16K & 64K & 32K \\
Training tokens & 3T & 3T & 1.4T & 20B & 300B \\
Recurrent steps & 8 & \multicolumn{4}{c}{4} \\
$\beta$ for KL divergence & 0.1 & \multicolumn{4}{c}{0.05} \\
RoPE base & 10K & 10K & 40K & 1M & 1M \\
\midrule
\textbf{Data Focus} \\
Web data & High & High & Medium & Low & Low \\
Math \& Code & Low & Low & High & Low & High \\
Long-context & None & None & Low & High & Medium \\
SFT-quality & None & None & Low & Low & High \\
\bottomrule
\end{tabular}
\end{table}

\subsection{Transformer Architecture} 

The Ouro models use a standard decoder-only Transformer~\cite{vaswani2017attention}, prioritizing a clean implementation of the looped computation mechanism without extraneous modifications. The core architecture consists of a stack of Transformer blocks applied recurrently. Each block uses Multi-Head Attention (MHA) with Rotary Position Embeddings (RoPE)~\cite{su2023roformerenhancedtransformerrotary}. The feed-forward network (FFN) in each block utilizes a SwiGLU activation~\cite{shazeer2020glu}. To enhance training stability, which is especially critical for deep recurrent computation, we employ a \textbf{sandwich normalization} structure, placing an \texttt{RMSNorm} layer before both the attention and FFN sub-layers~\cite{geiping2025scaling}. 
Both models use a 49,152-token vocabulary from the SmolLM2 model~\cite{allal2025smollm2}. This tokenizer is optimized for code and Latin-alphabet language. Architectural details are summarized in \Cref{tab:model_architecture}.

\begin{table}[htbp]
\centering
\caption{Ouro model architecture configurations. Both models share the same vocabulary and core component types, differing in parameter count and layer depth.}
\label{tab:model_architecture}
\small 
\begin{tabular}{@{}lccccccc@{}}
\toprule
\textbf{Model} & \textbf{Parameters} & \textbf{Layers} & \textbf{Hidden Size ($d_{\text{model}}$)} & \textbf{Attention} & \textbf{FFN} & \textbf{Pos. Embed.} & \textbf{Vocab Size} \\
\midrule
Ouro 1.4B & 1.4B & 24 & 2048 & MHA & SwiGLU & RoPE & 49,152 \\
Ouro 2.6B & 2.6B & 48 & 2048 & MHA & SwiGLU & RoPE & 49,152 \\
\bottomrule
\end{tabular}
\end{table}

\subsection{Data}
Data sets the capability bounds of foundation models. Our corpus spans web text, mathematics, code, and long-context documents across multiple stages, building core language understanding while strengthening reasoning, coding, and long-context skills. Beyond standard web crawls, we include targeted datasets for mathematical reasoning and code generation to improve complex problem solving. \Cref{tab:training_data_details} summarizes composition and scale at each training stage. 

\begin{table}[ht]
\label{tab:data_stat_of_training_corpus}
\small
\centering
\caption{\textbf{Statistics of the training corpus.} Since data are randomly sampled during pre-training, the dataset size does not directly correspond to the total number of seen tokens.}
\label{tab:training_data_details}
\begin{tabular}{l|ccc}
\toprule
\textbf{Data Source} & \textbf{Stage} & \textbf{\# Tokens (B)} & \textbf{\# Used Tokens (B)} \\
\hline
Nemotron-CC (Web Data) & Stage 1 & 6386 & 4404 \\
MAP-CC (Web Data) & Stage 1 & 800 & 780\\
Ultra-FineWeb-zh (Web Data) & Stage 1 & 120 & 120 \\
OpenCoder-pretrain & Stage 1 & 450 & 450 \\
MegaMath-web  & Stage 1 & 247 & 246 \\
MegaMath-high-quailty & Stage 2 & 64 & 64\\
Nemotron-CC-Math-v1 & Stage 2 & 210 & 210 \\
Nemotron-Code & Stage 2 & 53 & 53 \\
Nemotron-SFT-Code & Stage 2 & 48 & 48 \\
Nemotron-SFT-General & Stage 2 & 87 & 87 \\
OpenCoder-Annealing & Stage 2 & 7 & 7  \\
ProLong-64K & Stage 3 & 20 & 20 \\
Mid-training SFT Mix & Stage 4 & 182 & 90 \\
\bottomrule
\end{tabular}
\end{table}

\begin{table}[ht]
\small
\centering
\caption{\textbf{Data composition for Stage 1 (Stable Training I \& II).} Total dataset size: 6T tokens.}
\label{tab:stage1_composition}
\begin{tabular}{l|ccccc}
\toprule
\textbf{Data Source} & Nemotron-CC&MAP-CC&Ultra-FineWeb-zh&OpenCoder-pretrain&MegaMath-web \\
\midrule
\textbf{Proportion (\%)} & 73.4 & 13.0 & 2.0 & 7.5 & 4.1 \\
\bottomrule
\end{tabular}
\end{table}

To ensure reproducibility, our training corpus is composed entirely of open-source datasets, with data statistics summarized in \autoref{tab:stage1_composition}. We partition the data into four  stages, each with construction strategies aligned to the Warmup-Stable-Decay (WSD) learning rate scheduler~\cite{wen2024understanding} commonly used in modern pre-training.

\paragraph{Stage 1: Pre-training}
This stage supports the warmup and stable phases of training. The corpus is primarily composed of Web CommonCrawl (CC) data. Because we sought to train the model on >2T tokens, many popular open corpora are too small (e.g., Fineweb-Edu at 1.3T tokens~\cite{penedo2024fineweb}, DCLM at 2.6T tokens~\cite{li2024dclm}). We therefore use Nemotron-CC~\cite{su2024nemotron} (6.3T tokens) as the main dataset for the stable-phase. To provide the model with basic Chinese proficiency, we include Ultra-FineWeb-zh~\cite{wang2025ultra} and MAP-CC~\cite{du2024chinese}. However, without Chinese vocabulary in the tokenizer, characters would be fragmented into multiple byte-level sub-tokens, so we removed Chinese from Stage~2 onwards. To enhance coding and mathematical abilities, we incorporate OpenCoder~\cite{Huang2024OpenCoderTO} and MegaMath~\cite{zhou2025megamath}. See \Cref{tab:stage1_composition} for dataset proportions in further detail.

\paragraph{Stage 2: Continual Training (CT) Annealing}
The CT annealing stage incorporates higher-quality data to enhance the model under the annealing learning rate. Token sequence length is extended to 16K tokens, exceeding the length of most samples to minimize truncation. We construct the corpus from the high-quality subset of Nemotron-CC and augment with HQ MegaMath, Nemotron-CC-Math-v1~\cite{karimi2025nemotroncc,nvidia2025nvidianemotronnano2}, OpenCoder-Annealing~\cite{Huang2024OpenCoderTO}, Nemotron-pre-training-Code-v1~\cite{nvidia2025nvidianemotronnano2}, and Nemotron-pre-training-SFT-v1~\cite{nvidia2025nvidianemotronnano2}. Data composition is provided in\Cref{tab:stage2_composition}.

\begin{table}[ht]
\small
\centering
\caption{\textbf{Data composition for Stage 2 (CT Annealing).} Total dataset size: 1.4T tokens.}
\label{tab:stage2_composition}
\begin{tabular}{l|c}
\toprule
\textbf{Data Source} & \textbf{Proportion (\%)} \\
\hline
Nemotron-CC-high-quailty & 66.5 \\
Nemotron-CC-Math-v1 & 15.0 \\
MegaMath-high-quailty & 4.6 \\
OpenCoder-LLM/opc-annealing-corpus & 0.5 \\
Nemotron-pre-training-Code-v1/Synthetic-Code & 3.8 \\
Nemotron-pre-training-SFT-v1/Nemotron-SFT-Code & 3.4 \\
Nemotron-pre-training-SFT-v1/Nemotron-SFT-General & 6.2 \\

\bottomrule
\end{tabular}
\end{table}

\paragraph{Stage 3: Long Context Training (LongCT)}
The LongCT stage extends the long-context capabilities of the model. We adopt the 64K-length subset of ProLong~\cite{gao2024train}, consisting of 20B tokens, to train the model on longer sequences and improve its ability to handle long contexts.

\paragraph{Stage 4: Mid-training}
This stage uses a diverse set of extremely high-quality data, consisting of both $\langle$Question, Answer$\rangle$ and $\langle$Question, CoT, Answer$\rangle$ samples, to further develop advanced abilities. We integrate 20+ open-source SFT datasets to boost data breadth, with thorough decontamination to avoid overlap with mainstream evaluation benchmarks. All samples are converted to ChatML to reduce alignment tax in the subsequent post-training stage. After processing, we obtain 182B tokens, from which we randomly sample 90B tokens. To stabilize the training distribution, we replay 30B tokens from Stage 1 and 180B from Stage 2, yielding an effective volume of 300B tokens. Consequently, this stage consolidates and extends capabilities acquired during pre-training under diverse supervised signals.


\subsection{Training Stability and Adaptive Configuration}

We use the \textit{flame}~\cite{zhang2025flame} framework for pre-training, built on \textit{torchtitan}~\cite{liang2025torchtitan}. During training, we prioritized stability over aggressive scaling, making several key adjustments based on empirical observations of training dynamics. These decisions were critical for achieving stable convergence with recurrent architectures, which exhibit different optimization characteristics compared to standard transformers.

\paragraph{Recurrent Step Reduction for Stability.} Our initial experiments with 8 recurrent steps in Stage~1a (Stable Training I) led to loss spikes and gradient oscillations. We hypothesize this stems from compounded gradient flow through multiple recurrent iterations, which can amplify small perturbations. Consequently, we reduced the recurrent steps from 8 to 4 in Stage~1b (Stable Training II in \Cref{fig:training_process}), which balanced computational depth with training stability.

\paragraph{Batch Size Scaling.} To further enhance stability, we progressively increased the batch size from 4M to 8M tokens. Larger batch sizes provide more stable gradient estimates, which is particularly important for recurrent architectures where gradient flow through multiple iterations can introduce additional variance.

\paragraph{KL Divergence Coefficient Reduction.} We strategically reduced $\beta$ in \Cref{eqdef:entropy-reg-obj} from 0.1 in Stage 1a to 0.05 in later stages. This reduction serves dual purposes: (1) it decreases the conflicting gradients between task loss and the KL penalty, leading to more stable optimization, and (2) it reduces the ``pull'' from the uniform prior, allowing the model greater freedom to explore beneficial depth patterns without being artificially constrained. This adjustment allowed the model to learn useful depth patterns without undue constraint.

\paragraph{Optimization Configuration.} 
Throughout all stages, we use AdamW optimizer with weight decay set to 0.1, $\beta_1 = 0.9$, $\beta_2 = 0.95$, and gradient clipping at 1.0. These conservative settings were chosen specifically to maintain stability with recurrent architectures.

\paragraph{Learning Rate Considerations.} 
We empirically found that recurrent architectures require smaller learning rates than parameter-matched Transformers. Given compute constraints, we did not run exhaustive LR sweeps, but instead, adopted conservative rates that prioritized stable convergence over potentially faster but riskier schedules.

\paragraph{Sequence Length Progression.} 
The sequence length is progressively increased across stages: 4K tokens for both pre-training phases, 16K for CT annealing, 64K for long-context training, and 32K for mid-training. This progression stabilizes optimization while expanding context capacity with training throughput.

\subsubsection{Stage-wise Training Details}

\begin{itemize}[itemsep=0.0pt,topsep=0pt,leftmargin=*]
    \item \textbf{Stage 1a: Pre-training Phase I (Exploration).} We initialize training with 8 recurrent steps. The learning rate follows a Warmup-Stable schedule with a peak of $3 \times 10^{-4}$. The sequence length is 4K tokens with an initial batch size of 4M tokens, gradually increased to 8M for stability. During this phase, we observed training instabilities that prompted subsequent architectural adjustments.
    \item \textbf{Stage 1b: Pre-training Phase II with Stability-Driven Upcycling.} After identifying stability issues in Stage 1a, we reduced the recurrent steps from 8 to 4. To maintain computational efficiency while improving stability, we split our approach into two variants:
    \begin{itemize}
        \item \textbf{1.4B Ouro}: Uses the original 24 pre-trained layers with 4 recurrent steps
        \item \textbf{2.6B Ouro}: Upcycles 24 layers to 48 via layer duplication with 4 recurrent steps
    \end{itemize}
    The recurrent nature of our architecture makes this upcycling process particularly smooth, as the shared weights across iterations naturally facilitates layer duplication without the typical instabilities seen in standard transformer upcycling.
    
    \item \textbf{Stage 2: CT Annealing.} The learning rate is annealed to $3 \times 10^{-5}$ while exposing the model to high-quality training data. The recurrent steps remain at 4, having proven optimal for the stability-performance trade-off. The data composition is carefully balanced as shown in Table~\ref{tab:stage2_composition}.
    
    \item \textbf{Stage 3: LongCT.} The batch size is held at 8M tokens. The reduced KL coefficient ($\beta = 0.05$) continues to provide stable training dynamics even with the 64K-length sequences.
    
    \item \textbf{Stage 4: Mid-training.} The learning rate is further reduced to $1 \times 10^{-5}$ with a cosine scheduler to help the model better absorb on this diverse, high-quality dataset.
\end{itemize}

\subsection{Supervised Fine-Tuning}

\paragraph{Data Composition.} We perform SFT on a diverse corpus of approximately 8.3M examples drawn from high-quality public datasets. As shown in Table~\ref{tab:sft_data}, our training mixture emphasizes mathematical reasoning (3.5M examples) and code generation (3.2M examples), while also incorporating scientific reasoning (808K examples) and conversational abilities (767K examples). 

For mathematical reasoning, we combine OpenThoughts3 \citep{guha2025openthoughts} and AceReason-1.1-SFT \citep{liu2025acereason} to provide comprehensive coverage of problem-solving strategies. Our code training data aggregates multiple sources including AceReason-1.1-SFT, OpenCodeReasoning \citep{ahmad2025opencodereasoning}, Llama-Nemotron-Post-Training-Dataset \citep{bercovich2025llama}, and OpenThoughts3, ensuring broad exposure to diverse programming paradigms and reasoning patterns. Scientific reasoning capabilities are developed through OpenThoughts3 and Llama-Nemotron-Post-Training-Dataset, while conversational proficiency is enhanced using the OO1-Chat-747K\footnote{\url{https://huggingface.co/datasets/m-a-p/OO1-Chat-747K}} and DeepWriting-20K \citep{wang2025reverse} datasets.

\paragraph{Training Configuration.} We train for 2 epochs with a maximum sequence length of 32K tokens using the LlamaFactory codebase \citep{zheng2024llamafactory}. We employ the Adam optimizer with a learning rate of $2 \times 10^{-5}$ and $\beta = (0.9, 0.95)$, applying a cosine decay schedule for stable convergence.\footnote{Training was interrupted due to infrastructure issues; we resumed from the last saved checkpoint with a learning rate close to the original cosine decay schedule.}

\begin{table}[htbp]
    \caption{Supervised fine-tuning data composition. The training corpus comprises 8.3M examples across four key capability domains.}
    \label{tab:sft_data}
    \centering
    \begin{tabular}{l|p{10cm}|r}
    \toprule
       Topic  & Data Source & Size \\ \midrule
    Math   &  OpenThoughts3, AceReason-1.1-SFT & 3.5M \\
    Code   & AceReason-1.1-SFT, OpenCodeReasoning, Llama-Nemotron-Post-Training-Dataset, OpenThoughts3 & 3.2M \\ 
    Science & OpenThoughts3, Llama-Nemotron-Post-Training-Dataset & 808K \\
    Chat & OO1-Chat-747K, DeepWriting-20K & 767K \\ \bottomrule
    \end{tabular}
\end{table}

\subsection{Reinforcement Learning Attempts}

Following the SFT stage, we conducted exploratory RLVR (Reinforcement Learning with Verifiable Rewards) alignment experiments using DAPO~\cite{yu2025dapo} and GRPO~\cite{shao2024deepseekmath} on the DAPO-17K dataset. These attempts did not yield significant performance gains over the final SFT checkpoint. The primary issue stemmed from the model's dynamic early-exit mechanism. vLLM/SGLang provide fast rollouts via a fixed execution path, which breaks under \ut{}'s variable-depth computation.  

We tried two approaches, neither successful:

\begin{enumerate}
    \item \textbf{Off-policy rollouts:} We generate full four-step rollouts in vLLM, yielding four logit candidates per token. We then selected the first token to exceed the termination threshold to simulate an early exit. For updates, we used the cumulative loss up to that step, discarding later tokens and losses. This off-policy mismatch, i.e., where tokens are produced with the final depth, and losses computed at an earlier depth, did not improve performance.

    \item \textbf{Fixed 4-Round RL:} To avoid off-policy issues, we performed rollouts and updates at a fixed four recurrent steps. Training progressed normally but performance did not surpass the SFT checkpoint. A like cause is scale: after having already undergone extensive SFT, these smaller models may have limited headroom for RL gains. Interestingly, the model still used fewer rounds at inference when beneficial despite being trained at four rounds. The mechanism behind this generalization remains unclear.
\end{enumerate}

We will further explore RL alignment for this architecture as we continue to develop infrastructure that can fully support \ut{}'s dynamic computation.
\newpage
\section{Experiments}
\subsection{Base Model Evaluation}

We conduct comprehensive evaluations of the Ouro base models trained on 7.7T tokens using the \ut{} architecture. The evaluation focuses on their performance across general knowledge, reasoning, mathematics, science, coding, and multilingual capabilities. All benchmarks are evaluated using \texttt{lm-eval-harness}~\citep{eval-harness} and \texttt{evalplus}~\citep{evalplus} frameworks with settings detailed in Appendix.~\ref{sec:appendix_eval}.

For the base model baselines, we compare our Ouro models with leading open-source base models, including Qwen2.5 \citep{qwen2}, Qwen3 \citep{qwen3}, Gemma3 \citep{team2025gemma3}, Llama3.1 \citep{dubey2024llama}, and Llama3.2 \citep{dubey2024llama} series base models. All models are evaluated using the same evaluation pipeline to ensure fair comparison.

\begin{table}[htbp]
\centering
\caption{\textbf{Comparison of 1.4B  \ut{} model with 1-4B parameter baselines.} The best score is \textbf{bolded}, and the second-best is \underline{underlined}. \ut{}'s column is highlighted in gray.}
\label{tab:base-1.4B}
\small
\setlength{\tabcolsep}{2.5pt}
\begin{tabular}{@{}lcccccccccc@{}}
\toprule
 & \textbf{Gemma3} & \textbf{Llama3.2} & \textbf{Qwen2.5} & \textbf{Qwen3} & \textbf{Qwen2.5} & \textbf{Llama3.2} & \textbf{Qwen3} & \textbf{Gemma3} & \cellcolor{lightergray}\textbf{Ouro} \\
 & \textbf{1B} & \textbf{1.2B} & \textbf{1.5B} & \textbf{1.7B} & \textbf{3B} & \textbf{3B} & \textbf{4B} & \textbf{4B} & \cellcolor{lightergray}\textbf{1.4B R4} \\
\midrule
Architecture & Dense & Dense & Dense & Dense & Dense & Dense & Dense & Dense & \cellcolor{lightergray}\ut{}\\
\# Params & 1.0B & 1.0B & 1.5B & 1.7B & 3.0B & 3.0B & 4.0B & 4.0B & \cellcolor{lightergray}1.4B\\
\# Tokens & 2T & 9T & 18T & 36T & 18T & 9T & 36T & 4T & \cellcolor{lightergray}7.7T\\
\midrule
\multicolumn{10}{c}{\textit{General Tasks}} \\
\midrule
MMLU & 39.85 & 45.46 & 60.99 & 62.46 & 65.62 & 59.69 & \textbf{73.19} & 58.37 & \cellcolor{lightergray}\underline{67.35} \\
MMLU-Pro & 11.31 & 11.80 & 29.11 & 37.27 & 37.87 & 33.34 & \textbf{51.40} & 34.61 & \cellcolor{lightergray}\underline{48.62} \\
BBH & 30.26 & 30.72 & 43.66 & 53.51 & 55.37 & 39.45 & \underline{70.95} & 66.32 & \cellcolor{lightergray}\textbf{71.02} \\
ARC-C & 39.25 & 41.98 & 54.44 & 55.72 & 55.46 & 52.47 & \textbf{63.65} & \underline{60.92} & \cellcolor{lightergray}\underline{60.92} \\
HellaSwag & 56.12 & 59.35 & 67.73 & 67.09 & 74.54 & 73.09 & \textbf{75.66} & \underline{75.58} & \cellcolor{lightergray}74.29 \\
Winogrande & 58.72 & 62.75 & 66.77 & 66.30 & 70.17 & 69.14 & \underline{71.19} & 71.07 & \cellcolor{lightergray}\textbf{72.30} \\
\midrule
\multicolumn{10}{c}{\textit{Math \& Coding Tasks}} \\
\midrule
GSM8K & 2.05 & 7.05 & 60.73 & 70.28 & \underline{74.60} & 67.20 & 72.86 & 68.69 & \cellcolor{lightergray}\textbf{78.92} \\
MATH500 & 41.00 & 7.40 & 17.60 & 25.80 & 42.60 & 40.80 & 59.60 & \underline{68.60} & \cellcolor{lightergray}\textbf{82.40} \\
HumanEval & 6.70 & 19.50 & 52.40 & 66.50 & 68.90 & 29.90 & \textbf{77.40} & 34.80 & \cellcolor{lightergray}\underline{74.40} \\
HumanEval+ & 5.50 & 17.40 & 46.30 & 59.80 & 62.20 & 26.20 & \textbf{70.70} & 29.30 & \cellcolor{lightergray}\underline{67.40} \\
MBPP & 12.40 & 35.70 & 60.30 & 68.00 & 63.00 & 50.30 & \textbf{78.80} & 60.60 & \cellcolor{lightergray}\underline{73.00} \\
MBPP+ & 10.10 & 29.10 & 50.00 & 58.50 & 54.20 & 39.70 & \textbf{65.90} & 51.10 & \cellcolor{lightergray}\underline{62.70} \\
\bottomrule
\end{tabular}
\end{table}

\begin{table}[htbp]
\centering
\caption{\textbf{Comparison of 2.6B \ut{} model with 3-12B parameter baselines.} The best score is \textbf{bolded}, and the second-best is \underline{underlined}. \ut{}'s column is highlighted in gray.}
\label{tab:base-2.6B}
\small
\setlength{\tabcolsep}{2.5pt}
\begin{tabular}{@{}lccccccccc@{}}
\toprule
 & \textbf{Qwen2.5} & \textbf{Llama3.2} & \textbf{Qwen3} & \textbf{Gemma3} & \textbf{Qwen2.5} & \textbf{Llama3.1} & \textbf{Qwen3} & \textbf{Gemma3} & \cellcolor{lightergray}\textbf{Ouro} \\
 & \textbf{3B} & \textbf{3B} & \textbf{4B} & \textbf{4B} & \textbf{7B} & \textbf{8B} & \textbf{8B} & \textbf{12B} & \cellcolor{lightergray}\textbf{2.6B R4} \\
\midrule
Architecture & Dense & Dense & Dense & Dense & Dense & Dense & Dense & Dense & \cellcolor{lightergray}\ut{}\\
\# Total Params & 3.0B & 3.0B & 4.0B & 4.0B & 7.0B & 8.0B & 8.0B & 12.0B & \cellcolor{lightergray}2.6B\\
\# Trained Tokens & 18T & 9T & 36T & 4T & 18T & 15T & 36T & 12T & \cellcolor{lightergray}7.7T\\
\midrule
\multicolumn{10}{c}{\textit{General Tasks}} \\
\midrule
MMLU & 65.62 & 59.69 & 73.19 & 58.37 & 74.20 & 73.02 & \textbf{76.63} & 72.14 & \cellcolor{lightergray}\underline{74.60} \\
MMLU-Pro & 37.87 & 33.34 & 51.40 & 34.61 & 43.55 & 43.24 & \underline{53.72} & 49.21 & \cellcolor{lightergray}\textbf{55.73} \\
BBH & 55.37 & 39.45 & 71.14 & 66.32 & 53.72 & 71.56 & \underline{77.65} & 78.41 & \cellcolor{lightergray}\textbf{80.46} \\
ARC-C & 55.46 & 52.47 & 63.65 & 60.75 & 63.65 & 60.75 & 66.10 & \textbf{72.44} & \cellcolor{lightergray}\underline{66.40} \\
HellaSwag & 74.54 & 73.09 & 75.66 & 75.58 & 79.98 & \underline{81.97} & 79.60 & \textbf{83.68} & \cellcolor{lightergray}79.69 \\
Winogrande & 70.17 & 69.14 & 71.19 & 71.27 & 76.48 & \underline{77.11} & 76.80 & \textbf{77.74} & \cellcolor{lightergray}75.85 \\
\midrule
\multicolumn{10}{c}{\textit{Math \& Coding Tasks}} \\
\midrule
GSM8K & 74.60 & 67.20 & 72.86 & 68.69 & 81.50 & 78.17 & \textbf{83.09} & 77.18 & \cellcolor{lightergray}\underline{81.58} \\
MATH500 & 42.60 & 40.80 & 59.60 & 68.60 & 61.20 & 52.90 & 62.30 & \underline{83.20} & \cellcolor{lightergray}\textbf{90.85} \\
HumanEval & 68.90 & 29.90 & 77.70 & 34.80 & 79.30 & 38.40 & \textbf{84.80} & 46.30 & \cellcolor{lightergray}\underline{78.70} \\
HumanEval+ & 62.20 & 26.20 & 70.70 & 29.30 & 70.60 & 31.10 & \textbf{75.30} & 37.20 & \cellcolor{lightergray}\underline{70.70} \\
MBPP & 63.00 & 50.30 & 78.80 & 60.60 & 73.80 & 62.40 & \underline{79.00} & 73.50 & \cellcolor{lightergray}\textbf{80.40} \\
MBPP+ & 54.20 & 39.70 & 65.90 & 51.10 & 63.50 & 51.60 & \textbf{67.90} & \underline{66.10} & \cellcolor{lightergray}\underline{66.60} \\
\bottomrule
\end{tabular}
\end{table}
\paragraph{Summary of Evaluation Results}
Based on the overall evaluation results, we highlight key conclusions about our base models:
\begin{enumerate}[label=(\arabic*)]
\item Our 1.4B parameter Ouro model (with 4 recurrent steps) achieves performance comparable to the 4B Qwen3-Base across most benchmarks. Notably, it matches or exceeds the 4B model on challenging reasoning tasks such as BBH (71.02 vs 70.95), GSM8K (78.92 vs 72.86) and MATH500 (82.40 vs 59.60)

\item The 2.6B parameter Ouro model outperforms dense models up to 8B parameters on reasoning-intensive benchmarks. It achieves 55.73 on MMLU-Pro, 80.46 on BBH and 90.85 on MATH500, surpassing the 8B Qwen3-Base (53.72, 77.65 and 62.30 respectively).

\item The recurrent architecture shows particular strength on tasks requiring multi-step reasoning and knowledge manipulation, with the most pronounced gains observed on MMLU-Pro, BBH, GSM8K and MATH500 benchmarks, validating our hypothesis that iterative computation enhances reasoning capabilities.
\end{enumerate}


\subsection{Reasoning Model Evaluation}
We evaluate the reasoning capabilities of our Ouro reasoning models (\textbf{Ouro-Thinking}) with 4 recurrent steps on challenging mathematical and scientific benchmarks that require multi-step problem solving and deep reasoning. The evaluation includes AIME 2024/2025 (American Invitational Mathematics Examination), OlympiadBench, GPQA, SuperGPQA, BeyondAIME, and HLE, representing some of the most challenging reasoning tasks in the field.

\begin{table}[t]
\centering
\caption{Performance comparison across different benchmarks. For AIME24 and AIME25, we report pass@1/pass@10 metrics. The best score is \textbf{bolded}, and the second-best is \underline{underlined}.}
\label{tab:reasoning_comparison}
\resizebox{\textwidth}{!}{
\begin{tabular}{l|cc|cc|ccccc}
\toprule
\textbf{Model} & \multicolumn{2}{c|}{\textbf{AIME24}} & \multicolumn{2}{c|}{\textbf{AIME25}} & \textbf{Olympiad} & \textbf{Beyond} & \textbf{HLE} & \textbf{Super} & \textbf{GPQA} \\
 & pass@1 & pass@10 & pass@1 & pass@10 & \textbf{bench} & \textbf{AIME} &  & \textbf{GPQA} &  \\
\midrule
\rowcolor{lightergray}\textbf{Ouro-1.4B-Thinking-R4}  & \underline{65.0} & \underline{83.3} & 46.3 & 73.3 & 71.6 & 34.0 & \underline{5.21} & 47.4 & 45.5 \\
\rowcolor{lightergray}\textbf{Ouro-2.6B-Thinking-R4}  & 64.7 & \textbf{90.0} & 50.3 & \textbf{76.7} & \textbf{76.4} & \textbf{39.0} & \textbf{5.58} & \textbf{53.7} & 52.7 \\
\midrule
Qwen3-1.7B & 32.0 & 55.6 & 22.0 & 33.3 & 56.4 & 15.0 & 4.13 & 35.9 & 34.0 \\
Qwen3-4B & 61.3 & 75.0 & \underline{51.3} & 63.3 & 73.2 & 31.0 & \underline{5.21} & \underline{51.9} & \underline{54.5} \\
Qwen3-8B & \textbf{73.0} & \underline{86.7} & \textbf{66.7} & \textbf{81.3} & \underline{75.3} & \underline{38.0} & 2.22 & 48.0 & \textbf{59.1} \\
\midrule
Deepseek-Distill-Qwen-1.5B & 29.6 & 66.7 & 23.0 & 43.33 & 56.44 & 9.0 & 4.2 & 26.5 & 33.2 \\
Deepseek-Distill-Qwen-7B & 57.3 & 83.3 & 36.0 & 73.3 & 72.0 & 30.0 & 5.14 & 46.6 & 51.0 \\
\bottomrule
\end{tabular}
}
\end{table}

\paragraph{Benchmarks.}
\begin{itemize}
  \item \textbf{AIME 2024/2025}~\cite{HuggingFaceH4_2024_AIME2024}. 30 questions per year from AIME I and II; integer answers 0--999.
  \item \textbf{OlympiadBench}~\cite{He2024OlympiadBench}. Olympiad-level bilingual scientific problems; supports images for multimodal inputs.
  \item \textbf{GPQA}~\cite{rein2023gpqa}. 448 graduate-level multiple-choice questions in biology, physics, and chemistry; search-resistant design.
  \item \textbf{SuperGPQA}~\cite{MAPTeam2025SuperGPQA}. GPQA scaled to about 285 graduate disciplines; curated to remain challenging.
  \item \textbf{BeyondAIME}~\cite{ByteDanceSeed_2025_BeyondAIME}. Hard integer-answer math beyond AIME; emphasizes contamination resistance.
  \item \textbf{HLE}~\cite{Phan2025HLE}. Multi-disciplinary closed-ended benchmark; expert-written with public splits and a private test set.
\end{itemize}

\paragraph{Models compared.}
We report results for Ouro-1.4B-Thinking and Ouro-2.6B-Thinking, which are \ut{}-based looped language models with iterative depth. As baselines we include Qwen3-1.7B, Qwen3-4B, Qwen3-8B, DeepSeek-Distill-Qwen-1.5B, and DeepSeek-Distill-Qwen-7B. We use size-matched baselines whenever available, otherwise we compare to the next larger widely used model.

\paragraph{Evaluation protocol.}
All systems are evaluated with a single in-house harness and identical prompting. We adopt an LLM-as-judge protocol across benchmarks with a fixed rubric and tie-breaking policy. Unless otherwise noted, decoding uses \texttt{temperature = 1.0} and \texttt{top\_p = 0.7} for every model.

\paragraph{Evaluation results.}
Table~\ref{tab:reasoning_comparison} summarizes outcomes. Iterative reasoning in the \ut{} architecture provides consistent gains on these tasks. The 1.4B Ouro model with 4 recurrent steps reaches 71.55 on OlympiadBench (vs. 73.18 for Qwen3-4B) and 34.0 on BeyondAIME (vs. 31.0 for Qwen3-4B). The 2.6B with 4 recurrent steps variant scores 76.44 on OlympiadBench (vs. 75.25 for Qwen3-8B) and 39.0 on BeyondAIME (vs. 38.0 for Qwen3-8B).

\subsection{Performance by Recurrent Depth and Extrapolation}
\begin{table}[htbp]
\centering
\small
\caption{Performance of the Ouro 1.4B \textbf{base model} across different recurrent steps (C-QA is CommonsenseQA~\cite{talmor-etal-2019-commonsenseqa}). Steps 5-8 represent extrapolation, as the model was trained with a maximum of 4 steps. Performance peaks at the trained depth ($T=4$) and then degrades.}
\label{tab:extrapolation_base_1_4b}
\begin{tabular}{lcccccc}
\toprule
\textbf{UT Step} & \textbf{ARC-C} & \textbf{ARC-E} & \textbf{C-QA} & \textbf{HellaSwag} & \textbf{MMLU} & \textbf{Winogrande} \\
& (25-shot) & (8-shot) & (10-shot) & (10-shot) & (5-shot avg) & (5-shot) \\
\midrule
1 & 37.63 & 63.85 & 44.64 & 55.24 & 41.21 & 56.99 \\
2 & 54.86 & 80.30 & 67.98 & 71.15 & 60.43 & 66.69 \\
3 & 59.47 & 83.33 & 74.37 & 74.07 & 66.71 & 71.35 \\
4 & \textbf{60.92} & \textbf{83.96} & \textbf{75.43} & \textbf{74.29} & \textbf{67.45} & \textbf{72.30} \\
\midrule[0.8pt]
\multicolumn{7}{l}{\textit{Extrapolation (Trained on T=4)}} \\
\midrule
5 & 58.96 & 82.91 & 75.35 & 73.72 & 66.64 & 70.32 \\
6 & 59.73 & 82.58 & 74.94 & 72.77 & 65.77 & 71.03 \\
7 & 58.96 & 81.99 & 74.28 & 72.35 & 65.28 & 70.09 \\
8 & 58.19 & 82.07 & 73.55 & 71.60 & 64.49 & 69.30 \\
\bottomrule
\end{tabular}
\end{table}

\begin{table}[htbp]
\centering
\small
\caption{Performance of the Ouro 2.6B \textbf{base model} across different recurrent steps (C-QA is CommonsenseQA~\cite{talmor-etal-2019-commonsenseqa}). Steps 5-8 represent extrapolation, as the model was trained with a maximum of 4 steps. Performance is strongest around the trained depth ($T=4$) and shows varied degradation patterns during extrapolation.}
\label{tab:extrapolation_base_2_6b}
\begin{tabular}{lcccccc}
\toprule
\textbf{UT Step} & \textbf{ARC-C} & \textbf{ARC-E} & \textbf{C-QA} & \textbf{HellaSwag} & \textbf{MMLU} & \textbf{Winogrande} \\
& (25-shot) & (8-shot) & (10-shot) & (10-shot) & (5-shot avg) & (5-shot) \\
\midrule
1 & 47.95 & 72.39 & 57.58 & 68.94 & 51.55 & 61.48 \\
2 & 62.37 & 85.23 & 76.90 & 77.61 & 67.63 & 70.48 \\
3 & 65.36 & 87.33 & 79.77 & 79.12 & 73.57 & 74.35 \\
4 & \textbf{66.38} & \textbf{86.95} & \textbf{81.65} & \textbf{79.56} & \textbf{74.60} & 75.53 \\
\midrule[0.8pt]
\multicolumn{7}{l}{\textit{Extrapolation (Trained on T=4)}} \\
\midrule
5 & 65.36 & 86.83 & 81.24 & 79.57 & 74.43 & \textbf{75.93} \\
6 & 65.02 & 86.74 & 81.08 & 79.63 & 73.79 & 75.37 \\
7 & 65.44 & 86.57 & 80.75 & 79.59 & 72.92 & 75.77 \\
8 & 64.76 & 86.49 & 81.08 & 79.50 & 72.24 & 74.59 \\
\bottomrule
\end{tabular}
\end{table}

We analyze the Ouro model's performance as a function of its recurrent computational depth. Our models were trained with a maximum of 4 recurrent steps ($T=4$). We investigate this behavior for both our \textbf{base models} and our SFT \textbf{Ouro-Thinking} models.

\paragraph{Base Model Performance.}
Tables~\ref{tab:extrapolation_base_1_4b} and \ref{tab:extrapolation_base_2_6b} present the performance of the Ouro 1.4B and 2.6B \textbf{base models}, respectively, evaluated at depths from $T=1$ to $T=8$.

For both base models, performance on standard benchmarks (e.g., MMLU, ARC-C) generally improves up to the trained depth of $T=4$. Steps $T=5$ through $T=8$ represent extrapolation beyond the training configuration. As shown in both tables, benchmark performance sees a moderate degradation when extrapolating, with a noticeable drop compared to the peak at $T=4$.

However, this degradation in task-specific performance contrasts sharply with the model's safety alignment. As detailed in \Cref{sec:safety}, the model's safety improves as the number of recurrent steps increases, even into the extrapolated regime ($T>4$). This suggests that while the model's fine-grained knowledge for benchmarks may falter beyond its training depth, the iterative refinement process continues to enhance its safety alignment.

\paragraph{Reasoning Model (SFT) Performance.}

\begin{table*}
\centering
\caption{\textbf{Performance of Ouro-1.4B-Thinking model by recurrent step.} The model was trained at $T=4$. Performance peaks around $T=4$ or $T=5$. All scores are percentages (0-100).}
\label{tab:extrapolation_thinking_1_4b}
\small
\setlength{\tabcolsep}{5pt}
\begin{tabular}{@{}l|cccccccc@{}}
\toprule
\textbf{Benchmark} & \textbf{T=1} & \textbf{T=2} & \textbf{T=3} & \textbf{T=4} & \textbf{T=5} & \textbf{T=6} & \textbf{T=7} & \textbf{T=8} \\
\midrule
OlympiadBench & 2.22 & 59.70 & 70.67 & 71.55 & \textbf{72.30} & 69.48 & 69.04 & 66.81 \\
SuperGPQA     & 2.03 & 33.07 & 44.50 & 47.37 & \textbf{48.73} & 46.15 & 45.29 & 42.88 \\
AIME 2024     & 0.00 & 37.33 & 62.33 & \textbf{65.00} & 60.67 & 50.67 & 42.33 & 38.67 \\
AIME 2025     & 0.33 & 25.00 & 43.33 & 46.30 & \textbf{47.00} & 43.00 & 41.00 & 38.00 \\
\bottomrule
\end{tabular}
\end{table*}

\begin{table*}
\centering
\caption{\textbf{Performance of Ouro-2.6B-Thinking model by recurrent step.} The model was trained at $T=4$. Performance peaks at $T=3$ or $T=4$. All scores are percentages (0-100).}
\label{tab:extrapolation_thinking_2_6b}
\small
\setlength{\tabcolsep}{5pt}
\begin{tabular}{@{}l|cccccccc@{}}
\toprule
\textbf{Benchmark} & \textbf{T=1} & \textbf{T=2} & \textbf{T=3} & \textbf{T=4} & \textbf{T=5} & \textbf{T=6} & \textbf{T=7} & \textbf{T=8} \\
\midrule
OlympiadBench & 18.96 & 68.59 & 75.56 & \textbf{76.44} & 71.85 & 69.19 & 57.63 & 39.26 \\
SuperGPQA     & 15.66 & 48.58 & \textbf{56.70} & 53.68 & 56.45 & 55.44 & 53.32 & 46.84 \\
AIME 2024     & 3.00 & 52.00 & \textbf{70.33} & 64.70 & 57.00 & 56.33 & 49.67 & 39.00 \\
AIME 2025     & 2.00 & 40.67 & \textbf{50.67} & 50.30 & 49.33 & 46.00 & 38.00 & 24.33 \\
\bottomrule
\end{tabular}
\end{table*}
We conduct a similar analysis on our SFT models, \textbf{Ouro-Thinking}, to see how recurrent depth affects specialized reasoning tasks. Results for the 1.4B and 2.6B models are presented in Table~\ref{tab:extrapolation_thinking_1_4b} and Table~\ref{tab:extrapolation_thinking_2_6b}, respectively.

We conduct a similar analysis on our SFT models, \textbf{Ouro-Thinking}, to see how recurrent depth affects specialized reasoning tasks. Results for the 1.4B and 2.6B models are presented in Table~\ref{tab:extrapolation_thinking_1_4b} and Table~\ref{tab:extrapolation_thinking_2_6b}, respectively.

For both SFT models, performance at $T=1$ is very low, confirming that iterative refinement is essential for these complex tasks. Performance generally peaks at or near the trained depth, but shows slightly different patterns. The 1.4B model (Table~\ref{tab:extrapolation_thinking_1_4b}) peaks around $T=4$ or $T=5$. The 2.6B model (Table~\ref{tab:extrapolation_thinking_2_6b}) tends to peak slightly earlier, at $T=3$ or $T=4$. Interestingly, neither model peaks strictly at $T=4$ across all tasks, unlike the base model evaluations which are often logit-based. This may suggest that the longer decoding required for these reasoning tasks allows for a more active exploration of capabilities at different recurrent depths. For both models, performance degrades as they extrapolate to deeper, unseen recurrent steps ($T=6-8$), reinforcing that performance is optimized for the depth seen during training.

\subsection{Early Exit and Adaptive Computation Efficiency}

A defining advantage of the \ut{} architecture lies in its capacity for adaptive computation allocation. Unlike standard transformers with fixed computational budgets, our model can dynamically adjust the number of recurrent steps based on input complexity. This section investigates various strategies for implementing adaptive early exit, comparing their effectiveness in balancing computational efficiency with task performance.

\subsubsection{Early Exit Strategies}
\label{sec:early_exit_strategy}
We explore three distinct approaches to determining when the model should terminate its iterative computation and produce the final output.

\paragraph{Baseline: Static Exit.}
The simplest strategy forces the model to exit at a predetermined recurrent step, regardless of the input characteristics. While this approach provides predictable computational costs, it fails to leverage the model's potential for adaptive resource allocation. We evaluate static exit at steps 1 through 4 to establish performance bounds and understand the relationship between computational depth and accuracy.

\paragraph{Hidden State Difference Threshold.}
This heuristic-based approach monitors the magnitude of representational changes between consecutive recurrent steps. At each step $t$, we compute $\Delta h_t = \|h_t - h_{t-1}\|_2$ and trigger early exit when $\Delta h_t < \epsilon$ for some threshold $\epsilon$. 

\begin{wrapfigure}{r}{0.5\textwidth}
    \centering
    \includegraphics[width=\linewidth]{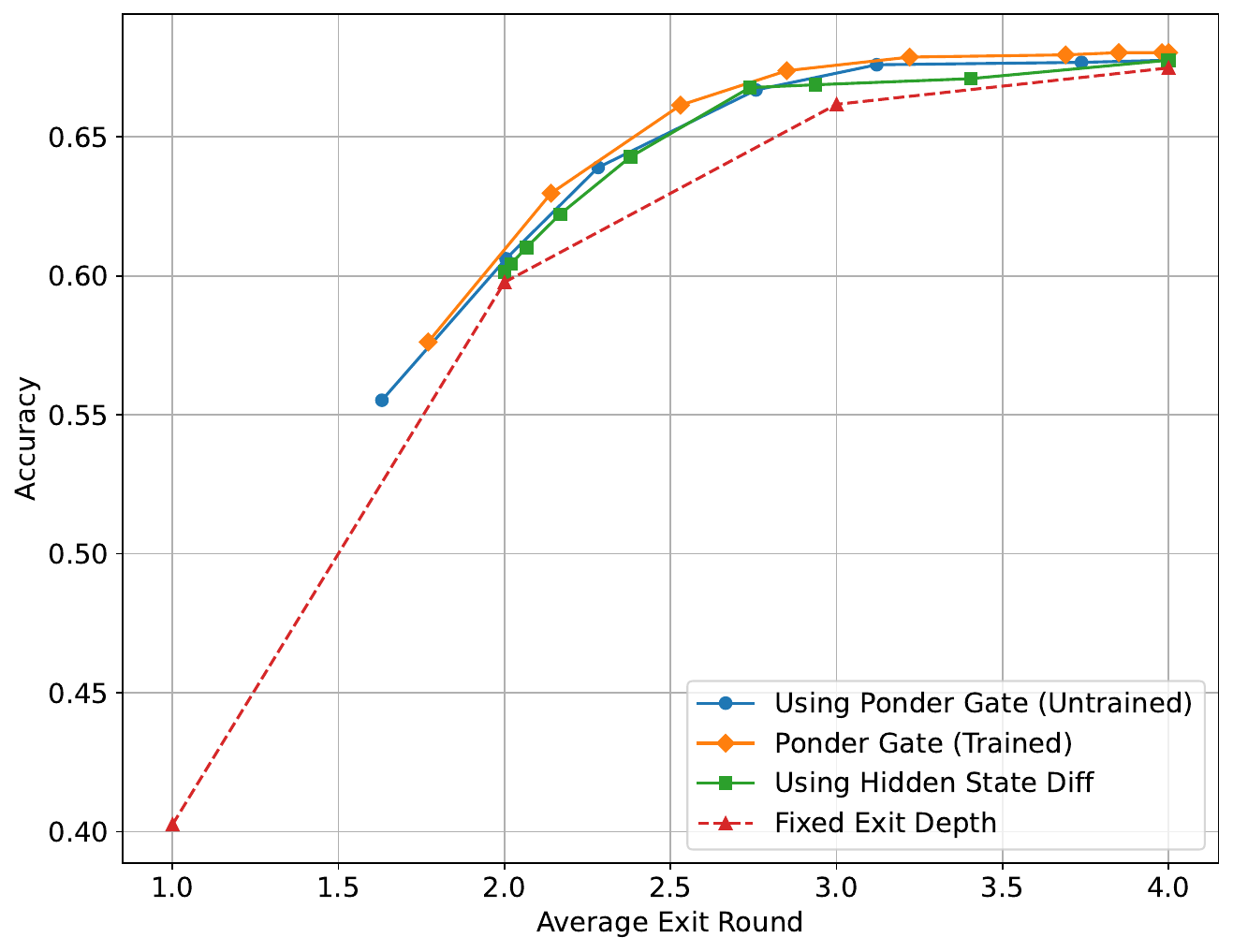}
    \caption{\textbf{Comparison of early exit strategies on MMLU.} We evaluate four approaches across different average exit rounds: static baseline (red triangle), hidden state difference threshold (green squares), Ponder gate from standard pre-training (blue circles), and Ponder gate with specialized adaptive exit training from Section~\ref{ut:adaptive_exit} (orange diamonds). }
    \label{fig:early_exit_comparison}
    \vspace{-0.6cm}
\end{wrapfigure}
\paragraph{Learned Gating with Q-Exit Criterion.}
Our primary approach employs the learned exit gate described in Section~\ref{sec:training_looplm}, which produces step-wise halting probabilities $\lambda_t$ based on the model's current hidden states. During inference, we apply the Q-exit criterion: at each step $t$, we compute the cumulative distribution function $\text{CDF}(t) = \sum_{i=1}^{t} p(i|x)$ and exit when $\text{CDF}(t)$ exceeds the threshold $q \in [0,1]$. The threshold $q$ serves as a deployment-time hyperparameter that controls the compute-accuracy trade-off without requiring model retraining.

We evaluate this strategy under two training configurations. The \textit{untrained} configuration uses the gate as trained during our standard pre-training pipeline with the entropy-regularized objective (uniform prior KL loss). This represents the gate's behavior when jointly optimized with language modeling throughout Stages 1-4. The \textit{trained} configuration additionally applies the specialized adaptive exit loss described in Section~\ref{ut:adaptive_exit}, which explicitly teaches the gate to base stopping decisions on observed task loss improvements.

\paragraph{Experimental Results.}
Figure~\ref{fig:early_exit_comparison} presents the accuracy-efficiency trade-off curves for all strategies on the MMLU benchmark. By varying the exit threshold (or static exit step for baseline), we obtain multiple operating points for each method, enabling direct comparison at equivalent computational budgets measured by average exit round.

Several key findings emerge from this analysis:
\begin{enumerate}
    \item The Ponder gate with specialized adaptive exit training  achieves the best accuracy at every computational budget, demonstrating that the loss improvement-based training signal described in Section~\ref{ut:adaptive_exit} provides clear benefits over standard entropy regularization. At an average exit round of 2.5, the specialized training reaches 66\% accuracy while the standard gate achieves approximately 64\%;
    \item Even without specialized training, the Ponder gate from standard pre-training substantially outperforms the static baseline, validating that the entropy-regularized objective with uniform prior successfully enables adaptive computation. The gate learns to differentiate input difficulty through the general training dynamics, though it lacks the explicit supervision to correlate stopping decisions with actual performance improvements. This demonstrates that our base training approach already captures useful signals for resource allocation;
    \item The hidden state difference threshold strategy performs surprisingly competitively, closely tracking both gate configurations. At moderate computational budgets (2-3 average rounds), it achieves accuracy within 1\%-2\% of the specialized trained gate, suggesting that representation stability provides a reasonable proxy for computational convergence. However, the consistently superior performance of the specialized trained gate across all operating points confirms that explicit supervision via the adaptive exit loss captures information beyond what can be inferred from representational dynamics alone.
    \item Comparing the untrained and trained gate configurations reveals the value proposition of the specialized training procedure. The gap between these curves, approximately 2\%-3\% accuracy at most operating points, represents the benefit of teaching the gate to explicitly monitor task loss improvements $I^{(n)}_t$ rather than relying solely on entropy regularization to discover stopping policies. This empirical result validates our design choice to introduce the adaptive exit loss as a specialized training objective.
    \item The baseline's monotonic improvement from 1 to 4 rounds confirms the ``deeper is better'' property while revealing diminishing returns. The dramatic jump from 1.0 to 2 rounds (40\% to 60\% accuracy) contrasts with the marginal gain from 3 to 4 rounds (67.35\% accuracy). This pattern explains why adaptive methods prove effective: most examples achieve near-maximal performance at intermediate depths, with only a minority requiring full computational depth. 
\end{enumerate}

\subsubsection{KV Cache Sharing for Inference Efficiency}

The recurrent nature of our architecture introduces a challenge: naively, each recurrent step requires maintaining its own KV cache, leading to 4$\times$ memory overhead for our 4-step model. We investigate strategies to reduce this overhead through KV cache reuse.

\paragraph{Prefilling Phase}
During the prefilling phase (processing the input prompt), we find that all four recurrent steps require their own KV caches, as each step transforms the representations in ways that cannot be approximated by earlier steps. Attempting to reuse KV caches during prefilling leads to performance degradation (>10 points on GSM8K).

\paragraph{Decoding Phase}
However, during the decoding phase (auto-regressive generation), we discover that KV cache reuse becomes viable. We explore two strategies:

\begin{enumerate}
    \item \textbf{Last-step reuse}: Only maintain KV cache from the final (4th) recurrent step
    \item   \textbf{First-step reuse}: Only maintain KV cache from the first (1st) recurrent step. 
    \item \textbf{Averaged reuse}: Maintain an averaged KV cache across all four steps
   
\end{enumerate}

\begin{table}[htbp]
\centering
\caption{KV cache sharing strategies during decoding. Both last-step and averaged strategies achieve minimal performance loss while reducing memory by 4$\times$.}
\label{tab:kv_cache}
\small
\begin{tabular}{@{}lccc@{}}
\toprule
\textbf{Strategy} & \textbf{GSM8K} & \textbf{MATH-500} & \textbf{Memory Reduction}  \\
\midrule
Full (4$\times$ cache) & 78.92 & 82.40 & 1.00$\times$  \\
First-step only & 18.73 & 8.43 & 4.00$\times$  \\
Last-step only & 78.85 & 80.40 & 4.00$\times$  \\
Averaged & 78.73 & 78.52 & 4.00$\times$ \\
\bottomrule
\end{tabular}
\end{table}

As shown in Table~\ref{tab:kv_cache}, these strategies yield dramatically different outcomes. Reusing only the first step's cache results in a catastrophic performance collapse (e.g., 18.73 on GSM8K, down from 78.92), indicating that the initial representations are insufficient for subsequent decoding steps. In contrast, both the last-step and averaged reuse strategies achieve nearly identical performance (within 0.3 points on GSM8K) to the full cache baseline, while successfully reducing memory requirements by 4×. The last-step strategy performs slightly better than the averaged approach on MATH-500, suggesting that the final recurrent step's representations are most informative for subsequent token generation. This finding enables practical deployment of \ut{} models with memory footprints comparable to standard transformers of similar parameter count.

\section{Understanding \ut{}s Superiority from a Parametric Knowledge Viewpoint}
\label{sec:understanding_ut}
Why \ut{}s achieve far better performance when the parameter counts do not increase? Although potential enhanced reasoning capabilities were observed in \cite{saunshi2025reasoning}, the source of the advantage remains unclear. Specifically, do \ut{}s perform better due to the models' increased \textbf{knowledge capacity} with the same size of parameters? Or do they have a better capability in \textbf{extracting and composing the knowledge} encoded within the parameters? Toward understanding the improvement of the phenomenon, we explore what capabilities are exactly enhanced by simply looping more times. 
In this section, we perform experiments to test the model's abilities to \textit{memorize} factual knowledge in its parameters, and the capabilities of \textit{manipulating and composing} existing knowledge encoded in the parameters based on a set of fully controllable synthetic tasks in \cite{AL2024-knowledge3,Allenzhu2025-canon,yao2025language}.

\subsection{\ut{}s does not increase knowledge capacity}
\label{subsec:capacity}

We first explore the \textit{knowledge capacity}, i.e. the model's storage capacity of facts in the parameters. We aim to answer the first question: do \ut{}s achieve better performance by \textbf{memorizing knowledge} when the parameter count is not increased?

\textbf{Settings.} Following the Capo task setting in Physics of language models \cite{AL2024-knowledge3,Allenzhu2025-canon}, we construct synthetic biographies to test \textbf{how much information the model memorizes}. 
Specifically, we generate several synthetic biographic datasets $\operatorname{bioS}(N)$ with different number of people $N$, and train a series of language models to memorize the information contained in the dataset. Each biography contains the individual's name and five attributes $a_1,a_2,...,a_5$ of the person: gender, birth date, university, major, and employer. The names $n$ and the attributes $a_i$ are randomly selected from a pre-defined set $\mathcal{N}$ and $\mathcal{A}_i$ and combined together as a biography using a random template. Based on the random generation process, we have an information-theoretic lower bound for the model in the minimum bits required to encode all the names and attributes. To check whether the models memorize the biographic information accurately, we look at the probability of predicting the ground-truth attributes with the trained models given the biography context.  
Calculating the sum of cross-entropy loss on each attribute token positions, we can estimate how much information (estimated in bits) has already been memorized in the trained language model, which is our \textbf{knowledge capacity} metric. 

With this metric, we can compare the knowledge capacity between the original model (with only one recurrent step) and the looped model (with 4 recurrent steps) with the same parameter count to investigate whether looping increases knowledge capacity. Moreover, as larger models should encode more information than smaller models, we also aim to investigate whether looped models have a better scaling effect when the size of the model grows. 
We thereby trained GPT-2 style models of different parameter numbers ranging from 1M to 40M (with depth and hidden dimension varied) and measured the number of bits of knowledge learned by each model. We trained on $\operatorname{bioS}(N)$ with $N$ ranging from 20K to 500K individuals for 1000 exposures. More training details are provided in \Cref{appendix:capo}.

\begin{figure}[t]
\centering
\setlength{\tabcolsep}{6pt}
\vspace{-3pt}
\begin{minipage}[t]{0.5\textwidth}
\vspace{-3pt} 
\small
\centering\hspace{-2pt}
\includegraphics[width=\linewidth]{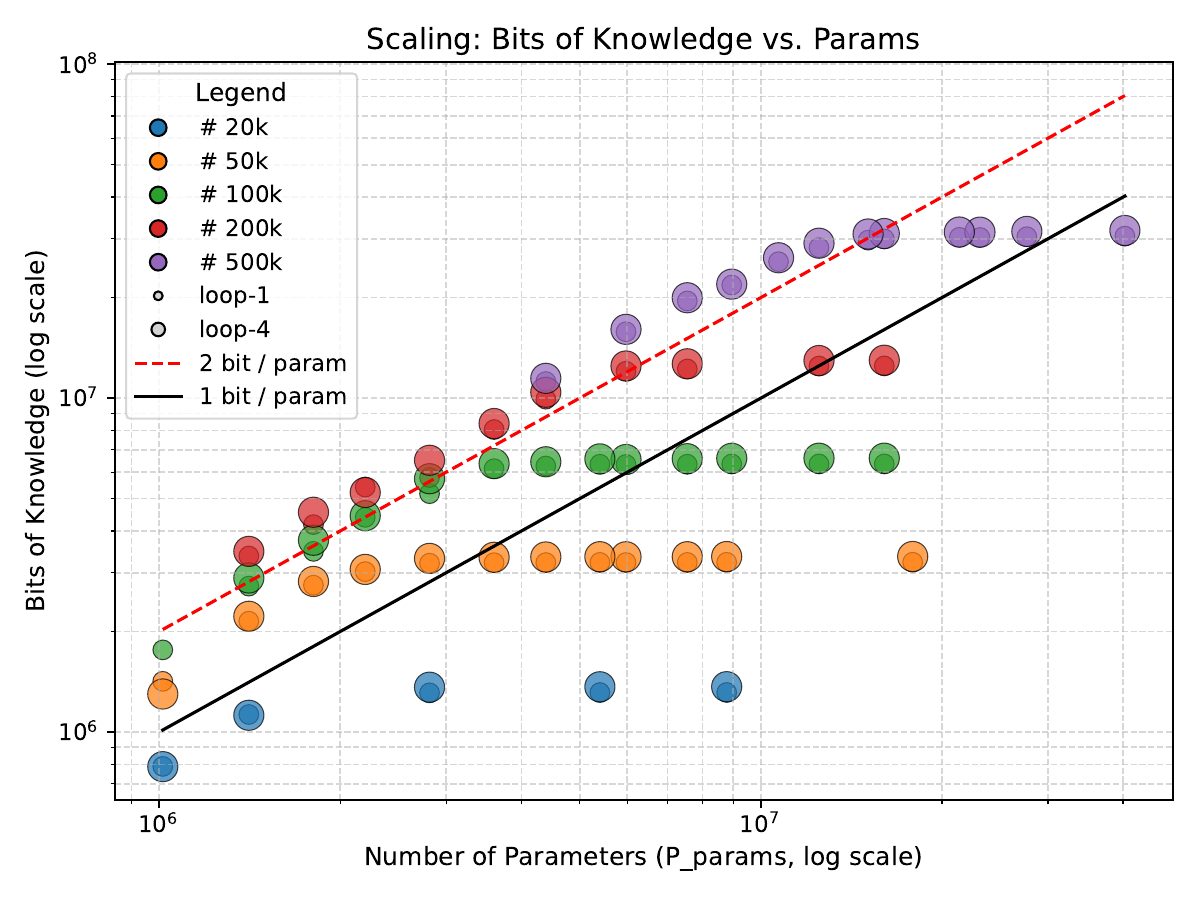}
\end{minipage}
\hfill
\begin{minipage}[t]{0.44\textwidth}
\vspace{-3pt} 
\centering
\small
\rowcolors{3}{white}{white} 
\begin{tabular}{@{}lccc@{}}
\toprule
\textbf{} & \textbf{$L=10$} & \textbf{$L=16$} & \textbf{$L=24$} \\
\rowcolor{gray!15}\multicolumn{4}{c}{\textbf{Baseline model}}\\
\midrule
Base $(12 \otimes 1)$ & 93.6 & 94.4&34.8  \\
\midrule
\rowcolor{gray!15}\multicolumn{4}{c}{\textbf{2 layer model}}\\
Base $(2 \otimes 1)$ &21.5 & 8.4& 7.5 \\
Loop $(2 \otimes 6)$ & \textbf{98.1}& \textbf{96.3} &78.0  \\
\midrule
\rowcolor{gray!15}\multicolumn{4}{c}{\textbf{3 layer model}}\\
Base $(3 \otimes 1)$ &75.4 & 29.8& 11.0 \\
Loop $(3 \otimes 4)$ & \textbf{97.9}& \textbf{95.8} &\textbf{92.2}  \\
\midrule
\rowcolor{gray!15}\multicolumn{4}{c}{\textbf{6 layer model}}\\
Base $(6 \otimes 1)$ & 84.7 &59.5 & 20.0\\
Loop $(6 \otimes 2)$ & 93.4 &88.5 & 35.1\\
\bottomrule
\end{tabular}
\end{minipage}
\caption{\textbf{Left.}We trained both \ut{} and a standard trasnformer baseline with the same parameters on Capo task to compare the knowledge capacity gain by looping more times. With the same parameter count, the looped model and its non-looped baseline has almost the same knowledge capacity measured in bits of knowledge on Capo task. 
\textbf{Right.}  Accuracy of looped/non-looped models on Mano task. Looped models are better than the iso-param $(\{2,3,6\} \otimes 1)$ models. They also achieve better or comparable performance comparing to the iso-flop baseline $(12 \otimes 1)$ model.}
\label{fig:knowledge_capacity_and_mano}
\end{figure}

\textbf{Results.} The results are visualized in the plot ``bits vs. \# of parameters'', where we can observe the comparison between iso-parameter looped and non-looped models. Our results are shown in \Cref{fig:knowledge_capacity_and_mano} (Left): \textbf{looping does not increase \textit{knowledge capacity} nor improve capacity scaling}. Models with and without loops all attain around a similar capacity ratio $\approx 2$ bits/parameter. Therefore, the number of parameters itself can be seen as a direct indicator of knowledge capacity, and \textbf{merely increasing looping does not help enhance knowledge capacity itself}.

\subsection{\ut{}s prevails in knowledge manipulation}
\label{subsec:manipulation}
We have already shown that reusing parameters cannot help the model memorize more atomic factual knowledge. However, natural language is not only about single-hop factual knowledge. In most of the scenarios, predicting the next token requires combining different piece of knowledge, which we called \textbf{knowledge manipulation} \cite{AL2024-knowledge3}. \textit{Does looping and reusing parameters help \ut{}s in tasks that require \textbf{flexible usage of knowledge}?} We further consider two synthetic tasks to investigate the hypothesis on {{knowledge manipulation}} capacity: the synthetic Mano task in \cite{Allenzhu2025-canon} based on modular arithmetic, and a multi-hop QA task in natural language \cite{yao2025language} composing individual facts. 

\textbf{Mano Task.} We first explore the knowledge manipulation task Mano in \cite{Allenzhu2025-canon}, based on a complex tree structure with restricted modular arithmetic knowledge. Models need to solve the task without intermediate thinking process. As illustration, an example could be \texttt{<bos> + * a b c <eos>} requires the model to directly output $(a*b)+c$ mod 23. To solve this task, the model needs to (1) apply the arithmetic rules modulo 23 as the factual knowledge encoded in the parameters, and (2) parse the binary tree structure of the arithmetic to compose all calculations. 

To evaluate the manipulation capability thoroughly, we consider the test accuracy across different difficulty levels based on maximum expression length $L$, which accounts for the number of operations in the sample. The model is trained with online samples with all possible expression lengths $\ell\in[1, L]$ and tested on the maximum expression length $L$. We prepare three levels of difficulties $L=[10, 16, 24]$ to test \ut{}'s superiority over non-looped models given fixed training budget. We train ($\{2,3,6,12\}\otimes 1$) standard transformers as the baselines and several looped models ($k\otimes 12/k$) with $k=2,3,6$. More details are included in Appendix~\ref{appendix:mano}. 

\textbf{Results.} The results in \Cref{fig:knowledge_capacity_and_mano} show that given the same parameters, looped models \textbf{always} outperform their non-looped counterpart for all possible $k\in\{2,3,6\}$. Even with the same number of FLOPs in the model, the looped models can often perform better. This indicates that\textbf{ \ut{} has a better inductive bias towards \textit{knowledge manipulation}}: with the same budget on training samples and computation, \ut{} can achieve comparable or even better performance after training when the task requires manipulation capability (e.g., parsing the arithmetic tree) given limited amount of required knowledge (e.g., modular arithmetic rules).  

\begin{figure}[t]
\centering
\vspace{-5pt}
\setlength{\tabcolsep}{6pt}
\begin{minipage}[t]{0.48\textwidth}
\small
\centering
\includegraphics[width=\linewidth]{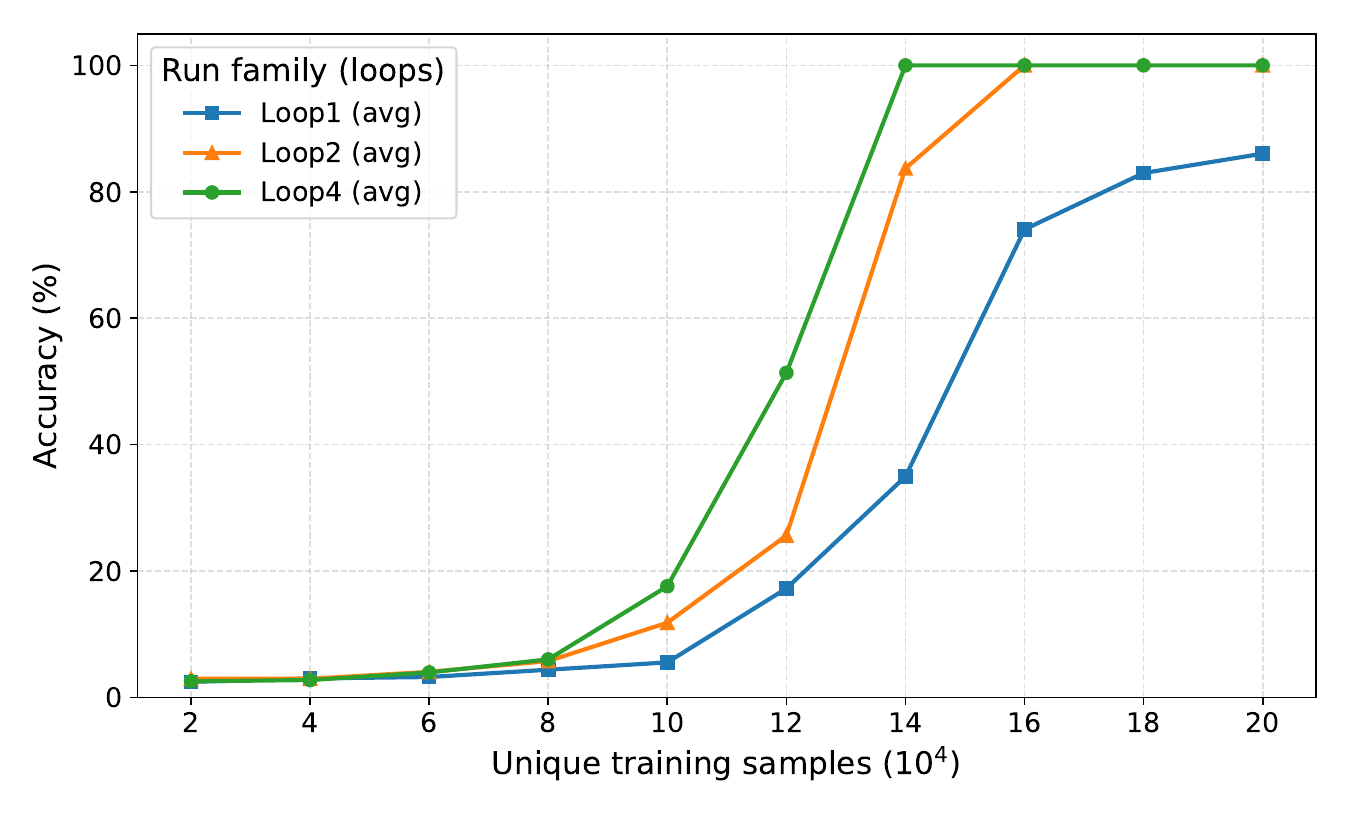}
\end{minipage}
\hfill
\begin{minipage}[t]{0.48\textwidth}
\centering
\includegraphics[width=\linewidth]{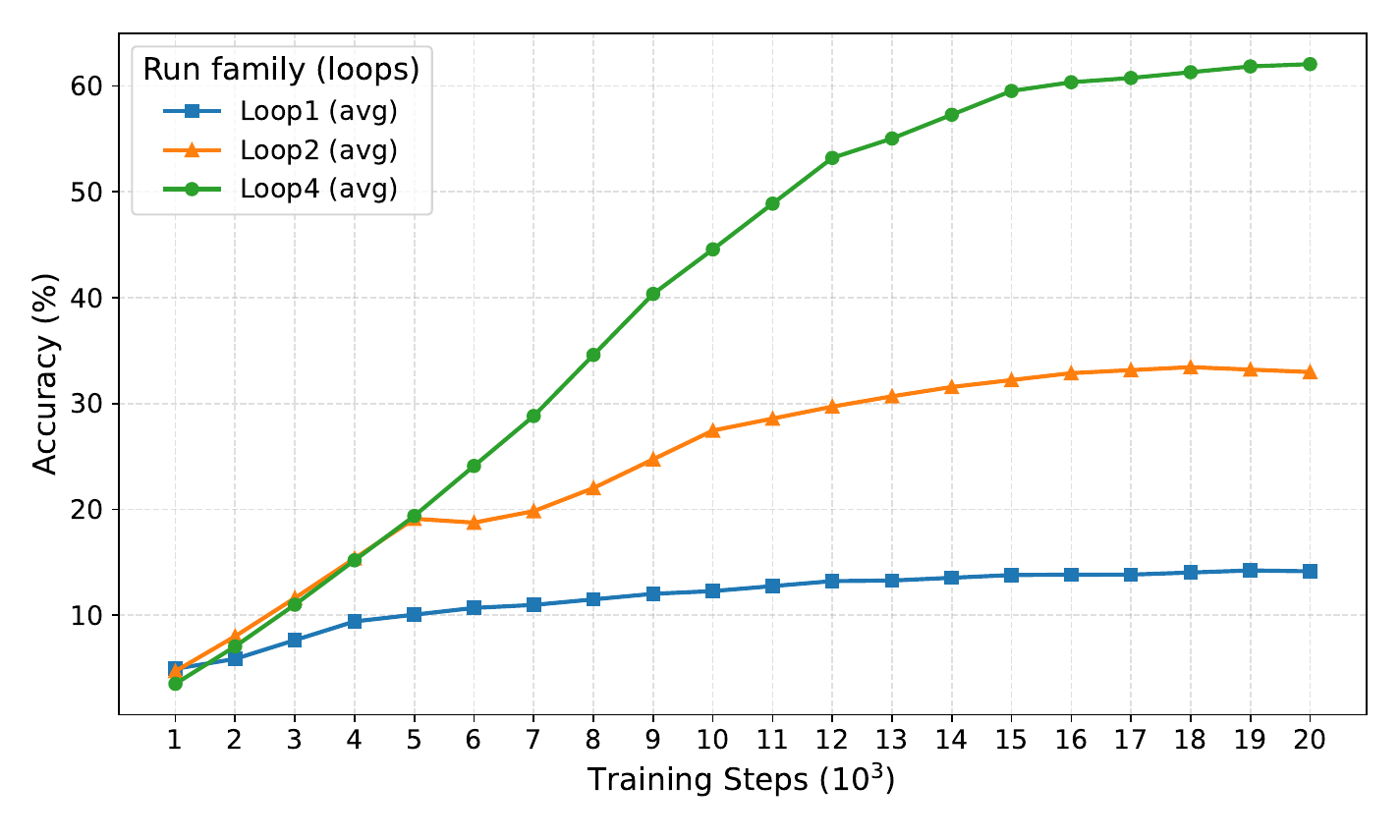}
\end{minipage}
\caption{We trained \ut{}s and standard transformer baselines with the same parameters on Multi-hop QA tasks. To investigate the \textit{sample efficiency} of \ut{}s, we \textbf{vary the number of unique training samples} (from $2.5\%$ to $25\%$ all possible QA pairs) for models with different loops. We compare the final performance using \textbf{the same compute budget in total training tokens}. \textbf{Left.} As shown, models with more loops requires fewer samples to learn the 3-hop QA task.
\textbf{Right.} As an example, we train with $15\%$ of all possible QA pairs (120,000 unique samples) for 20000 steps with context length 1024 and batch size 2048. Models with more loops learn faster and achieve better performance comparing with models without loops.}
\label{fig:multi_hop}
\vspace{-5pt}
\end{figure}

\textbf{Multi-hop QA. } Next, we corroborate our conjecture with a natural language multi-hop reasoning task proposed in \cite{yao2025language}, based on synthetic facts on relations $\mathcal{R}$ between $|\mathcal{E}|$ different individuals, like \textit{The instructor of A is B} and \textit{The teacher of B is C}. The target is to answer multi-hop questions like \textit{`Who is the teacher of the instructor of A?'.} We aim to study whether looping enables the original transformer better learn to perform internal multi-hop reasoning in a natural language setting. Compared to the Mano task, the task requires the model to memorize more factual knowledge with layer-wise data structure, which is closer to practical natural language multi-hop reasoning.

Multi-hop QA tasks require huge amount of samples to learn according to \cite{yao2025language} when training standard transformers. To study whether \ut{}s accelerate the learning process of this multi-hop knowledge manipulation task, 
we consider \textit{sample efficiency} in learning. Specifically, we study \textbf{how many different QA pairs are necessary for the trained model to achieve 100\% accuracy, as well as the performance after training on a fixed budget of unique training samples.} For simplicity, we focus on the task with 3-hop QA pairs. We separate all possible QA pairs into training subsets of different sizes, and compare when each model perfectly generalizes on the leave-out test set.
Similarly to the Mano task, we train a standard ($6\otimes1$) transformer as the baseline, and compare it with looped models ($6\otimes \{2,4\}$) to study the effect of the universal transformer. We also train an iso-flop model ($24\otimes 1$) for comparison.  More details are included in Appendix~\ref{appendix:multihop}.

\textbf{Results.} The results in \Cref{fig:multi_hop} show that looped models generally learn the multi-hop QA task with fewer examples compared to both the non-looped iso-parameter model when the training budget is the same. Moreover, \ut{}s learn the multi-hop task much faster than the non-looped model with the same number of unique QA samples. The improved \textit{sample efficiency} on the multi-hop reasoning task further demonstrates that \ut{} has a better ability to learn to compose and manipulate atomic factual knowledge. 

Based on the results in both Mano and multi-hop QAs, we can conclude that \ut{}s have a better inductive bias towards more \textbf{flexible manipulation} of learned knowledge, instead of increasing the knowledge capacity. The conclusion holds for both synthetic tasks regardless of whether the task is more reasoning-heavy (Mano) or knowledge-heavy (Multi-hop QA). This also corresponds to the analysis (see Appendix \ref{appendix:mmlu_case_study}) on the existing benchmarks (e.g. MMLU): adding more recurrent steps significantly improves the performance on more reasoning-heavy categories, while the improvements on more knowledge-heavy tasks is limited.

\subsection{Discussion: towards understanding why \ut{} helps knowledge manipulation}
\label{subsec:theory_and_discussion}
Why does \ut{} naturally bias towards better manipulation of the knowledge encoded in the parameter space? We conjecture that the reason lies in the inherent recurrent structure of \ut{}. Given that the knowledge capacity is limited by the parameter counts, looping enables \ut{} to better utilize the knowledge encoded in the parameters.
\ut{} can reuse the knowledge in each looped block, \textbf{retrieve} new necessary factual information, or apply \textbf{structured procedures} to obtain the final prediction.

\textbf{Search on the parametric knowledge graph.} During pre-training, language models often obtain an enormous amount of factual knowledge and learn analysis procedures with a rather shallow thinking depth. To perform more challenging tasks, the model needs to use multiple pieces of knowledge in the parameter space, which requires the model to search in-depth in the knowledge graph with directional dependencies formed by the atomic facts or knowledge. \ut{} naturally support an efficient reuse of the knowledge and algorithms stored in the parameter spaces: even though the knowledge piece is not retrieved or used in the previous calculations, the recurrent structure enables \ut{} to redo the procedure and extract necessary information.

Based on the abstraction above, we try to understand why \ut{}s are able to search on knowledge graph without adding more parameters. Specifically, we study the expressivity of \ut{} on a synthetic task. We consider the extensively studied search problem in the literature of \textbf{latent reasoning} \cite{hao2024training, zhu2025reasoning, zhong2025understanding}: \textit{graph reachability} on a knowledge graph. Here, we consider that only part of the knowledge graph $G_{\text{ctx}}$ is included in the context, and most of the knowledge relations $G$ must be encoded in the parameters. The model must learn to compose the context knowledge $G_{\text{ctx}}$ and the learned knowledge $G$. Compared to traditional CoT and recent proposed latent CoT \cite{zhu2025reasoning, hao2024training}, we show that \ut{} is a parallelizable latent reasoning paradigm that requires fewer sequential reasoning steps.
\begin{theorem}[Informal]
    Fix $n\ge 1$ and a hidden knowledge graph $G$ on $n$ vertices. Under the adjacency-column input encoding, full attention, and elementwise threshold operator specified in Appendix~\ref{appendix:theory}, there exists a one-layer, single-head looped transformer of hidden dimension $2n$, with parameters independent of the context graph $G_{\text{ctx}}$, that answers every reachability query on $G+G_{\text{ctx}}$ in $O(1+\log_2\max\{1,D\})$ loops. Here $D$ is the maximum finite shortest-path distance over reachable ordered pairs, and zero-length paths are allowed.
    \label{thm:ut_graph_connectivity}
\end{theorem}
\begin{table*}[ht]
\centering
    \vspace{-7pt}
    \begin{tabular}{@{}cl|cccc@{}}
        \toprule
        &\textbf{Latent reasoning method} & {Discrete CoT} & Continuous CoT & Universal Transformer \\
        \midrule
        &\textbf{Sequential computation steps} & {$O(n^2)$} & {$O(1+D)$} & {$O(1+\log\max\{1,D\})$} \\
        \bottomrule
    \end{tabular}
\end{table*}

The proof and its assumptions are given in Appendix~\ref{appendix:theory}. In this simplified setting, parallel all-pairs path composition yields logarithmically many sequential loops. The comparison above concerns sequential steps under the respective methods' assumptions, rather than matched total computation: our construction uses width $2n$, full attention, and a Boolean threshold. Extending this construction to the causal attention and standard normalization used in our language models remains outside the theorem. The result suggests one mechanism by which recurrent computation can support knowledge manipulation, but does not by itself explain the empirical performance gap.

\textbf{Recurrence improves sample efficiency.} The expressiveness result of \ut{} does not explain why the transformers with loops often learns knowledge manipulation tasks with samples much fewer than its iso-FLOP counterpart. We conjecture that the reason lies again in the \textbf{recurrent structure} of \ut{}. Assuming the reasoning tasks require multiple manipulation and recursion using learned parametric knowledge or algorithmic procedure, the models have to learn a \textit{repeated structure} across layers of different depth. For deep transformer models without looping, they potentially have to explore a large function class where each block of parameters are not tied. The parameter-sharing layers may help the model explore a much smaller realizable hypothesis class, thus reducing the sample complexity of learning those manipulation tasks. It could be a possible statistical reason that \ut{} enjoys a better sample complexity on those reasoning/manipulation tasks.

\section{Safety, Faithfulness and Consistency}
\subsection{Safety}
\label{sec:safety}
\begin{figure}[t]
    \centering
    \begin{subfigure}{\textwidth}
        \centering
        \includegraphics[width=\linewidth]{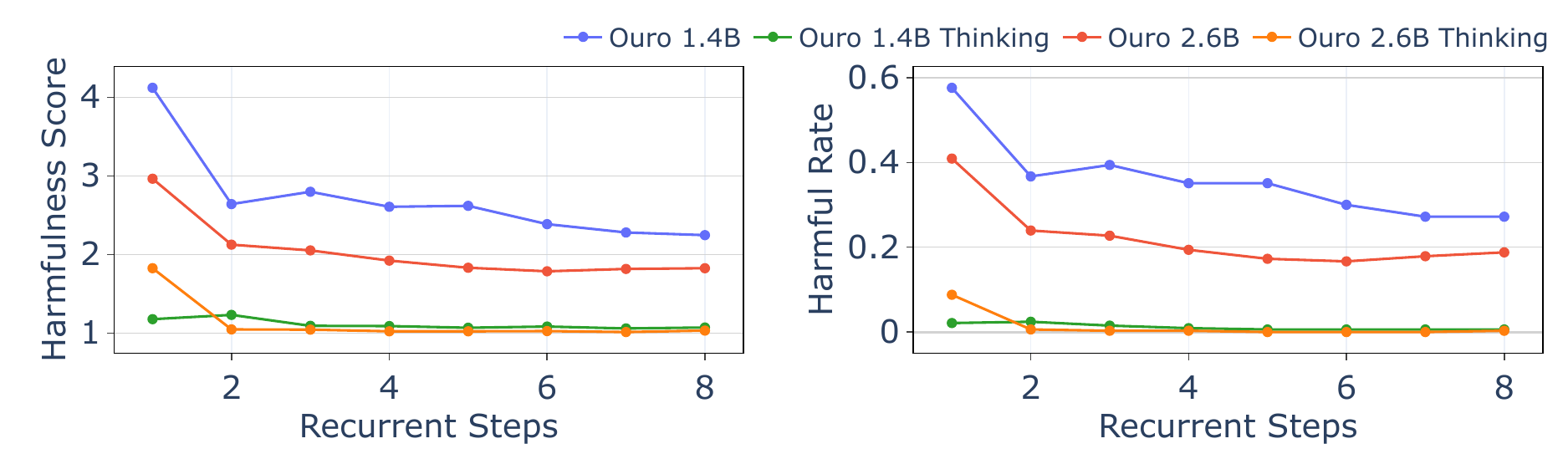}
        \caption{HEx-PHI evaluation}
    \label{fig:ut_safety}
    \end{subfigure}
    \begin{subfigure}{\textwidth}
    \centering
    \includegraphics[width=\linewidth]{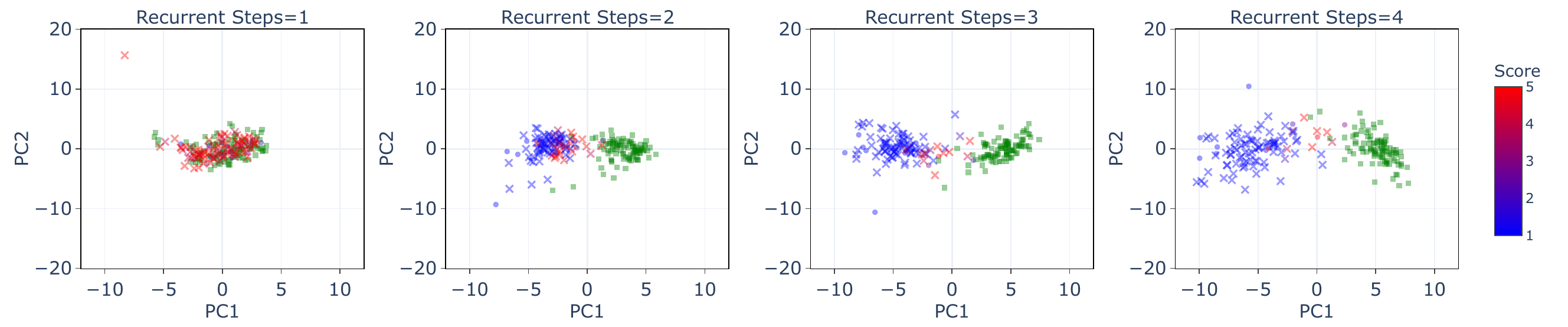}
    \caption{PCA analysis on Ouro 1.4B}
    \label{fig:safety_pca_1.4B}
    \end{subfigure}
    \caption{\textbf{(a) For both 1.4B and 2.6B models, Ouro demonstrates improved safety alignment on HEx-PHI as the recurrent steps increase.} Note that models were trained with 4 recurrent steps; evaluations at steps 5-8 demonstrate successful extrapolation beyond the training configuration. \textbf{(b) As the recurrent steps increase, Ouro 1.4B can better distinguish the benign prompts and harmful prompts, leading to safer responses.} We perform PCA on the hidden representation of the last input token from the model’s top layer. Harmful prompts with a harmfulness score of 4 or 5 at recurrent step 1 are marked with $\times$, while other harmful prompts are shown as circles. The color of each point reflects the harmfulness score of the corresponding response. Benign prompts are shown as green squares.}
    \label{fig:pca_hex_phi}
    
\end{figure}

We assess model safety using HEx-PHI dataset~\citep{qifine}, which contains 330 examples covering 11 prohibited categories. HEx-PHI employs GPT-4o as a judge to assign each model response a harmfulness score from 1 to 5; a higher score indicates a less safe output. Additionally, we compute the harmfulness rate, defined as the proportion of the test cases that receive the highest harmfulness score of 5.  For Ouro Base models, we use greedy decoding with max\_new\_tokens=128; For Ouro Thinking models, we sample with temperature=1.0, top\_p=0.7 with max\_new\_tokens=8192.
We evaluate Ouro 1.4B and 2.6B models with recurrent steps ranging from 1 to 8, and report the result in \Cref{fig:ut_safety}. Notably, while our models were trained with only 4 recurrent steps, both models show their \textbf{extrapolation capability} by extending recurrence steps to 5-8 during inference. This demonstrates the model's ability to generalize to deeper computation than seen during training. The Ouro Thinking checkpoints further enhance safety alignment, reducing harmful rates to 0.009 for Ouro 1.4B Thinking and 0.003 for Ouro 2.6B Thinking at 4 recurrent steps, comparable to Qwen3-4B-Thinking (0.009).

To further investigate how increasing recurrent steps affects the model’s safety alignment, we conduct Principal Component Analysis (PCA) on the hidden representation of the last input token from the top model layer. For a controlled analysis, we select 100 benign and 100 harmful questions with identical formats (all the examples are the questions starting with ``How to'') from Zheng et al.(2024)~\citep{zheng2024prompt}\footnote{Harmful questions: \url{https://github.com/chujiezheng/LLM-Safeguard/blob/main/code/data/custom.txt}; Benign questions: \url{https://github.com/chujiezheng/LLM-Safeguard/blob/main/code/data_harmless/custom.txt}}. Additionally, we evaluate the model’s responses to the 100 harmful questions and compute a 5-level harmfulness score (same as the one used in HEx-PHI) for each response. We plot our PCA analysis on Ouro 1.4B in \Cref{fig:safety_pca_1.4B}, from which we have the following observations. First, as the number of recurrent steps increases, the model becomes more capable of separating benign and harmful prompts, resulting in safer responses, as indicated by the decreasing number of red points. Furthermore, most points associated with unsafe responses appear near the middle of the plot, which represents the boundary between the ``benign'' and ``harmful'' clusters. This suggests that difficulty in distinguishing harmfulness may lead to unsafe responses, which can be alleviated by increasing the number of recurrent steps.

\begin{figure}[h]
    \centering
    \includegraphics[width=0.9\linewidth]{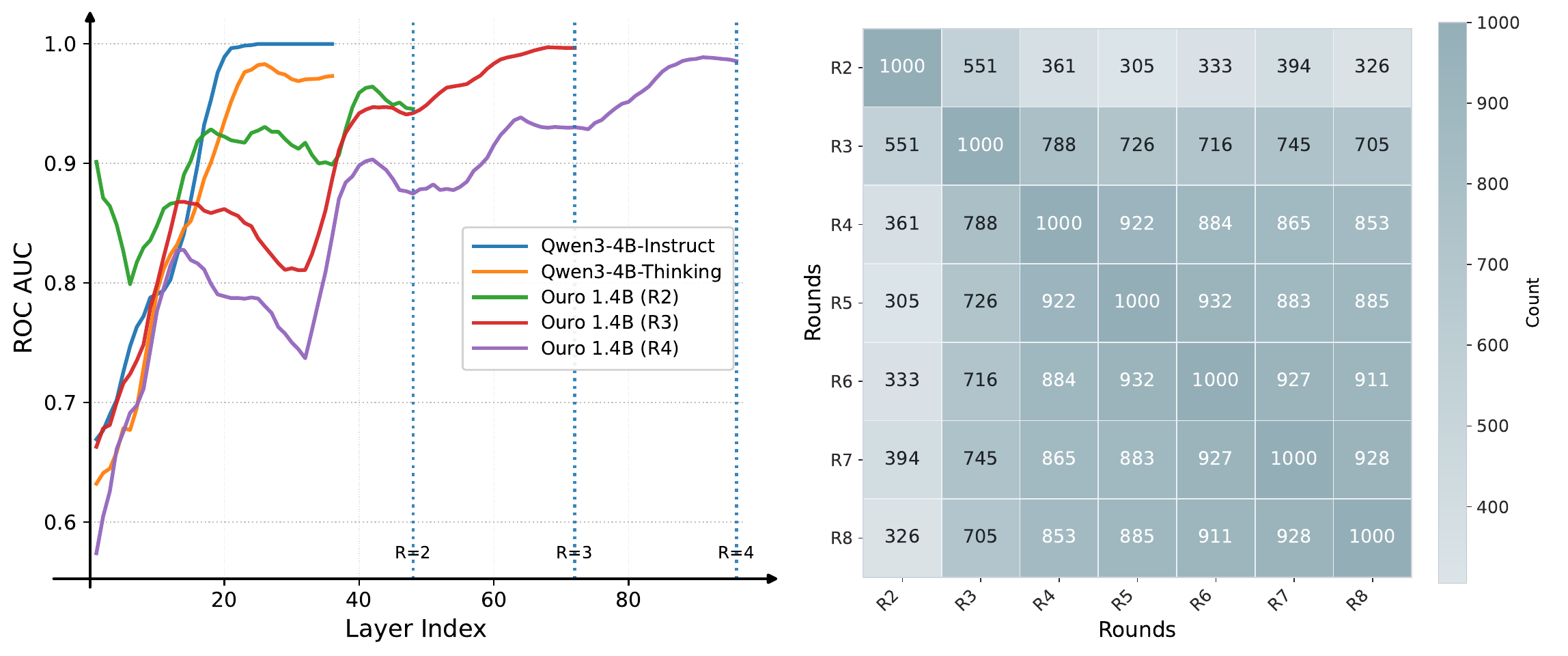}
    \caption{\textbf{Left.} \textbf{ROCAUC of linear probes by layer on Quora Question Pairs.} Each colored curve shows a probe trained on hidden states within a given 2 to 8 recurrent steps to predict that loop’s answer; Qwen3-4B models are the baselines. Vertical dotted lines mark loop boundaries. In recurrent step $i=2,3,4$, the ROC AUC rises quickly within a recurrent step, then partially resets at the next loop, indicating that intra-step answers are determined early while cross-step updates modify the provisional answer. \textbf{Right.} \textbf{Agreement across recurrent steps.} Heat map (A) over 1,000 Quora Question Pairs. Entry $A[i,j]$ is the number of items for which steps (i) and (j) assign the same label.}
    \label{fig:linear_probing}
\end{figure}

\subsection{Faithfulness}
We call a model's thinking process faithful if it is (i) procedurally correct and (ii) causally coupled to the final answer. Concretely, a faithful process should satisfy a counterfactual criterion: if the justification is intervened on (e.g., altered to a different intermediate state), the final prediction should change accordingly. A growing body of work~\citep{cox_post_hoc_cot_2024,arcuschin2025chain0of0thought,barez-chain-2025,korbak2025chain} shows that standard LLMs often appear to decide on an answer before generating chain-of-thought text and then use that text to rationalize the already-formed decision.

In \ut{}, the reasoning substrate is the sequence of latent states 
$h^{(1)} \!\rightarrow\! h^{(2)} \!\rightarrow\! \cdots \!\rightarrow\! h^{(T)}$. 
Each transition $h^{(k)} \!\rightarrow\! h^{(k+1)}$ performs non-trivial computation using the same shared-weight block, and each step is trained to improve the task objective. 
Thus, the causal path to the answer is this latent trajectory, not any optional natural-language trace. 
When we decode intermediate text $\text{Text}(R_k)$ from $h^{(k)}$ via the LM head, we treat it as an instrumented readout of the internal state rather than the mechanism itself. 
Because $h^{(k)}$ is directly supervised by the LM loss, its projection into token space provides a faithful snapshot of what the model currently represents.

Standard evaluation of faithfulness is often based on the manipulation of the reasoning process, CoT, and check if the average treatment effect of CoT is significant. In our case, we cannot manipulate the latent reasoning process. Instead, we adopt an observational proxy for mediation: we read out intermediate hidden representations and test whether predictions change as recurrence deepens on inputs that admit multiple plausible labels. Concretely, we assess whether intermediate ``thinking'' genuinely mediates decisions by measuring step-by-step predictability and agreement patterns. We use the Quora Question Pairs dataset~\citep{quora_question_pairs_kaggle}, which asks whether two short questions are semantically equivalent: a setting with ambiguity and weakly-defined decision boundaries. There are a lot of ambiguous questions in this dataset:
\begin{theobox*}{Ambiguous questions in Quora dataset}
\small
\vspace{0.05cm}
\textbf{Question: does the following two questions have the same intent?}
\footnotesize{
\vspace{-0.23cm}
    \begin{multicols}{2}
        \colorbox{seedblue!20}{\textbf{Pair 1:}}
        \begin{enumerate}[leftmargin=12pt]
          \item What are the questions should not ask on Quora?
          \item Which question should I ask on Quora?
        \end{enumerate}
        \colorbox{red!12}{\textbf{Answer: False}}
        \columnbreak
        
        \hspace{-10pt}
        \colorbox{seedblue!20}{\textbf{Pair 2:}}
        \begin{enumerate}[leftmargin=5pt]
          \item How do we prepare for Union Public Service Commission?
          \item How do I prepare for civil service?
        \end{enumerate}
        \hspace{-10pt}
        \colorbox{blue!12}{\textbf{Answer: True}}
    \end{multicols}}
\end{theobox*}

If a thinking process merely rationalizes the pre-committed answer, even if the questions are very ambiguous, the answers will not change after the reasoning process. This has been reported for Gemma-2 9B and reproduced by us on Qwen-3-4B-Thinking. As shown in the left part of Figure \ref{fig:linear_probing}, the simple linear probe on the final-token logits on the Qwen3-4B-Thinking model shows 0.99 ROC AUC predicting the model's eventual answer, which means the thinking process almost does not affect the results.

In our model, the situation is very different. our $1.4\mathrm{B}\!\times\!4$ model uses 24 layers per recurrent step. 
We train linear probes on hidden states from layers $1$ through $24i$ to predict the step-$i$ answer, for $i\!\in\!\{2,3,4\}$. Within a single recurrent step, the step-$i$ answer is well predicted by a probe on representation within layer $24i$, indicating strong intra-step alignment between state and decision, which is similar to the non-reasoning model Qwen-4B-Instruct, showing in left part of Figure \ref{fig:linear_probing}.
Crucially, probes on the preceding representation (layer $24(i\!-\!1)$) do not reliably predict the step-$i$ decision for $i\!\in\!\{2,3,4\}$, showing that the new recurrent pass performs additional computation that can revise a provisional choice.

To further examine the consistency between the results of different rounds. We also compute a step-by-step agreement matrix $A$ over 1{,}000 Quora Question Pairs, where $A[i,j]$ counts identical labels between step $i$ and step $j$ (diagonal $=\!1000$ by construction). See the right side of Figure \ref{fig:linear_probing}.
Adjacent steps never reach full agreement; for example, $A[2,4]\!=\!361$ indicates only $36.1\%$ of step-2 answers match step-4. $A[2,3]\!=\!551$ indicates only $55
.1\%$ of step-2 answers match step-3. We also notice that when $i\geq4$, the overlap consistency between step-$i$ and step-$i+1$, $A[i,i+1]$, is close to 1000. We think this phenomenon comes from: (1) the model does not learn to reason recursively when $i>4$. The model is trained within 4 loops; (2) as the number of loops increases, the answer gradually converges to a fixed point.

All in all, this systematic disagreement across steps when $i\leq4$ is precisely what a faithful latent process should exhibit: the model is updating its decision as recurrence deepens, and intermediate predictions are not frozen rationalizations of the final output.

\subsection{More Discussion}

The practical barrier for safety-critical deployment is that a model’s articulated reasoning and its final answer may diverge. The \ut{} architecture reduces this gap by exposing a sequence of intermediate predictors that are strongly aligned with the final predictor and can be used both for acceleration and for pre-emptive control. We summarize three deployment advantages.

\paragraph{Built-in draft model for speculative decoding.}
Let $Text(R_t)$ denote the language-model head attached to the latent state after recurrent step $t$, and let $T$ be the maximum step used at deployment. The pair
\[
\bigl(\underbrace{\text{Text}(R_s)}_{\text{proposal}},\;
\underbrace{\text{Text}(R_T)}_{\text{verifier}}\bigr),\qquad 1\le s<T.
\]
forms a native proposal–verification decomposition for speculative decoding without training an external draft model. Proposals are sampled from $Text(R_s)$ and verified under $Text(R_T)$ using standard acceptance tests; rejected tokens are rolled back as usual. Because both heads share the same parameters up to step $s$, cached activations and KV states can be reused, reducing verifier overhead. This turns the recurrent structure into an architectural primitive for draft–verify decoding rather than an add-on.

\paragraph{Joint acceleration and pre-emptive safety.}
Using the same proposal–verification split, safety checks can be interleaved with speculative decoding without extra models. At step $s$:
\begin{enumerate}[leftmargin=1.2em,itemsep=0pt,topsep=2pt]
\item Generate draft tokens with $\text{Text}(R_t)$ and compute their acceptance under $\text{Text}(R_T)$.
\item Run safety screening on the draft distribution or sampled drafts before any token is surfaced to the user. Screening can operate on logits, beams, or short candidate spans.
\item If a violation is detected, halt or reroute the response before streaming; otherwise, accept tokens that pass both verification and safety checks.
\end{enumerate}
Because $\text{Text}(R_s)$ and $\text{Text}(R_T)$ share the latent trajectory, intermediate predictions are well-aligned with the final answer distribution. This alignment makes the step-$s$ output a reliable proxy for the step-$T$ output for the purpose of early screening, while the verifier maintains final quality. The Q-exit threshold $q$ further provides a single deployment knob that simultaneously adjusts compute, consistency, and safety strictness by shifting the average exit depth.

\paragraph{Anytime prediction with adaptive depth.}
\ut{} exposes a predictive distribution at every recurrent step, allowing inference to select an intermediate depth when latency constraints apply. The Stage I objective in Section~\ref{sec:entropy} trains these predictions jointly, while Stage II in Section~\ref{ut:adaptive_exit} trains the exit gate with the language model frozen. Neither objective guarantees that next-token prediction loss decreases at every additional loop; the benefit of deeper computation must be evaluated for the chosen task and depth. This supports a choice of computation budget before emitting each token. Speculative generation can use intermediate predictions as drafts, but tokens should be verified before they are committed to the output.

\section{Conclusion}

In this work, we introduced Ouro, a family of Looped Language Models that demonstrate exceptional parameter efficiency by integrating iterative computation and adaptive depth directly into pre-training on 7.7T tokens. Our 1.4B and 2.6B models consistently match or exceed the performance of 4B and 8B standard transformers, showcasing a 2-3$\times$ efficiency gain. We demonstrated this advantage stems not from increased knowledge storage, but from a fundamentally superior capability for knowledge manipulation, supported by synthetic experiments and theoretical analysis. We also presented a practical training objective using entropy regularization with a uniform prior to learn adaptive depth, and validated efficient KV cache sharing strategies that make \ut{}s viable for real-world deployment.

Beyond performance, the \ut{} architecture exhibits unique properties: its iterative refinement process provides a causally faithful reasoning trace, mitigating the post-hoc rationalization issues seen in standard CoT, and its safety alignment uniquely improves with increased recurrent steps, even when extrapolating. This work establishes iterative latent computation as a critical third scaling axis beyond parameters and data. Future research should focus on enhancing performance extrapolation at greater depths and exploring more complex recurrent mechanisms, solidifying this parameter-efficient approach as a necessary direction in a data-constrained era.

\section*{Acknowledgement}
We sincerely thank Zeyuan Allen-Zhu for his in-depth discussion on the physics of language model part and his enlightening insights on knowledge manipulation. We thank Yuekun Yao for providing valuable insights and discussions on the multi-hop QA task. We also thank Yonghui Wu, Guang Shi, Shu Zhong, Tenglong Ao, Chen Chen, Songlin Yang, Wenhao Chai, and Yuhong Chou for their insightful discussions. Special thanks to Wenjia Zhu -- his words opened our eyes to what the real problems are in current models, and inspired us to explore this direction.
\newpage

\section*{Contributions}

\textbf{Project Lead}

Rui-Jie Zhu, Zixuan Wang, Kai Hua, Ge Zhang

\textbf{Core Contributors}

Rui-Jie Zhu: Proposes the project and leads the pre-training of Ouro. Optimizes pre-training and inference infrastructure, develops the initial vLLM implementation, and explores RLVR.

Zixuan Wang: Leads the analysis on understanding \ut{} superiority and is responsible for related experiments. He contributes to the design of adaptive early exit strategies, training, and the safety analysis. 

Kai Hua: Designs and curates all pre-training data mixtures and provides key insights during the pre-training process.

Ge Zhang: Co-leads and supervises the Ouro. Provides several key insights during the pre-training and post-training process.

Tianyu Zhang: Leads the analysis of Ouro on consistency, safety, and faithfulness. He designs the pipeline evaluation on faithfulness. He contributes to post-training, probing and efficient KV cache design.

Ziniu Li: Leads the post-training phase, developing supervised fine-tuning and providing key contributions to RLVR exploration.

Haoran Que: Leads the scaling law analysis for \ut{}, investigating the relationship between performance, model size, and recurrent depth.

Boyi Wei: Contributes to the safety analysis, conducting evaluations on the HEx-PHI benchmark and performing PCA on model representations.

Zixin Wen: Contributes to the theoretical analysis, the design of adaptive exit strategies, Physics of LLMs experiments, paper writing, and RLVR.

Fan Yin: Optimizes the vLLM and SGLang implementations for Ouro, contributing core pull requests to improve inference efficiency.

He Xing: Contributes to the vLLM infrastructure development and optimization.

\textbf{Contributors}

Lu Li, Jiajun Shi, Kaijing Ma, Shanda Li, Taylor Kergan, Andrew Smith, Xingwei Qu, Mude Hui, Bohong Wu, Xun Zhou, Qiyang Min, Hongzhi Huang,  Wei Ye, Jiaheng Liu, Jian Yang, Yunfeng Shi, Chenghua Lin, Enduo Zhao, Tianle Cai

\textbf{Supervision}

Ge Zhang, Wenhao Huang, Yoshua Bengio, Jason Eshraghian

\bibliographystyle{unsrt}
\bibliography{main}

\begin{thebibliography}{10}

\bibitem{brown2020language}
Tom Brown, Benjamin Mann, Nick Ryder, Melanie Subbiah, Jared~D Kaplan, Prafulla Dhariwal, Arvind Neelakantan, Pranav Shyam, Girish Sastry, Amanda Askell, et~al.
\newblock Language models are few-shot learners.
\newblock {\em Advances in neural information processing systems}, 33:1877--1901, 2020.

\bibitem{qwen2}
Qwen Team et~al.
\newblock Qwen2 technical report.
\newblock {\em arXiv preprint arXiv:2407.10671}, 2:3, 2024.

\bibitem{qwen3}
An~Yang, Anfeng Li, Baosong Yang, Beichen Zhang, Binyuan Hui, Bo~Zheng, Bowen Yu, Chang Gao, Chengen Huang, Chenxu Lv, et~al.
\newblock Qwen3 technical report.
\newblock {\em arXiv preprint arXiv:2505.09388}, 2025.

\bibitem{team2025gemma3}
Gemma Team, Aishwarya Kamath, Johan Ferret, Shreya Pathak, Nino Vieillard, Ramona Merhej, Sarah Perrin, Tatiana Matejovicova, Alexandre Ram{\'e}, Morgane Rivi{\`e}re, et~al.
\newblock Gemma 3 technical report.
\newblock {\em arXiv preprint arXiv:2503.19786}, 2025.

\bibitem{dubey2024llama}
Abhimanyu Dubey, Abhinav Jauhri, Abhinav Pandey, Abhishek Kadian, Ahmad Al-Dahle, Aiesha Letman, Akhil Mathur, Alan Schelten, Amy Yang, Angela Fan, et~al.
\newblock The llama 3 herd of models.
\newblock {\em arXiv e-prints}, pages arXiv--2407, 2024.

\bibitem{wei2022chain}
Jason Wei, Xuezhi Wang, Dale Schuurmans, Maarten Bosma, Fei Xia, Ed~Chi, Quoc~V Le, Denny Zhou, et~al.
\newblock Chain-of-thought prompting elicits reasoning in large language models.
\newblock {\em Advances in neural information processing systems}, 35:24824--24837, 2022.

\bibitem{saunshi2025reasoning}
Nikunj Saunshi, Nishanth Dikkala, Zhiyuan Li, Sanjiv Kumar, and Sashank~J Reddi.
\newblock Reasoning with latent thoughts: On the power of looped transformers.
\newblock {\em arXiv preprint arXiv:2502.17416}, 2025.

\bibitem{gatmiry2024can}
Khashayar Gatmiry, Nikunj Saunshi, Sashank~J Reddi, Stefanie Jegelka, and Sanjiv Kumar.
\newblock Can looped transformers learn to implement multi-step gradient descent for in-context learning?
\newblock {\em arXiv preprint arXiv:2410.08292}, 2024.

\bibitem{gatmiry2024role}
Khashayar Gatmiry, Nikunj Saunshi, Sashank~J Reddi, Stefanie Jegelka, and Sanjiv Kumar.
\newblock On the role of depth and looping for in-context learning with task diversity.
\newblock {\em arXiv preprint arXiv:2410.21698}, 2024.

\bibitem{huang2025transformers}
Jianhao Huang, Zixuan Wang, and Jason~D Lee.
\newblock Transformers learn to implement multi-step gradient descent with chain of thought.
\newblock {\em arXiv preprint arXiv:2502.21212}, 2025.

\bibitem{merrill2025little}
William Merrill and Ashish Sabharwal.
\newblock A little depth goes a long way: The expressive power of log-depth transformers.
\newblock {\em arXiv preprint arXiv:2503.03961}, 2025.

\bibitem{merrill2025exact}
William Merrill and Ashish Sabharwal.
\newblock Exact expressive power of transformers with padding.
\newblock {\em arXiv preprint arXiv:2505.18948}, 2025.

\bibitem{giannou2023looped}
Angeliki Giannou, Shashank Rajput, Jy-yong Sohn, Kangwook Lee, Jason~D Lee, and Dimitris Papailiopoulos.
\newblock Looped transformers as programmable computers.
\newblock In {\em International Conference on Machine Learning}, pages 11398--11442. PMLR, 2023.

\bibitem{yang2023looped}
Liu Yang, Kangwook Lee, Robert Nowak, and Dimitris Papailiopoulos.
\newblock Looped transformers are better at learning learning algorithms.
\newblock {\em arXiv preprint arXiv:2311.12424}, 2023.

\bibitem{dehghani2018universal}
Mostafa Dehghani, Stephan Gouws, Oriol Vinyals, Jakob Uszkoreit, and {\L}ukasz Kaiser.
\newblock Universal transformers.
\newblock {\em arXiv preprint arXiv:1807.03819}, 2018.

\bibitem{bae2024relaxed}
Sangmin Bae, Adam Fisch, Hrayr Harutyunyan, Ziwei Ji, Seungyeon Kim, and Tal Schuster.
\newblock Relaxed recursive transformers: Effective parameter sharing with layer-wise lora.
\newblock {\em arXiv preprint arXiv:2410.20672}, 2024.

\bibitem{geiping2025scaling}
Jonas Geiping, Sean McLeish, Neel Jain, John Kirchenbauer, Siddharth Singh, Brian~R Bartoldson, Bhavya Kailkhura, Abhinav Bhatele, and Tom Goldstein.
\newblock Scaling up test-time compute with latent reasoning: A recurrent depth approach.
\newblock {\em arXiv preprint arXiv:2502.05171}, 2025.

\bibitem{zeng2025pretraining}
Boyi Zeng, Shixiang Song, Siyuan Huang, Yixuan Wang, He~Li, Ziwei He, Xinbing Wang, Zhiyu Li, and Zhouhan Lin.
\newblock Pretraining language models to ponder in continuous space.
\newblock {\em arXiv preprint arXiv:2505.20674}, 2025.

\bibitem{zhu2025survey}
Rui-Jie Zhu, Tianhao Peng, Tianhao Cheng, Xingwei Qu, Jinfa Huang, Dawei Zhu, Hao Wang, Kaiwen Xue, Xuanliang Zhang, Yong Shan, et~al.
\newblock A survey on latent reasoning.
\newblock {\em arXiv preprint arXiv:2507.06203}, 2025.

\bibitem{qifine}
Xiangyu Qi, Yi~Zeng, Tinghao Xie, Pin-Yu Chen, Ruoxi Jia, Prateek Mittal, and Peter Henderson.
\newblock Fine-tuning aligned language models compromises safety, even when users do not intend to!
\newblock In {\em The Twelfth International Conference on Learning Representations}.

\bibitem{chen2025inner}
Yilong Chen, Junyuan Shang, Zhenyu Zhang, Yanxi Xie, Jiawei Sheng, Tingwen Liu, Shuohuan Wang, Yu~Sun, Hua Wu, and Haifeng Wang.
\newblock Inner thinking transformer: Leveraging dynamic depth scaling to foster adaptive internal thinking.
\newblock {\em arXiv preprint arXiv:2502.13842}, 2025.

\bibitem{bae2025mixture}
Sangmin Bae, Yujin Kim, Reza Bayat, Sungnyun Kim, Jiyoun Ha, Tal Schuster, Adam Fisch, Hrayr Harutyunyan, Ziwei Ji, Aaron Courville, et~al.
\newblock Mixture-of-recursions: Learning dynamic recursive depths for adaptive token-level computation.
\newblock {\em arXiv preprint arXiv:2507.10524}, 2025.

\bibitem{lan2019albert}
Zhenzhong Lan, Mingda Chen, Sebastian Goodman, Kevin Gimpel, Piyush Sharma, and Radu Soricut.
\newblock Albert: A lite bert for self-supervised learning of language representations.
\newblock {\em arXiv preprint arXiv:1909.11942}, 2019.

\bibitem{dabre2019recurrent}
Raj Dabre and Atsushi Fujita.
\newblock Recurrent stacking of layers for compact neural machine translation models.
\newblock In {\em Proceedings of the AAAI Conference on Artificial Intelligence}, volume~33, pages 6292--6299, 2019.

\bibitem{takase2021lessons}
Sho Takase and Shun Kiyono.
\newblock Lessons on parameter sharing across layers in transformers.
\newblock {\em arXiv preprint arXiv:2104.06022}, 2021.

\bibitem{li2025megrez2}
Boxun Li, Yadong Li, Zhiyuan Li, Congyi Liu, Weilin Liu, Guowei Niu, Zheyue Tan, Haiyang Xu, Zhuyu Yao, Tao Yuan, et~al.
\newblock Megrez2 technical report.
\newblock {\em arXiv preprint arXiv:2507.17728}, 2025.

\bibitem{hao2024training}
Shibo Hao, Sainbayar Sukhbaatar, DiJia Su, Xian Li, Zhiting Hu, Jason Weston, and Yuandong Tian.
\newblock Training large language models to reason in a continuous latent space.
\newblock {\em arXiv preprint arXiv:2412.06769}, 2024.

\bibitem{mohtashami2023cotformer}
Amirkeivan Mohtashami, Matteo Pagliardini, and Martin Jaggi.
\newblock Cotformer: More tokens with attention make up for less depth.
\newblock In {\em Workshop on Advancing Neural Network Training: Computational Efficiency, Scalability, and Resource Optimization (WANT@ NeurIPS 2023)}, 2023.

\bibitem{wu2025efficient}
Bohong Wu, Shen Yan, Sijun Zhang, Jianqiao Lu, Yutao Zeng, Ya~Wang, and Xun Zhou.
\newblock Efficient pretraining length scaling.
\newblock {\em arXiv preprint arXiv:2504.14992}, 2025.

\bibitem{ye2024physics}
Tian Ye, Zicheng Xu, Yuanzhi Li, and Zeyuan Allen-Zhu.
\newblock Physics of language models: Part 2.1, grade-school math and the hidden reasoning process.
\newblock {\em arXiv preprint arXiv:2407.20311}, 2024.

\bibitem{wang2025beyond}
Shenzhi Wang, Le~Yu, Chang Gao, Chujie Zheng, Shixuan Liu, Rui Lu, Kai Dang, Xionghui Chen, Jianxin Yang, Zhenru Zhang, et~al.
\newblock Beyond the 80/20 rule: High-entropy minority tokens drive effective reinforcement learning for llm reasoning.
\newblock {\em arXiv preprint arXiv:2506.01939}, 2025.

\bibitem{balagansky2022palbert}
Nikita Balagansky and Daniil Gavrilov.
\newblock {PALBERT}: Teaching {ALBERT} to Ponder.
\newblock In {\em Advances in Neural Information Processing Systems}, volume~35, pages 14002--14012, 2022.

\bibitem{banino2021pondernet}
Andrea Banino, Jan Balaguer, and Charles Blundell.
\newblock Pondernet: Learning to ponder.
\newblock {\em arXiv preprint arXiv:2107.05407}, 2021.

\bibitem{vaswani2017attention}
Ashish Vaswani, Noam Shazeer, Niki Parmar, Jakob Uszkoreit, Llion Jones, Aidan~N Gomez, {\L}ukasz Kaiser, and Illia Polosukhin.
\newblock Attention is all you need.
\newblock {\em Advances in neural information processing systems}, 30, 2017.

\bibitem{su2023roformerenhancedtransformerrotary}
Jianlin Su, Yu~Lu, Shengfeng Pan, Ahmed Murtadha, Bo~Wen, and Yunfeng Liu.
\newblock Roformer: Enhanced transformer with rotary position embedding, 2023.

\bibitem{shazeer2020glu}
Noam Shazeer.
\newblock Glu variants improve transformer.
\newblock {\em arXiv preprint arXiv:2002.05202}, 2020.

\bibitem{allal2025smollm2}
Loubna~Ben Allal, Anton Lozhkov, Elie Bakouch, Gabriel~Mart{\'\i}n Bl{\'a}zquez, Guilherme Penedo, Lewis Tunstall, Andr{\'e}s Marafioti, Hynek Kydl{\'\i}{\v{c}}ek, Agust{\'\i}n~Piqueres Lajar{\'\i}n, Vaibhav Srivastav, et~al.
\newblock Smollm2: When smol goes big--data-centric training of a small language model.
\newblock {\em arXiv preprint arXiv:2502.02737}, 2025.

\bibitem{wen2024understanding}
Kaiyue Wen, Zhiyuan Li, Jason Wang, David Hall, Percy Liang, and Tengyu Ma.
\newblock Understanding warmup-stable-decay learning rates: A river valley loss landscape perspective.
\newblock {\em arXiv preprint arXiv:2410.05192}, 2024.

\bibitem{penedo2024fineweb}
Guilherme Penedo, Hynek Kydl{\'\i}{\v{c}}ek, Anton Lozhkov, Margaret Mitchell, Colin~A Raffel, Leandro Von~Werra, Thomas Wolf, et~al.
\newblock The fineweb datasets: Decanting the web for the finest text data at scale.
\newblock {\em Advances in Neural Information Processing Systems}, 37:30811--30849, 2024.

\bibitem{li2024dclm}
Jeffrey Li, Alex Fang, Georgios Smyrnis, Maor Ivgi, Matt Jordan, Samir~Yitzhak Gadre, Hritik Bansal, Etash Guha, Sedrick~Scott Keh, Kushal Arora, et~al.
\newblock Datacomp-lm: In search of the next generation of training sets for language models.
\newblock {\em Advances in Neural Information Processing Systems}, 37:14200--14282, 2024.

\bibitem{su2024nemotron}
Dan Su, Kezhi Kong, Ying Lin, Joseph Jennings, Brandon Norick, Markus Kliegl, Mostofa Patwary, Mohammad Shoeybi, and Bryan Catanzaro.
\newblock Nemotron-cc: Transforming common crawl into a refined long-horizon pretraining dataset.
\newblock {\em arXiv preprint arXiv:2412.02595}, 2024.

\bibitem{wang2025ultra}
Yudong Wang, Zixuan Fu, Jie Cai, Peijun Tang, Hongya Lyu, Yewei Fang, Zhi Zheng, Jie Zhou, Guoyang Zeng, Chaojun Xiao, et~al.
\newblock Ultra-fineweb: Efficient data filtering and verification for high-quality llm training data.
\newblock {\em arXiv preprint arXiv:2505.05427}, 2025.

\bibitem{du2024chinese}
Xinrun Du, Zhouliang Yu, Songyang Gao, Ding Pan, Yuyang Cheng, Ziyang Ma, Ruibin Yuan, Xingwei Qu, Jiaheng Liu, Tianyu Zheng, et~al.
\newblock Chinese tiny llm: Pretraining a chinese-centric large language model.
\newblock {\em arXiv preprint arXiv:2404.04167}, 2024.

\bibitem{Huang2024OpenCoderTO}
Siming Huang, Tianhao Cheng, Jason~Klein Liu, Jiaran Hao, Liuyihan Song, Yang Xu, J.~Yang, J.~H. Liu, Chenchen Zhang, Linzheng Chai, Ruifeng Yuan, Zhaoxiang Zhang, Jie Fu, Qian Liu, Ge~Zhang, Zili Wang, Yuan Qi, Yinghui Xu, and Wei Chu.
\newblock Opencoder: The open cookbook for top-tier code large language models.
\newblock 2024.

\bibitem{zhou2025megamath}
Fan Zhou, Zengzhi Wang, Nikhil Ranjan, Zhoujun Cheng, Liping Tang, Guowei He, Zhengzhong Liu, and Eric~P. Xing.
\newblock Megamath: Pushing the limits of open math corpora.
\newblock {\em arXiv preprint arXiv:2504.02807}, 2025.
\newblock Preprint.

\bibitem{karimi2025nemotroncc}
Rabeeh~Karimi Mahabadi, Sanjeev Satheesh, Shrimai Prabhumoye, Mostofa Patwary, Mohammad Shoeybi, and Bryan Catanzaro.
\newblock Nemotron-cc-math: A 133 billion-token-scale high quality math pretraining dataset.
\newblock 2025.

\bibitem{nvidia2025nvidianemotronnano2}
NVIDIA, Aarti Basant, Abhijit Khairnar, Abhijit Paithankar, Abhinav Khattar, Adithya Renduchintala, Aditya Malte, Akhiad Bercovich, Akshay Hazare, Alejandra Rico, Aleksander Ficek, Alex Kondratenko, Alex Shaposhnikov, Alexander Bukharin, Ali Taghibakhshi, Amelia Barton, Ameya~Sunil Mahabaleshwarkar, Amy Shen, Andrew Tao, Ann Guan, Anna Shors, Anubhav Mandarwal, Arham Mehta, Arun Venkatesan, Ashton Sharabiani, Ashwath Aithal, Ashwin Poojary, Ayush Dattagupta, Balaram Buddharaju, Banghua Zhu, Barnaby Simkin, Bilal Kartal, Bita~Darvish Rouhani, Bobby Chen, Boris Ginsburg, Brandon Norick, Brian Yu, Bryan Catanzaro, Charles Wang, Charlie Truong, Chetan Mungekar, Chintan Patel, Chris Alexiuk, Christian Munley, Christopher Parisien, Dan Su, Daniel Afrimi, Daniel Korzekwa, Daniel Rohrer, Daria Gitman, David Mosallanezhad, Deepak Narayanan, Dima Rekesh, Dina Yared, Dmytro Pykhtar, Dong Ahn, Duncan Riach, Eileen Long, Elliott Ning, Eric Chung, Erick Galinkin, Evelina Bakhturina, Gargi Prasad, Gerald Shen, Haifeng Qian,
  Haim Elisha, Harsh Sharma, Hayley Ross, Helen Ngo, Herman Sahota, Hexin Wang, Hoo~Chang Shin, Hua Huang, Iain Cunningham, Igor Gitman, Ivan Moshkov, Jaehun Jung, Jan Kautz, Jane~Polak Scowcroft, Jared Casper, Jian Zhang, Jiaqi Zeng, Jimmy Zhang, Jinze Xue, Jocelyn Huang, Joey Conway, John Kamalu, Jonathan Cohen, Joseph Jennings, Julien~Veron Vialard, Junkeun Yi, Jupinder Parmar, Kari Briski, Katherine Cheung, Katherine Luna, Keith Wyss, Keshav Santhanam, Kezhi Kong, Krzysztof Pawelec, Kumar Anik, Kunlun Li, Kushan Ahmadian, Lawrence McAfee, Laya Sleiman, Leon Derczynski, Luis Vega, Maer~Rodrigues de~Melo, Makesh~Narsimhan Sreedhar, Marcin Chochowski, Mark Cai, Markus Kliegl, Marta Stepniewska-Dziubinska, Matvei Novikov, Mehrzad Samadi, Meredith Price, Meriem Boubdir, Michael Boone, Michael Evans, Michal Bien, Michal Zawalski, Miguel Martinez, Mike Chrzanowski, Mohammad Shoeybi, Mostofa Patwary, Namit Dhameja, Nave Assaf, Negar Habibi, Nidhi Bhatia, Nikki Pope, Nima Tajbakhsh, Nirmal~Kumar Juluru, Oleg
  Rybakov, Oleksii Hrinchuk, Oleksii Kuchaiev, Oluwatobi Olabiyi, Pablo Ribalta, Padmavathy Subramanian, Parth Chadha, Pavlo Molchanov, Peter Dykas, Peter Jin, Piotr Bialecki, Piotr Januszewski, Pradeep Thalasta, Prashant Gaikwad, Prasoon Varshney, Pritam Gundecha, Przemek Tredak, Rabeeh~Karimi Mahabadi, Rajen Patel, Ran El-Yaniv, Ranjit Rajan, Ria Cheruvu, Rima Shahbazyan, Ritika Borkar, Ritu Gala, Roger Waleffe, Ruoxi Zhang, Russell~J. Hewett, Ryan Prenger, Sahil Jain, Samuel Kriman, Sanjeev Satheesh, Saori Kaji, Sarah Yurick, Saurav Muralidharan, Sean Narenthiran, Seonmyeong Bak, Sepehr Sameni, Seungju Han, Shanmugam Ramasamy, Shaona Ghosh, Sharath~Turuvekere Sreenivas, Shelby Thomas, Shizhe Diao, Shreya Gopal, Shrimai Prabhumoye, Shubham Toshniwal, Shuoyang Ding, Siddharth Singh, Siddhartha Jain, Somshubra Majumdar, Soumye Singhal, Stefania Alborghetti, Syeda~Nahida Akter, Terry Kong, Tim Moon, Tomasz Hliwiak, Tomer Asida, Tony Wang, Tugrul Konuk, Twinkle Vashishth, Tyler Poon, Udi Karpas, Vahid Noroozi,
  Venkat Srinivasan, Vijay Korthikanti, Vikram Fugro, Vineeth Kalluru, Vitaly Kurin, Vitaly Lavrukhin, Wasi~Uddin Ahmad, Wei Du, Wonmin Byeon, Ximing Lu, Xin Dong, Yashaswi Karnati, Yejin Choi, Yian Zhang, Ying Lin, Yonggan Fu, Yoshi Suhara, Zhen Dong, Zhiyu Li, Zhongbo Zhu, and Zijia Chen.
\newblock Nvidia nemotron nano 2: An accurate and efficient hybrid mamba-transformer reasoning model, 2025.

\bibitem{gao2024train}
Tianyu Gao, Alexander Wettig, Howard Yen, and Danqi Chen.
\newblock How to train long-context language models (effectively).
\newblock {\em arXiv preprint arXiv:2410.02660}, 2024.

\bibitem{zhang2025flame}
Yu~Zhang and Songlin Yang.
\newblock Flame: Flash language modeling made easy, January 2025.

\bibitem{liang2025torchtitan}
Wanchao Liang, Tianyu Liu, Less Wright, Will Constable, Andrew Gu, Chien-Chin Huang, Iris Zhang, Wei Feng, Howard Huang, Junjie Wang, Sanket Purandare, Gokul Nadathur, and Stratos Idreos.
\newblock Torchtitan: One-stop pytorch native solution for production ready {LLM} pretraining.
\newblock In {\em The Thirteenth International Conference on Learning Representations}, 2025.

\bibitem{guha2025openthoughts}
Etash Guha, Ryan Marten, Sedrick Keh, Negin Raoof, Georgios Smyrnis, Hritik Bansal, Marianna Nezhurina, Jean Mercat, Trung Vu, Zayne Sprague, et~al.
\newblock Openthoughts: Data recipes for reasoning models.
\newblock {\em arXiv preprint arXiv:2506.04178}, 2025.

\bibitem{liu2025acereason}
Zihan Liu, Zhuolin Yang, Yang Chen, Chankyu Lee, Mohammad Shoeybi, Bryan Catanzaro, and Wei Ping.
\newblock Acereason-nemotron 1.1: Advancing math and code reasoning through sft and rl synergy.
\newblock {\em arXiv preprint arXiv:2506.13284}, 2025.

\bibitem{ahmad2025opencodereasoning}
Wasi~Uddin Ahmad, Sean Narenthiran, Somshubra Majumdar, Aleksander Ficek, Siddhartha Jain, Jocelyn Huang, Vahid Noroozi, and Boris Ginsburg.
\newblock Opencodereasoning: Advancing data distillation for competitive coding.
\newblock {\em arXiv preprint arXiv:2504.01943}, 2025.

\bibitem{bercovich2025llama}
Akhiad Bercovich, Itay Levy, Izik Golan, Mohammad Dabbah, Ran El-Yaniv, Omri Puny, Ido Galil, Zach Moshe, Tomer Ronen, Najeeb Nabwani, et~al.
\newblock Llama-nemotron: Efficient reasoning models.
\newblock {\em arXiv preprint arXiv:2505.00949}, 2025.

\bibitem{wang2025reverse}
Haozhe Wang, Haoran Que, Qixin Xu, Minghao Liu, Wangchunshu Zhou, Jiazhan Feng, Wanjun Zhong, Wei Ye, Tong Yang, Wenhao Huang, et~al.
\newblock Reverse-engineered reasoning for open-ended generation.
\newblock {\em arXiv preprint arXiv:2509.06160}, 2025.

\bibitem{zheng2024llamafactory}
Yaowei Zheng, Richong Zhang, Junhao Zhang, Yanhan Ye, Zheyan Luo, Zhangchi Feng, and Yongqiang Ma.
\newblock Llamafactory: Unified efficient fine-tuning of 100+ language models.
\newblock {\em arXiv preprint arXiv:2403.13372}, 2024.

\bibitem{yu2025dapo}
Qiying Yu, Zheng Zhang, Ruofei Zhu, Yufeng Yuan, Xiaochen Zuo, Yu~Yue, Weinan Dai, Tiantian Fan, Gaohong Liu, Lingjun Liu, et~al.
\newblock Dapo: An open-source llm reinforcement learning system at scale.
\newblock {\em arXiv preprint arXiv:2503.14476}, 2025.

\bibitem{shao2024deepseekmath}
Zhihong Shao, Peiyi Wang, Qihao Zhu, Runxin Xu, Junxiao Song, Xiao Bi, Haowei Zhang, Mingchuan Zhang, YK~Li, Yang Wu, et~al.
\newblock Deepseekmath: Pushing the limits of mathematical reasoning in open language models.
\newblock {\em arXiv preprint arXiv:2402.03300}, 2024.

\bibitem{eval-harness}
Leo Gao, Jonathan Tow, Baber Abbasi, Stella Biderman, Sid Black, Anthony DiPofi, Charles Foster, Laurence Golding, Jeffrey Hsu, Alain Le~Noac'h, Haonan Li, Kyle McDonell, Niklas Muennighoff, Chris Ociepa, Jason Phang, Laria Reynolds, Hailey Schoelkopf, Aviya Skowron, Lintang Sutawika, Eric Tang, Anish Thite, Ben Wang, Kevin Wang, and Andy Zou.
\newblock The language model evaluation harness, 07 2024.

\bibitem{evalplus}
Jiawei Liu, Chunqiu~Steven Xia, Yuyao Wang, and Lingming Zhang.
\newblock Is your code generated by chat{GPT} really correct? rigorous evaluation of large language models for code generation.
\newblock In {\em Thirty-seventh Conference on Neural Information Processing Systems}, 2023.

\bibitem{HuggingFaceH4_2024_AIME2024}
{HuggingFaceH4}.
\newblock Aime 2024.
\newblock \url{https://huggingface.co/datasets/HuggingFaceH4/aime_2024}, 2024.
\newblock 30 problems from AIME I \& II 2024.

\bibitem{He2024OlympiadBench}
Chaoqun He, Renjie Luo, Yuzhuo Bai, et~al.
\newblock Olympiadbench: A challenging benchmark for promoting agi with olympiad-level bilingual multimodal scientific problems.
\newblock {\em arXiv preprint arXiv:2402.14008}, 2024.

\bibitem{rein2023gpqa}
David Rein, Betty~Li Hou, Asa Cooper~Stickland, Jackson Petty, Richard~Yuanzhe Pang, Julien Dirani, Julian Michael, and Samuel~R. Bowman.
\newblock {GPQA}: A graduate-level google-proof q\&a benchmark.
\newblock {\em arXiv preprint arXiv:2311.12022}, 2023.

\bibitem{MAPTeam2025SuperGPQA}
{M-A-P Team}, Xinrun Du, Yifan Yao, et~al.
\newblock Supergpqa: Scaling llm evaluation across 285 graduate disciplines.
\newblock {\em arXiv preprint arXiv:2502.14739}, 2025.

\bibitem{ByteDanceSeed_2025_BeyondAIME}
{ByteDance-Seed}.
\newblock Beyondaime.
\newblock \url{https://huggingface.co/datasets/ByteDance-Seed/BeyondAIME}, 2025.
\newblock CC0-1.0 license.

\bibitem{Phan2025HLE}
Long Phan, Alice Gatti, Ziwen Han, Nathaniel Li, et~al.
\newblock Humanity's last exam.
\newblock {\em arXiv preprint arXiv:2501.14249}, 2025.

\bibitem{talmor-etal-2019-commonsenseqa}
Alon Talmor, Jonathan Herzig, Nicholas Lourie, and Jonathan Berant.
\newblock {C}ommonsense{QA}: A question answering challenge targeting commonsense knowledge.
\newblock In {\em Proceedings of the 2019 Conference of the North {A}merican Chapter of the Association for Computational Linguistics: Human Language Technologies, Volume 1 (Long and Short Papers)}, pages 4149--4158, Minneapolis, Minnesota, June 2019. Association for Computational Linguistics.

\bibitem{AL2024-knowledge3}
Zeyuan {Allen-Zhu} and Yuanzhi Li.
\newblock {Physics of Language Models: Part 3.3, Knowledge Capacity Scaling Laws}.
\newblock In {\em Proceedings of the 13th International Conference on Learning Representations}, ICLR~'25, April 2025.
\newblock Full version available at \url{https://ssrn.com/abstract=5250617}.

\bibitem{Allenzhu2025-canon}
Zeyuan {Allen-Zhu}.
\newblock {Physics of Language Models: Part 4.1, Architecture Design and the Magic of Canon Layers}.
\newblock {\em SSRN Electronic Journal}, May 2025.
\newblock \url{https://ssrn.com/abstract=5240330}.

\bibitem{yao2025language}
Yuekun Yao, Yupei Du, Dawei Zhu, Michael Hahn, and Alexander Koller.
\newblock Language models can learn implicit multi-hop reasoning, but only if they have lots of training data.
\newblock {\em arXiv preprint arXiv:2505.17923}, 2025.

\bibitem{zhu2025reasoning}
Hanlin Zhu, Shibo Hao, Zhiting Hu, Jiantao Jiao, Stuart Russell, and Yuandong Tian.
\newblock Reasoning by superposition: A theoretical perspective on chain of continuous thought.
\newblock {\em arXiv preprint arXiv:2505.12514}, 2025.

\bibitem{zhong2025understanding}
Shu Zhong, Mingyu Xu, Tenglong Ao, and Guang Shi.
\newblock Understanding transformer from the perspective of associative memory.
\newblock {\em arXiv preprint arXiv:2505.19488}, 2025.

\bibitem{zheng2024prompt}
Chujie Zheng, Fan Yin, Hao Zhou, Fandong Meng, Jie Zhou, Kai-Wei Chang, Minlie Huang, and Nanyun Peng.
\newblock On prompt-driven safeguarding for large language models.
\newblock In {\em Proceedings of the 41st International Conference on Machine Learning}, pages 61593--61613, 2024.

\bibitem{cox_post_hoc_cot_2024}
Kyle Cox.
\newblock Post-hoc reasoning in chain of thought, December 2024.
\newblock Blog post.

\bibitem{arcuschin2025chain0of0thought}
Iván Arcuschin, Jett Janiak, Robert Krzyzanowski, Senthooran Rajamanoharan, Neel Nanda, and Arthur Conmy.
\newblock Chain-of-thought reasoning in the wild is not always faithful.
\newblock {\em arXiv preprint arXiv: 2503.08679}, 2025.

\bibitem{barez-chain-2025}
Fazl Barez, Tung-Yu Wu, Iván Arcuschin, Michael Lan, Vincent Wang, Noah Siegel, Nicolas Collignon, Clement Neo, Isabelle Lee, Alasdair Paren, Adel Bibi, Robert Trager, Damiano Fornasiere, John Yan, Yanai Elazar, and Yoshua Bengio.
\newblock Chain-of-thought is not explainability.
\newblock 2025.

\bibitem{korbak2025chain}
Tomek Korbak, Mikita Balesni, Elizabeth Barnes, Yoshua Bengio, Joe Benton, Joseph Bloom, Mark Chen, Alan Cooney, Allan Dafoe, Anca Dragan, Scott Emmons, Owain Evans, David Farhi, Ryan Greenblatt, Dan Hendrycks, Marius Hobbhahn, Evan Hubinger, Geoffrey Irving, Erik Jenner, Daniel Kokotajlo, Victoria Krakovna, Shane Legg, David Lindner, David Luan, Aleksander Mądry, Julian Michael, Neel Nanda, Dave Orr, Jakub Pachocki, Ethan Perez, Mary Phuong, Fabien Roger, Joshua Saxe, Buck Shlegeris, Martín Soto, Eric Steinberger, Jasmine Wang, Wojciech Zaremba, Bowen Baker, Rohin Shah, and Vlad Mikulik.
\newblock Chain of thought monitorability: A new and fragile opportunity for ai safety.
\newblock {\em arXiv preprint arXiv: 2507.11473}, 2025.

\bibitem{quora_question_pairs_kaggle}
Quora.
\newblock Quora question pairs.
\newblock \url{https://www.kaggle.com/competitions/quora-question-pairs/}, 2017.
\newblock Kaggle competition.

\bibitem{NEURIPS2024_8f395480}
Clayton Sanford, Bahare Fatemi, Ethan Hall, Anton Tsitsulin, Mehran Kazemi, Jonathan Halcrow, Bryan Perozzi, and Vahab Mirrokni.
\newblock Understanding transformer reasoning capabilities via graph algorithms.
\newblock In A.~Globerson, L.~Mackey, D.~Belgrave, A.~Fan, U.~Paquet, J.~Tomczak, and C.~Zhang, editors, {\em Advances in Neural Information Processing Systems}, volume~37, pages 78320--78370. Curran Associates, Inc., 2024.

\bibitem{sanford2024transformers}
Clayton Sanford, Daniel Hsu, and Matus Telgarsky.
\newblock Transformers, parallel computation, and logarithmic depth.
\newblock {\em arXiv preprint arXiv:2402.09268}, 2024.

\bibitem{liu2022transformers}
Bingbin Liu, Jordan~T Ash, Surbhi Goel, Akshay Krishnamurthy, and Cyril Zhang.
\newblock Transformers learn shortcuts to automata.
\newblock {\em arXiv preprint arXiv:2210.10749}, 2022.

\bibitem{wang2025learning}
Zixuan Wang, Eshaan Nichani, Alberto Bietti, Alex Damian, Daniel Hsu, Jason~D Lee, and Denny Wu.
\newblock Learning compositional functions with transformers from easy-to-hard data.
\newblock {\em arXiv preprint arXiv:2505.23683}, 2025.

\bibitem{hendrycks2020measuring}
Dan Hendrycks, Collin Burns, Steven Basart, Andy Zou, Mantas Mazeika, Dawn Song, and Jacob Steinhardt.
\newblock Measuring massive multitask language understanding.
\newblock In {\em International Conference on Learning Representations (ICLR)}, 2021.

\bibitem{wang2024mmlu}
Yizhong Wang, Yada Pruksachatkun, Sheng Chen, Zexuan Zhong, Pengfei Chen, et~al.
\newblock {MMLU-Pro}: A more challenging and reliable evaluation for massive multitask language understanding.
\newblock {\em arXiv preprint arXiv:2406.01574}, 2024.

\bibitem{suzgun2022challenging}
Mirac Suzgun, Nathan Scales, Nathanael Sch{\"a}rli, Sebastian Gehrmann, Yi~Tay, Hyung~Won Chung, Aakanksha Chowdhery, Quoc~V. Le, Ed~H. Chi, Denny Zhou, and Jason Wei.
\newblock Challenging big-bench tasks and whether chain-of-thought can solve them.
\newblock {\em arXiv preprint arXiv:2210.09261}, 2022.

\bibitem{clark2018think}
Peter Clark, Isaac Cowhey, Oren Etzioni, Tushar Khot, Ashish Sabharwal, Carissa Schoenick, and Oyvind Tafjord.
\newblock Think you have solved question answering? try arc, the ai2 reasoning challenge.
\newblock {\em arXiv preprint arXiv:1803.05457}, 2018.

\bibitem{zellers2019hellaswag}
Rowan Zellers, Ari Holtzman, Yonatan Bisk, Ali Farhadi, and Yejin Choi.
\newblock {H}ella{S}wag: Can a machine really finish your sentence?
\newblock In {\em Proceedings of the 57th Annual Meeting of the Association for Computational Linguistics}, pages 4791--4800, Florence, Italy, 2019. Association for Computational Linguistics.

\bibitem{sakaguchi2021winogrande}
Keisuke Sakaguchi, Ronan Le~Bras, Chandra Bhagavatula, and Yejin Choi.
\newblock Winogrande: An adversarial winograd schema challenge at scale.
\newblock {\em arXiv preprint arXiv:1907.10641}, 2019.

\bibitem{cobbe2021training}
Karl Cobbe, Vineet Kosaraju, Mohammad Bavarian, Mark Chen, Heewoo Jun, Lukasz Kaiser, Matthias Plappert, Jerry Tworek, Jacob Hilton, Reiichiro Nakano, Christopher Hesse, and John Schulman.
\newblock Training verifiers to solve math word problems.
\newblock {\em arXiv preprint arXiv:2110.14168}, 2021.

\bibitem{hendrycks2021measuring}
Dan Hendrycks, Collin Burns, Saurav Kadavath, Akul Arora, Steven Basart, Eric Tang, Dawn Song, and Jacob Steinhardt.
\newblock Measuring mathematical problem solving with the {MATH} dataset.
\newblock In {\em NeurIPS 2021 Datasets and Benchmarks Track}, 2021.

\bibitem{chen2021codex}
Mark Chen, Jerry Tworek, Heewoo Jun, Qiming Yuan, Henrique~Ponde de~Oliveira~Pinto, Jared Kaplan, Harri Edwards, Yuri Burda, Nicholas Joseph, Greg Brockman, Alex Ray, Raul Puri, Gretchen Krueger, Michael Petrov, Heidy Khlaaf, Girish Sastry, Pamela Mishkin, Brooke Chan, Scott Gray, Nick Ryder, Mikhail Pavlov, Alethea Power, Lukasz Kaiser, Mohammad Bavarian, Clemens Winter, Philippe Tillet, Felipe~Petroski Such, Dave Cummings, Matthias Plappert, Fotios Chantzis, Elizabeth Barnes, Ariel Herbert-Voss, William~Hebgen Guss, Alex Nichol, Alex Paino, Nikolas Tezak, Jie Tang, Igor Babuschkin, Suchir Balaji, Shantanu Jain, William Saunders, Christopher Hesse, Andrew~N. Carr, Jan Leike, Josh Achiam, Vedant Misra, Evan Morikawa, Alec Radford, Matthew Knight, Miles Brundage, Mira Murati, Katie Mayer, Peter Welinder, Bob McGrew, Dario Amodei, Sam McCandlish, Ilya Sutskever, and Wojciech Zaremba.
\newblock Evaluating large language models trained on code.
\newblock {\em arXiv preprint arXiv:2107.03374}, 2021.

\bibitem{austin2021program}
Jacob Austin, Augustus Odena, Maxwell Nye, Maarten Bosma, Henryk Michalewski, David Dohan, Ellen Jiang, Carrie Cai, Michael Terry, Quoc Le, and Charles Sutton.
\newblock Program synthesis with large language models.
\newblock {\em arXiv preprint arXiv:2108.07732}, 2021.

\bibitem{allenai:arc}
Peter Clark, Isaac Cowhey, Oren Etzioni, Tushar Khot, Ashish Sabharwal, Carissa Schoenick, and Oyvind Tafjord.
\newblock Think you have solved question answering? try arc, the ai2 reasoning challenge.
\newblock {\em arXiv:1803.05457v1}, 2018.

\bibitem{lambada}
Denis Paperno, Germán Kruszewski, Angeliki Lazaridou, Quan~Ngoc Pham, Raffaella Bernardi, Sandro Pezzelle, Marco Baroni, Gemma Boleda, and Raquel Fernández.
\newblock The lambada dataset, Aug 2016.

\bibitem{mihaylov2018can}
Todor Mihaylov, Peter Clark, Tushar Khot, and Ashish Sabharwal.
\newblock Can a suit of armor conduct electricity? a new dataset for open book question answering.
\newblock {\em arXiv preprint arXiv:1809.02789}, 2018.

\bibitem{bisk2020piqa}
Yonatan Bisk, Rowan Zellers, Jianfeng Gao, Yejin Choi, et~al.
\newblock Piqa: Reasoning about physical commonsense in natural language.
\newblock In {\em Proceedings of the AAAI conference on artificial intelligence}, volume~34, pages 7432--7439, 2020.

\bibitem{hoffmann2022training}
Jordan Hoffmann, Sebastian Borgeaud, Arthur Mensch, Elena Buchatskaya, Trevor Cai, Eliza Rutherford, Diego de~Las Casas, Lisa~Anne Hendricks, Johannes Welbl, Aidan Clark, et~al.
\newblock Training compute-optimal large language models.
\newblock {\em arXiv preprint arXiv:2203.15556}, 2022.

\bibitem{que2024d}
Haoran Que, Jiaheng Liu, Ge~Zhang, Chenchen Zhang, Xingwei Qu, Yinghao Ma, Feiyu Duan, Zhiqi Bai, Jiakai Wang, Yuanxing Zhang, et~al.
\newblock D-cpt law: domain-specific continual pre-training scaling law for large language models.
\newblock In {\em Proceedings of the 38th International Conference on Neural Information Processing Systems}, pages 90318--90354, 2024.

\end{thebibliography}

\newpage
\appendix
\section{Empirical Validation of Prior Choice}
\label{app:prior_empirical}

\begin{figure}[htbp]
    \centering
    \includegraphics[width=\linewidth]{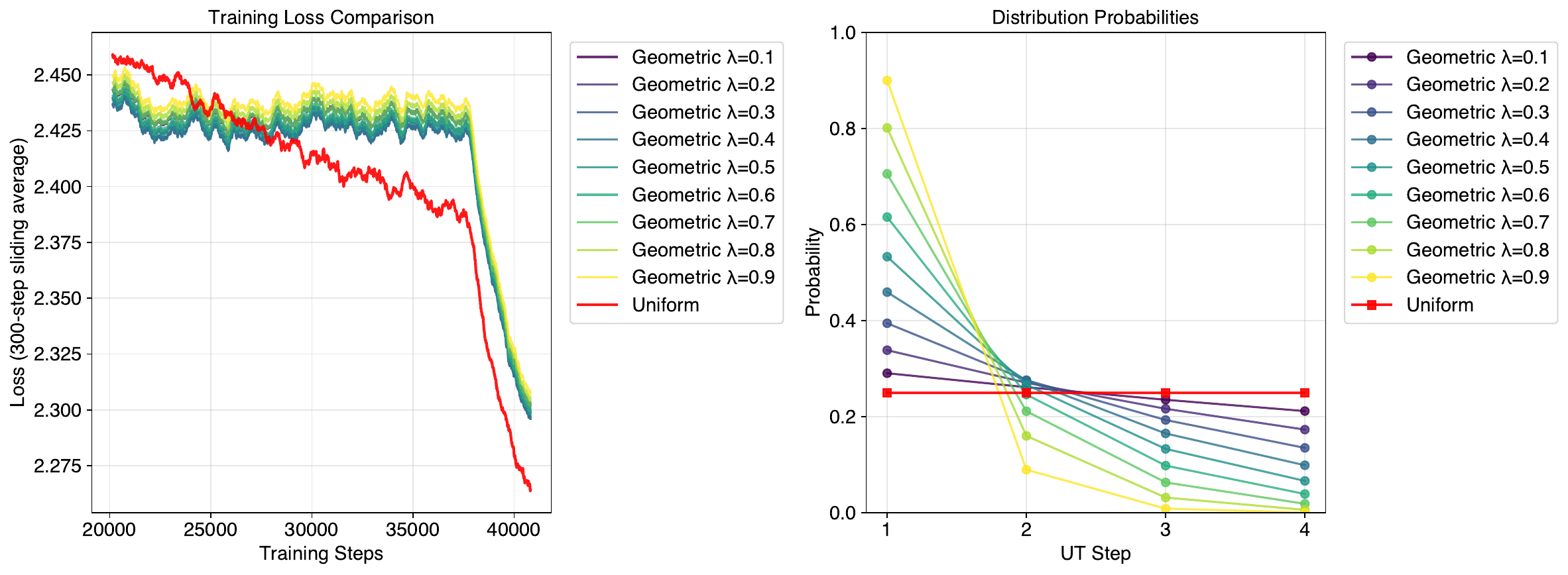}
    \vspace{-0.6em}
    \caption{\textbf{Effect of the prior over exit steps.} 
    \emph{Left:} training loss (300-step sliding average) for a \ut{} with $T_{\max}=4$ under different priors on $z$. 
    Colored curves correspond to geometric priors with parameter $\eta\in\{0.1,\dots,0.9\}$; the red curve uses a uniform prior. 
    Shaded regions indicate variability across runs. 
    \emph{Right:} prior probability over \ut{} steps induced by each $\eta$ (uniform shown in red). 
    Stronger geometric bias (larger $\eta$) concentrates mass on shallow steps, reducing credit assignment to deeper computation.}
    \label{fig:prior_loss}
\end{figure}

\paragraph{Experimental setup.}
Unless otherwise noted, we keep the model, data, optimizer, and schedule identical across conditions and \emph{only} change the prior $\pi$ used in the KL term of the loss.
All results are obtained on a 776M-parameter \ut{} with $T_{\max}=4$ recurrent steps.
Training is performed on the FineWeb-Edu corpus~\cite{penedo2024fineweb} for a total of 20B tokens with a global batch of 50K tokens per optimization step, i.e., roughly 40K steps in total.%
\footnote{The loss curves plot a 300-step sliding average over the training trajectory.}
For geometric priors we sweep $\lambda\!\in\!\{0.1,0.2,\dots,0.9\}$; the uniform prior assigns equal mass to all steps.
To assess variability, we repeat each condition with multiple random seeds; shaded areas in \Cref{fig:prior_loss} denote the variability across runs.
All other hyperparameters follow our training recipe, keeping $\beta$ fixed across prior choices.

\paragraph{Convergence and final loss.}
As shown on the left of \Cref{fig:prior_loss}, the uniform prior consistently achieves lower training loss and cleaner convergence on the 776M \ut{}.
Geometric priors plateau higher, with the gap widening as $\lambda$ grows (i.e., stronger bias toward early exit), reflecting weaker supervision for deeper iterations.

\paragraph{Stability and exploration.}
Geometric priors exhibit larger late-training oscillations, consistent with premature collapse of $q_\phi(z\!\mid\!x)$ onto shallow steps and reduced entropy.
The uniform prior imposes no structural depth preference, so the KL term behaves as pure entropy regularization: exploration is maintained longer, and the model can allocate probability mass across multiple depths until it has learned which examples benefit from deeper computation.

\paragraph{Depth utilization.}
The right panel of \Cref{fig:prior_loss} visualizes the priors.
Large-$\lambda$ geometric priors concentrate mass at $t{=}1,2$, neglecting deeper steps ($t{\ge}3$) of credit assignment; this undermines the ``deeper is better'' property.
With a uniform prior, all depths receive comparable signal, enabling later iterations to specialize and deliver higher accuracy when maximum depth is allowed at inference.

\paragraph{Compute--accuracy trade-off.}
Although the uniform prior does not explicitly favor early exit, it does not preclude efficient inference: at test time we can still cap steps or apply a halting threshold.
For a fixed average step budget, models trained with a uniform prior achieve a strictly better accuracy-compute Pareto frontier than those trained with geometric priors, indicating that unbiased depth exploration during pre-training turns into better deployment trade-offs.
\section{Physics of \ut{}s}
\label{appendix:physics}
In this appendix, we conclude all the experimental settings and details in \Cref{sec:understanding_ut}. \Cref{appendix:capo} includes the experiments on knowledge capacity; \cref{appendix:mano} includes the settings on knowledge manipulation synthetic tasks. \Cref{appendix:multihop} introduces the detailed setting on the synthetic QA task following \cite{yao2025language}. Finally, \Cref{appendix:theory} provides the theoretical results, detailed proof, and the discussion with the current theoretical results.
\subsection{Capo: knowledge capacity}
\label{appendix:capo}
In this section, we introduce the knowledge capacity proposed in \cite{AL2024-knowledge3,Allenzhu2025-canon}. The task evaluates models' efficiency in memorizing factual knowledge within its parameters, which is measured by  \textit{bits per parameter}. We tested different sizes of models and visualize the knowledge scaling law through plotting \textit{bits v.s. parameter number}.

\paragraph{Dataset: Synthetic Biographies} We synthesize fake biographies following the $\mathrm{bioS}(N)$ dataset in \cite{AL2024-knowledge3}. Specifically, we generate $N$ biographies of a random generated person together with their date of birth, city of birth, university, major, and employer. In our work, we online sample the individual attributes and generate the biographies in natural language using a random selected fixed template. An illustrative example is:

\begin{center}
    \emph{Layla Jack Beasley celebrates their birthday on January 24, 1914. They spent formative years in Portland, ME. They focused on Business Analytics. They supported operations for Delta Air Lines Inc. in Atlanta, GA. They received their education at Pepperdine University.}
\end{center}

\paragraph{Model} We use original GPT2 architecture and replace the positional encoding with RoPE \cite{su2023roformerenhancedtransformerrotary}. In the Capo task, we tie the LM head and the embedding layer. To test the capability of universal transformer, we also added looping module s.t. the transformer blocks can be looped several times. We explore a broad range of model sizes varying in hidden dimension and depth. The notation $a$-$b$-l$c$ represents the model with $64a$ hidden dimensions ($a$ attention heads with each head 64 dimensions), $b$ layers, and $c$ \ut{} steps (loops). The context length is set to 512. 

\paragraph{Training details} We use AdamW optimizer by setting $(\beta_1,\beta_2)=(0.9,0.98), \epsilon=10^{-6}$ with 1000 steps of warmup followed by a cosine learning rate schedule from 1 to 0.1$\times$ of the original learning rate. We use bf16 training and packing is used during training. We masked different pieces of biographies from each other in each concatenated chunk.

We pass each data piece for 1000 times (similar to the 1000-exposure in \cite{AL2024-knowledge3}) during training. Since the final performance is not sensitive to learning rate choices, we consider learning rate $\eta = 0.001, wd=0.02$, and total batch size 192. We pick $N\in\{20K, 50K, 100K, 200K, 500K\}$. 

\paragraph{Evaluation: Knowledge Capacity Ratio} 
After pre-training on the bioS($N$) dataset, we assess a model’s \emph{knowledge capacity}, 
defined as the number of bits of information it can reliably store. To make this measure 
comparable across models of different sizes, the raw bit count is normalized by the number 
of model parameters, yielding a ``bits per parameter'' metric. 
The derivation and motivation of the metric in discussed in \cite{AL2024-knowledge3}. For readers, we refer the detailed setting to Section 2.1 of \cite{AL2024-knowledge3}.

\begin{definition}
    Given a model $F$ with $P$ parameters trained over the bioS($N$) dataset $Z$, suppose it gives 
$p_1=\text{loss}_{\text{name}}(Z)$ and $p_2=\text{loss}_{\text{value}}(Z)$. These losses are measured in nats: for each person, sum the negative log-likelihoods over the name tokens for $p_1$, or over the tokens of all attribute values for $p_2$, and then average over the $N$ people. Name loss is evaluated without conditioning on the person's identity, while each attribute value is predicted given the person's name and the attribute being queried.
The \emph{capacity ratio} 
and the maximum achievable capacity ratio are
 defined as
\[
\begin{aligned}
R(F)&\;\overset{\text{def}}{=}\;\frac{N\log_2\!\left(N_0/e^{p_1}\right)+N\log_2\!\left(S_0/e^{p_2}\right)}{P},\\
R_{\max}(F)&\;\overset{\text{def}}{=}\;\frac{N\log_2(N_0/N)+N\log_2 S_0}{P}.
\end{aligned}
\]
for 
$N_0 = 400 \times 400 \times 1000$, $S_0 = 2 \times (12 \cdot 28 \cdot 200) \times 200 \times 300 \times 100 \times 263$ as all possible configurations.
The maximum corresponds to $p_1=\log N$ and $p_2=0$, with the natural logarithm: even a model that memorizes all biographies must represent the uncertainty over the $N$ equally likely names \cite{AL2024-knowledge3}.
\end{definition}

Ignoring names, each person encodes approximately $\log_2(S_0) \approx 47.6 \; \text{bits of knowledge}.$ 
The evaluation accounts for \emph{partial correctness}. For instance, if a model 
recalls the year of a person’s birth but not the exact date, the partially correct information 
still contributes to the overall bit-level computation. This approach allows for a fine-grained 
measurement of knowledge retention, rather than relying on a strict all-or-nothing scoring.

\subsection{Mano: knowledge manipulation}
\label{appendix:mano}
We followed \cite{Allenzhu2025-canon} and used the Mano task to investigate the models' capability of manipulating stored knowledge within the parameters without intermediate thoughts. 

\textbf{Dataset} The dataset consists of modular arithmetic instances with tree structures of $\ell$ operations, where the number of operations $\ell\le L$ as the maximum length. $\ell$ is uniformly sampled from $[1,L]$. The expressions are presented in prefix notation. For example, a length-3 instance is:
$$\texttt{<bos> <len\_3> - * a b + c d <ans> ans}$$
which corresponds to $(a*b)+(c-d)\mod 23$. All the operations are on $\mathbb{F}_{23}$. The task only involves $(+,-,*)$. The only tokens we use are the operations, numbers from 0 to 22, and the special \texttt{<bos>, <ans>} and length tokens \texttt{len\_\{i\}} with $i\in [0,L]$.

\paragraph{Training details} We use AdamW optimizer with $(\beta_1,\beta_2)=(0.9,0.98), \epsilon=10^{-6}$ and gradient clipping with maximum norm 1.0. We employ 1000 steps of warmup followed by a cosine learning rate schedule to minimal learning rate 0.1 of the peak learning rate. We use bf16 training with packing and set the context length to 1024 tokens. Different pieces of mano problems are masked from each other in each concatenated chunk during training.

We conduct hyperparameter search over learning rates $lr \in \{0.00005, 0.0001, 0.0002, 0.0005\}$
with weight decay 0.1 and global batch size 128. We experiment with model depths $L \in \{10, 16, 24\}$ layers and hidden dimension 1024. Training is performed for $\{80K, 110K, 200K\}
$ steps respectively for different difficulties. We run all experiments across 3 random seeds and report the best performance. 

\paragraph{Evaluation} During evaluation, we only use the expressions with the hardest length $\ell=L$. Accuracy is computed separately due to the masks. We consider exact match accuracy since the final answer is single-token.

\subsection{Multi-hop question answering on synthetic relations}
\label{appendix:multihop}
We followed \cite{yao2025language} to construct the natural language multi-hop QA task. Comparing with Mano, the QA task is more knowledge-heavy and with a slightly simpler structure. \cite{yao2025language} found that the model needs exponential many $k$-hop data for traditional transformer to learn. We chose this task to investigate if recursive structure in the reused-parameters can improve the sample efficiency of the task, showing better manipulation capability of \ut{}.

\paragraph{Dataset} The dataset contains 
$|\mathcal{E}|$ entities—each with a unique name—and 
$N$ relation types. We created 500 distinct single-token person names (e.g., Jennifer) and 20 single-token relation names (e.g., instructor) to serve as namespaces for entities and relations. We reused the name list in \cite{yao2025language}. The complete list of relation names and a partial list of entity names appear in Tables 5 and 6 in \cite{yao2025language}. The multi-hop questions are generated through a $K=5$ hierarchical layers, where each layer has 100 individuals. Each entity is connected to $|\mathcal{R}|$ randomly chosen person in the next layer. This structure naturally generates $|\mathcal{E}|/5 \times |\mathcal{R}|^k$ $k$-hop questions. In our setting, since we only consider 3-hop questions, the number should be $8\times 10^5$.

For training, we use part of all the 3-hop training set and test on the leave-out 3000 test questions. For each test instance, we greedy decode the single token answer given the question prompt (e.g. `Who is the instructor of the teacher of Bob? $\backslash n$ Answer:'). We evaluate the exact match accuracy.

\paragraph{Training details} We use AdamW optimizer with $(\beta_1,\beta_2)=(0.9,0.98), \epsilon=10^{-6}$ and gradient clipping 1.0. We run 1000 steps of linear warmup followed by a cosine learning rate schedule to minimal learning rate 0.1 of the peak learning rate. We use bf16 training with packing with context length 1024 tokens. QA pairs from distinct samples are masked from each other during training.

We use a base model architecture with 1024 hidden dimensions, 16 attention heads, and 6 layers. We allow it to loop in $\{1,2,4\}$ times. Following the experimental setup in \cite{yao2025language}, we set the learning rate to 0.0005 with 1000 warmup steps and train for a total of 20,000 steps using batch size 2048. We run all experiments across 4 random seeds and report the average performance. 

\subsubsection{Additional experimental results}
\begin{figure}[t]
\centering
\vspace{-5pt}
\setlength{\tabcolsep}{6pt}
\begin{minipage}[t]{0.48\textwidth}
\small
\centering
\includegraphics[width=\linewidth]{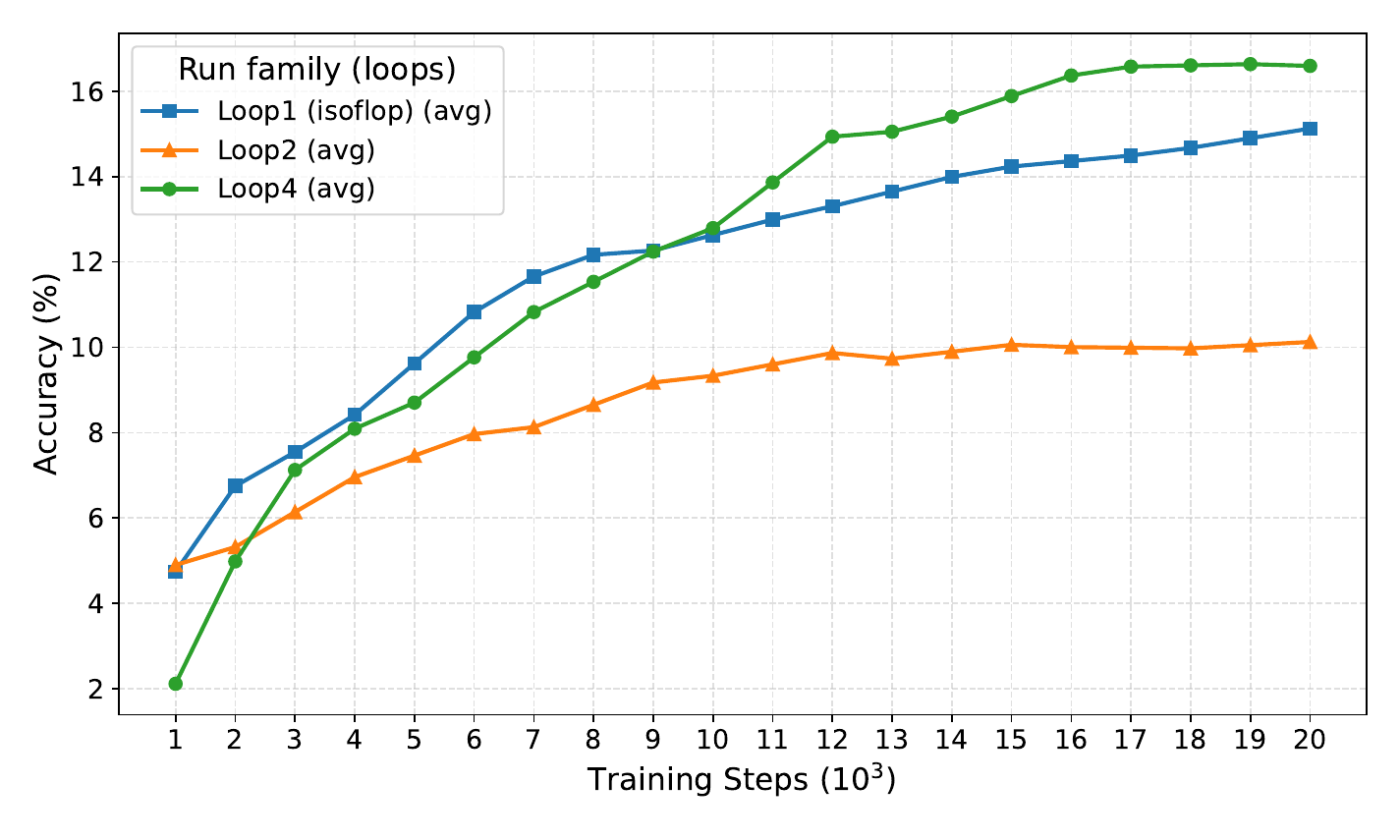}
\end{minipage}
\hfill
\begin{minipage}[t]{0.48\textwidth}
\centering
\includegraphics[width=\linewidth]{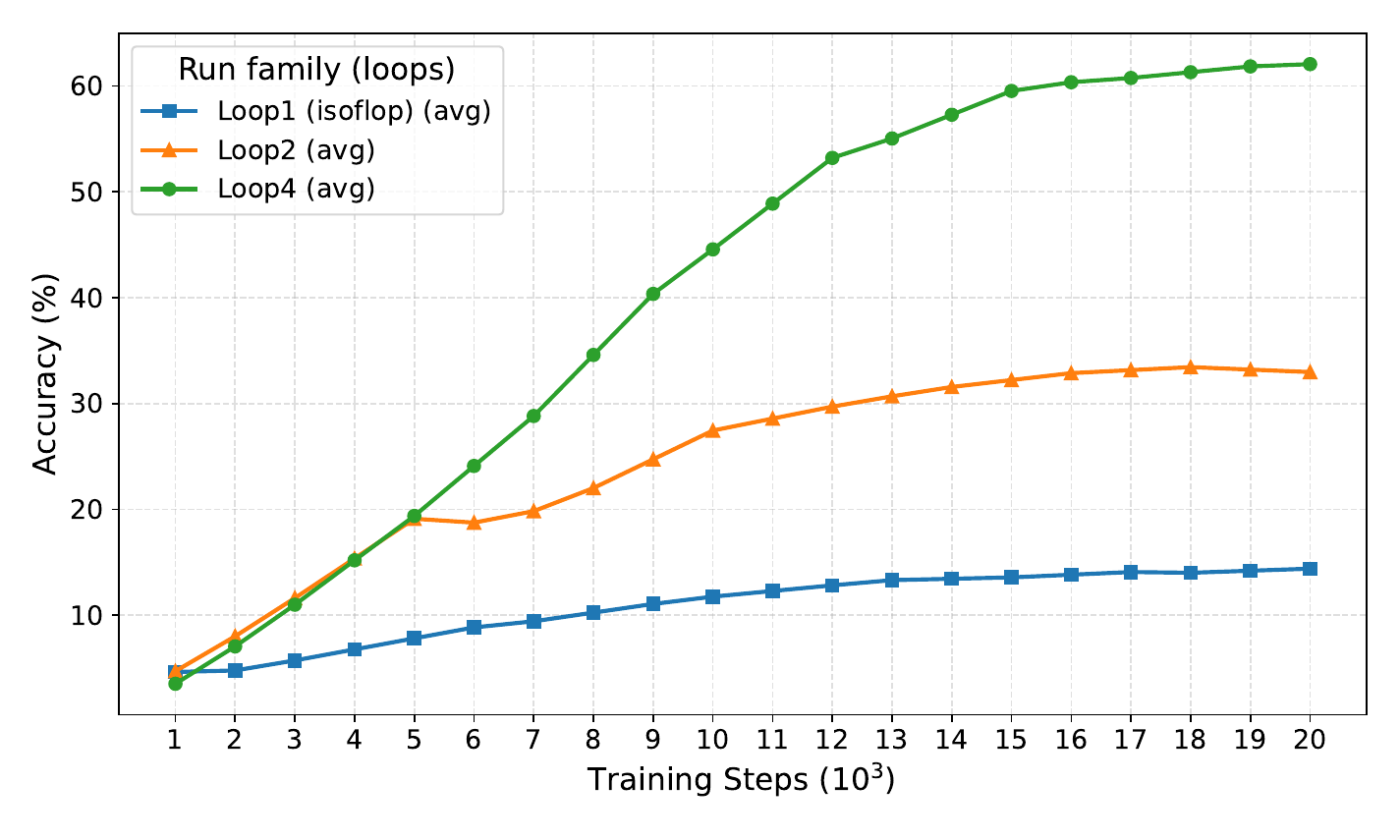}
\end{minipage}
\caption{
\textbf{Left \& Right.} We further train with $100000$ and $120000$ unique QA pairs for 20000 steps with context length 1024 and batch size 2048. We train the baseline with 24 layers, which is equivalent flops with the loop 4 transformers. Similar to the main text, models with more loops learn faster and achieve better performance comparing with models without loops, even with iso-flop transformers. The loop 2 average performance is weaker than the iso-flop version transformer since it has less equivalent depth when $N=10^5$, but it surpasses the baseline with more data provided. }
\label{fig:multi_hop_additional_isoflop}
\vspace{-5pt}
\end{figure}

As the supplement of the main text, we present additional experiments to show that the superiority of \ut{} is general across different number of unique samples. For presentation, we only consider the interval of $\{10^5, 1.2\times 10^5, 1.4\times 10^5\}$ to exhibit the difference between looped models and non-looped baselines. We also checked iso-flop baseline models with the same hidden dimension and 24 layers\footnote{We note that in \Cref{fig:multi_hop_additional_isoflop}, the iso-flop baseline with $N=1.2\times 10^5$ does not perform significantly better than the shallower version in the main paper. We conjecture that it could be because of the randomness, or insufficient hyperparameter tuning. We believe further follow-up experiments should be necessary to further validate this conclusion here in the appendix.}. The results are presented below in \Cref{fig:multi_hop_additional} and \Cref{fig:multi_hop_additional_isoflop}.

\begin{figure}[t]
\centering
\vspace{-5pt}
\setlength{\tabcolsep}{6pt}
\begin{minipage}[t]{0.48\textwidth}
\small
\centering
\includegraphics[width=\linewidth]{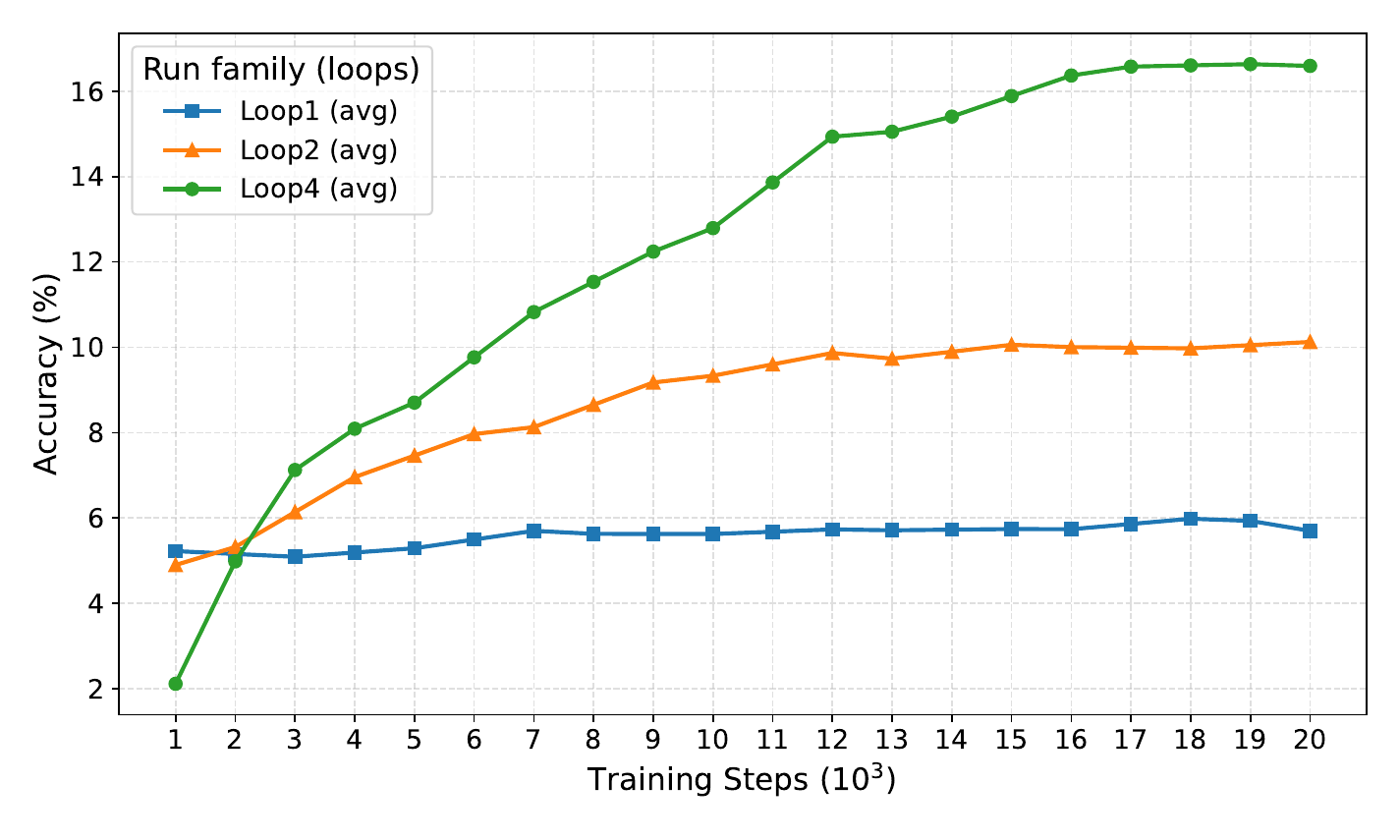}
\end{minipage}
\hfill
\begin{minipage}[t]{0.48\textwidth}
\centering
\includegraphics[width=\linewidth]{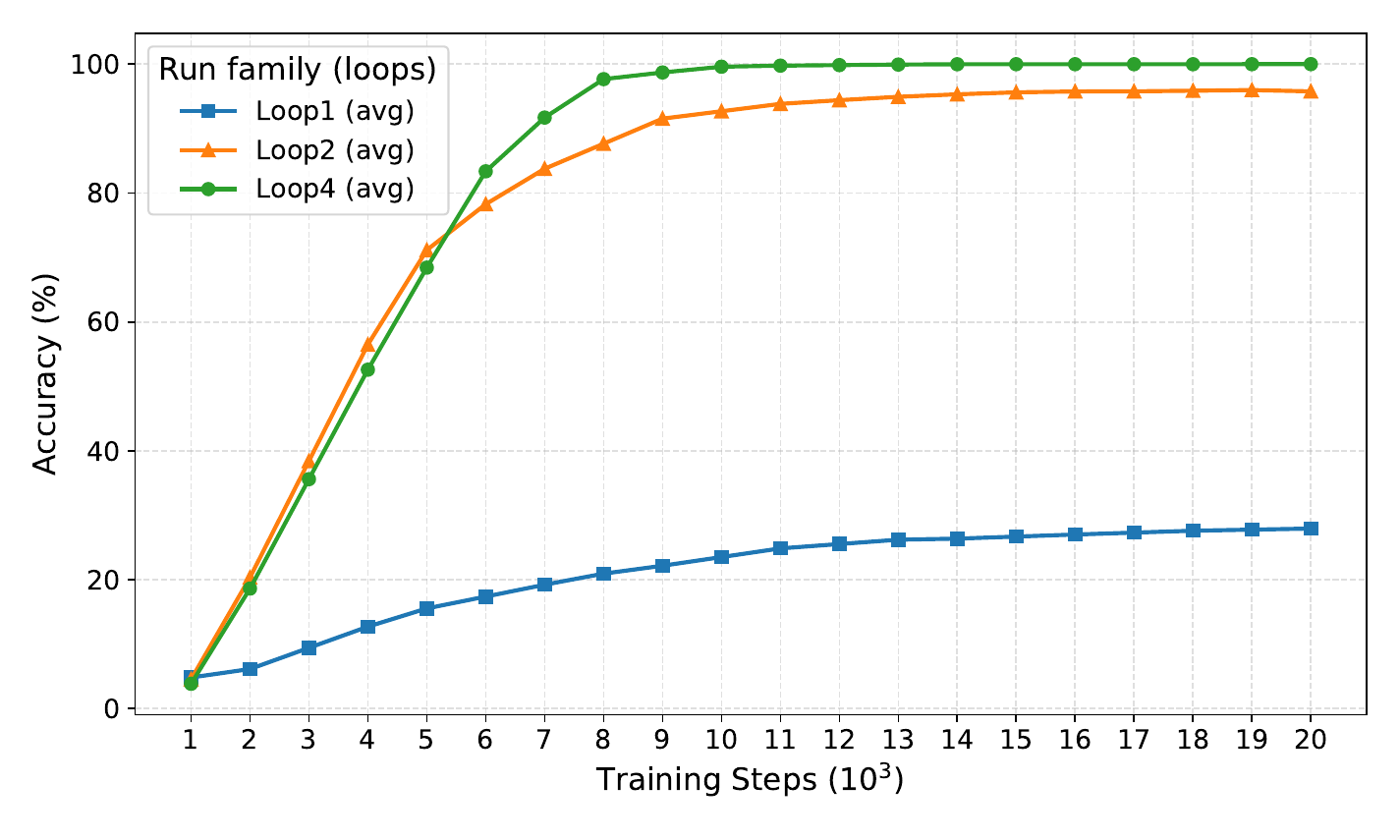}
\end{minipage}
\caption{
\textbf{Left \& Right.} We further train with $100000$ and $140000$ unique QA pairs for 20000 steps with context length 1024 and batch size 2048. Similar to the main text, models with more loops learn faster and achieve better performance comparing with models without loops.}
\label{fig:multi_hop_additional}
\vspace{-5pt}
\end{figure}

\subsection{Case study: improvements across different categories in MMLU}
\label{appendix:mmlu_case_study}
To validate our findings from synthetic tasks on a broad, real-world benchmark, we conducted a granular analysis of performance gains across all 57 sub-categories of MMLU. Our hypothesis is that if \ut{}s primarily enhance knowledge manipulation and reasoning, the largest performance gains should appear in procedural, reasoning-heavy tasks, while knowledge-heavy, retrieval-based subjects should see less improvement.

We measured the relative improvement by comparing the accuracy at the single recurrent step (Loop 1) against the accuracy at our fully trained depth (Loop 4). The detailed results for all 57 categories are available in \textbf{Table}~\ref{tab:performance_metrics}.
The analysis strongly supports our hypothesis. The categories with the most significant improvements are those requiring logical, maths, or procedural reasoning; conversely, categories that depend more on retrieving specific, memorized facts or nuanced world knowledge showed the most modest gains.

\begin{theobox*}{Categories with the most significant/modest improvements}
\vspace{0.05cm}
\small{
\vspace{0.05cm}

    \begin{multicols}{2}
        \hspace{-10pt}
        \colorbox{red!12}{\textbf{With most significant improvements:}}
        \vspace{0.1cm}

        \begin{itemize}[itemsep=0.0pt,topsep=0pt,leftmargin=*]
            \item {Elementary Mathematics}: +155.6\%
            \item {Formal Logic}: +143.3\%
            \item {Logical Fallacies}: +127.8\%
            \item {High School Statistics}: +126.9\%
        \end{itemize}
        \columnbreak
        \hspace{-10pt}
        \colorbox{blue!12}{\textbf{With most modest improvements:}}
        \vspace{0.1cm}

        \begin{itemize}[itemsep=0.0pt,topsep=0pt,leftmargin=*]
            \item {Moral Scenarios}: +7.8\%
            \item {Global Facts}: +8.3\%
            \item {Virology}: +13.7\%
            \item {Anatomy}: +21.4\%
        \end{itemize}
    \end{multicols}}
\end{theobox*}

This stark contrast indicates that the iterative computation is not simply increasing the model's accessible knowledge (as seen in the nearly flat `global\_facts` improvement) but is actively performing the multi-step symbolic manipulation required for complex subjects like logic and math. This real-world benchmark result corroborates our synthetic findings in Section~\ref{subsec:manipulation}, confirming that the \ut{} architecture's primary advantage lies in enhancing knowledge manipulation, not raw storage.

\begin{table}[htbp]
\centering
\caption{Performance metrics across different depths by category in MMLU.}
\label{tab:performance_metrics}
\small
\renewcommand{\arraystretch}{0.94}
\begin{tabular}{lrrrrl}
\hline
Category & Loop 1 & Loop 2 & Loop 3 & Loop 4 & Improvement\% \\
\hline
elementary\_mathematics & 0.3095 & 0.6190 & 0.7460 & 0.7910 & 155.5556 \\
formal\_logic & 0.2381 & 0.4841 & 0.5238 & 0.5794 & 143.3333 \\
logical\_fallacies & 0.3313 & 0.7178 & 0.7546 & 0.7546 & 127.7778 \\
high\_school\_statistics & 0.3102 & 0.5833 & 0.6620 & 0.7037 & 126.8657 \\
high\_school\_macroeconomics & 0.3692 & 0.6564 & 0.7513 & 0.7718 & 109.0278 \\
management & 0.3883 & 0.6699 & 0.7864 & 0.7961 & 105.0000 \\
high\_school\_government\_and\_politics & 0.3938 & 0.7202 & 0.8238 & 0.7979 & 102.6316 \\
high\_school\_microeconomics & 0.4076 & 0.7143 & 0.8361 & 0.8193 & 101.0309 \\
high\_school\_psychology & 0.4183 & 0.7817 & 0.8294 & 0.8385 & 100.4386 \\
high\_school\_biology & 0.4129 & 0.7452 & 0.8129 & 0.8161 & 97.6563 \\
college\_chemistry & 0.2600 & 0.5000 & 0.5500 & 0.5000 & 92.3077 \\
conceptual\_physics & 0.3830 & 0.6340 & 0.7149 & 0.7319 & 91.1111 \\
college\_biology & 0.3889 & 0.6806 & 0.7569 & 0.7431 & 91.0714 \\
machine\_learning & 0.2500 & 0.4286 & 0.5089 & 0.4732 & 89.2857 \\
miscellaneous & 0.3870 & 0.6564 & 0.7101 & 0.7178 & 85.4785 \\
high\_school\_geography & 0.4444 & 0.7121 & 0.7980 & 0.8182 & 84.0909 \\
high\_school\_physics & 0.2649 & 0.3841 & 0.5033 & 0.4834 & 82.5000 \\
high\_school\_chemistry & 0.3054 & 0.5320 & 0.5665 & 0.5517 & 80.6452 \\
college\_medicine & 0.3584 & 0.5607 & 0.6416 & 0.6358 & 77.4194 \\
college\_computer\_science & 0.3500 & 0.5100 & 0.6000 & 0.6100 & 74.2857 \\
professional\_accounting & 0.2872 & 0.4539 & 0.4929 & 0.5000 & 74.0741 \\
world\_religions & 0.4211 & 0.6374 & 0.6959 & 0.7251 & 72.2222 \\
high\_school\_computer\_science & 0.4600 & 0.6800 & 0.7600 & 0.7800 & 69.5652 \\
clinical\_knowledge & 0.4151 & 0.5887 & 0.6415 & 0.6943 & 67.2727 \\
astronomy & 0.4013 & 0.6184 & 0.6711 & 0.6711 & 67.2131 \\
prehistory & 0.3765 & 0.5586 & 0.6389 & 0.6235 & 65.5738 \\
high\_school\_us\_history & 0.4314 & 0.6912 & 0.7304 & 0.7108 & 64.7727 \\
professional\_psychology & 0.3709 & 0.5392 & 0.5850 & 0.6095 & 64.3172 \\
philosophy & 0.4405 & 0.6495 & 0.7042 & 0.7106 & 61.3139 \\
business\_ethics & 0.4200 & 0.6400 & 0.6500 & 0.6700 & 59.5238 \\
high\_school\_mathematics & 0.3000 & 0.4444 & 0.5037 & 0.4778 & 59.2593 \\
high\_school\_european\_history & 0.5030 & 0.6909 & 0.7273 & 0.8000 & 59.0361 \\
medical\_genetics & 0.4500 & 0.6900 & 0.7100 & 0.7100 & 57.7778 \\
human\_sexuality & 0.4580 & 0.6336 & 0.7023 & 0.7176 & 56.6667 \\
computer\_security & 0.4500 & 0.6200 & 0.6700 & 0.7000 & 55.5556 \\
college\_physics & 0.2549 & 0.3333 & 0.3922 & 0.3922 & 53.8462 \\
international\_law & 0.5124 & 0.7107 & 0.7769 & 0.7851 & 53.2258 \\
marketing & 0.5726 & 0.8547 & 0.8803 & 0.8761 & 52.9851 \\
nutrition & 0.4510 & 0.6634 & 0.6863 & 0.6863 & 52.1739 \\
college\_mathematics & 0.2900 & 0.3700 & 0.4200 & 0.4400 & 51.7241 \\
econometrics & 0.3421 & 0.4123 & 0.5088 & 0.5175 & 51.2821 \\
sociology & 0.5373 & 0.7413 & 0.7910 & 0.8010 & 49.0741 \\
professional\_medicine & 0.3787 & 0.5368 & 0.5735 & 0.5625 & 48.5437 \\
high\_school\_world\_history & 0.5274 & 0.7300 & 0.7595 & 0.7722 & 46.4000 \\
human\_aging & 0.4484 & 0.6143 & 0.6457 & 0.6502 & 45.0000 \\
security\_studies & 0.5265 & 0.6898 & 0.7510 & 0.7592 & 44.1860 \\
professional\_law & 0.3246 & 0.4055 & 0.4596 & 0.4570 & 40.7631 \\
public\_relations & 0.4636 & 0.6182 & 0.6727 & 0.6364 & 37.2549 \\
us\_foreign\_policy & 0.5900 & 0.7200 & 0.8100 & 0.8000 & 35.5932 \\
electrical\_engineering & 0.4483 & 0.5655 & 0.5862 & 0.6069 & 35.3846 \\
abstract\_algebra & 0.2700 & 0.3000 & 0.3700 & 0.3600 & 33.3333 \\
moral\_disputes & 0.5491 & 0.6503 & 0.6850 & 0.6994 & 27.3684 \\
anatomy & 0.4148 & 0.5111 & 0.5333 & 0.5037 & 21.4286 \\
jurisprudence & 0.5926 & 0.6944 & 0.7593 & 0.7130 & 20.3125 \\
virology & 0.4398 & 0.5000 & 0.4880 & 0.5000 & 13.6986 \\
global\_facts & 0.3600 & 0.3700 & 0.3600 & 0.3900 & 8.3333 \\
moral\_scenarios & 0.2436 & 0.2693 & 0.2492 & 0.2626 & 7.7982 \\
\hline
\end{tabular}
\end{table}
\subsection{Theory: latent thought with \ut{}}
\label{appendix:theory}
In this section, we prove that a simplified looped transformer can solve graph reachability, with part of the graph stored in its parameters, in $O(1+\log n)$ loops. Related expressivity results study transformers and looped transformers with padding \cite{merrill2025exact, merrill2025little, NEURIPS2024_8f395480, sanford2024transformers}, as well as state tracking \cite{liu2022transformers, wang2025learning}. Our construction uses full attention over graph tokens and an explicit elementwise threshold. These assumptions differ from the causal attention and normalization used in our language models.

We consider a fixed hidden knowledge graph $G$, encoded in the model parameters, and an input context graph $G_{\mathrm{ctx}}$. The model must compose edges from both graphs to answer a reachability query.
\begin{definition}[Graph reachability on a knowledge graph]
Let $V=\{v_1,\ldots,v_n\}$, with $n\ge 1$, and let $G=(V,E)$ and $G_{\mathrm{ctx}}=(V,E_{\mathrm{ctx}})$ be directed graphs. Given source and target indices $s,t\in\{1,\ldots,n\}$, output $1$ if there is a directed path from $v_s$ to $v_t$ in $G+G_{\mathrm{ctx}}:=(V,E\cup E_{\mathrm{ctx}})$, and $0$ otherwise. We allow paths of length zero, so every vertex reaches itself. Let $D$ be the maximum finite shortest-path distance over reachable ordered pairs in the combined graph. Thus $0\le D\le n-1$, including when the graph is disconnected.
\end{definition}

\paragraph{Transformer architecture}
We use a single attention head and a two-layer ReLU MLP, with all parameters shared across loops. Tokens are columns. Attention is unmasked, and softmax normalizes each column over keys:
\[
\mathrm{Attn}_{Q,K,V}(H)=VH\,\mathrm{softmax}_{\mathrm{col}}(H^\top K^\top QH).
\]
For this theoretical construction, define the elementwise threshold operator
\[
\mathrm{LN}_{\tau}(X)_{k,j}=\mathbf{1}\{X_{k,j}>\tau\},
\qquad \tau=\frac{1}{2n}.
\]
This is a Boolean threshold, not standard LayerNorm or RMSNorm. Each loop computes
\begin{align*}
H_{i+\frac12}&=\mathrm{LN}_{\tau}\bigl(H_i+\mathrm{Attn}_{Q,K,V}(H_i)\bigr),\\
H_{i+1}&=\mathrm{LN}_{\tau}\bigl(H_{i+\frac12}+W_2\mathrm{ReLU}(W_1H_{i+\frac12})\bigr).
\end{align*}

\paragraph{Input format}
Let $A,A_{\mathrm{ctx}}\in\{0,1\}^{n\times n}$ be the adjacency matrices, with $(A)_{k,j}=1$ exactly when $v_j\to v_k$ is an edge of $G$, and the same convention for $A_{\mathrm{ctx}}$. Write $\lor$ for elementwise Boolean OR. The input consists of $n$ tokens of dimension $2n$:
\[
H_0=\begin{bmatrix}I_n\\R_0\end{bmatrix},
\qquad R_0=I_n\lor A_{\mathrm{ctx}}.
\]
The identity in $R_0$ encodes zero-length paths and ensures that each query has at least one selected key, even at an isolated vertex. As before, we assume this input encoding is supplied and use a direct readout: for query $(s,t)$, return $(R_L)_{t,s}$ after $L$ loops.

\begin{theorem}[\ut{} solves reachability in logarithmically many loops]
\label{appendix_thm:ut_graph_connectivity}
For each $n\ge 1$ and fixed hidden graph $G$ on $n$ vertices, there is a one-layer, single-head looped transformer of hidden dimension $2n$, with the full attention and threshold operator defined above, whose parameters are independent of $G_{\mathrm{ctx}}$ and $(s,t)$. For every context graph, it solves all reachability queries after
\[
L=1+\left\lceil\log_2\max\{1,D\}\right\rceil
\]
loops, and remains correct after any additional loops. In particular, $1+\lceil\log_2\max\{1,n-1\}\rceil$ loops suffice without knowing $D$.
\end{theorem}

The construction implements Boolean repeated squaring: attention composes the paths currently represented in the state, while the MLP inserts the edges of the hidden graph.
\begin{proof}
Choose a finite $\beta=\log(4n^2)$, with the natural logarithm, and set
\[
K=\beta\begin{bmatrix}I_n&0\\0&0\end{bmatrix},
\qquad Q=\begin{bmatrix}0&I_n\\0&0\end{bmatrix},
\qquad V=\begin{bmatrix}0&0\\0&I_n\end{bmatrix},
\]
\[
W_1=\begin{bmatrix}0&0\\A&0\end{bmatrix},
\qquad W_2=I_{2n},
\]
where each block in $K,Q,V,W_1$ has size $n\times n$. Attention and the MLP write only to the lower half of the state. Since $0<\tau<1$, the upper half remains $I_n$. We will show that the lower half remains a Boolean matrix $R_i$ with $I_n\preceq R_i$, where $\preceq$ denotes entrywise inclusion.

\textbf{One attention step.} Suppose $H_i=[I_n;R_i]$, and define
\[
S_j=\{k:(R_i)_{k,j}=1\},\qquad d_j=|S_j|\ge 1.
\]
The score matrix is $H_i^\top K^\top QH_i=\beta R_i$. Thus the weight of key $k$ for query $j$ is
\[
\alpha_{k,j}=\frac{\exp(\beta(R_i)_{k,j})}{d_j e^\beta+n-d_j}.
\]
For any selected key $k\in S_j$, we have $\alpha_{k,j}\ge 1/n$. The total weight on unselected keys satisfies
\[
\sum_{k\notin S_j}\alpha_{k,j}
=\frac{n-d_j}{d_j e^\beta+n-d_j}
\le ne^{-\beta}=\frac{1}{4n}<\tau.
\]
Consequently, a coordinate already present in the residual, or contributed by at least one selected value, exceeds $\tau$. If neither occurs, its entire contribution comes from unselected keys and is below $\tau$. Thresholding therefore gives exactly
\[
R_{i+\frac12}=R_i\lor R_i^2,
\]
where multiplication and powers of Boolean matrices use AND and OR. In particular, the diagonal remains one. The positive threshold is essential: for any finite $\beta$, softmax assigns positive weight to unselected keys, so a zero threshold would introduce false positives.

\textbf{One MLP step.} Because the upper half is $I_n$ and $A$ is nonnegative,
\[
W_2\mathrm{ReLU}\left(W_1\begin{bmatrix}I_n\\R_{i+\frac12}\end{bmatrix}\right)
=\begin{bmatrix}0\\A\end{bmatrix}.
\]
The threshold therefore implements the exact recurrence
\begin{equation}\label{eq:reachability_recurrence}
R_{i+1}=R_i\lor R_i^2\lor A.
\end{equation}

\textbf{Coverage and soundness.} Let $C=I_n\lor A_{\mathrm{ctx}}\lor A$, and let $C^*$ be its reflexive transitive closure. We claim that for every $i\ge 1$,
\[
C^{2^{i-1}}\preceq R_i\preceq C^*.
\]
At $i=1$, the recurrence includes both $R_0$ and $A$, hence $C\preceq R_1$. Every entry of $R_0$, $R_0^2$, and $A$ describes a genuine path in the combined graph, giving $R_1\preceq C^*$. For the induction step, monotonicity of Boolean multiplication gives
\[
C^{2^i}=\bigl(C^{2^{i-1}}\bigr)^2\preceq R_i^2\preceq R_{i+1}.
\]
Conversely, composing two genuine paths or inserting a hidden-graph edge cannot create an unreachable pair, so $R_{i+1}\preceq C^*$. This proves both coverage of short paths and absence of false positives; it does not assert that intermediate states contain only paths below that length bound.

Since $C$ includes the identity, $C^m$ contains every path of length at most $m$. For $L=1+\lceil\log_2\max\{1,D\}\rceil$, every reachable pair has a shortest path of length at most $2^{L-1}$, so $C^{2^{L-1}}=C^*$ and $R_L=C^*$. The closure is a fixed point of the recurrence, proving correctness for all subsequent loops as well.
\end{proof}

\paragraph{Scope and computational cost}
The bound concerns sequential loops under the stated input and readout conventions. Full attention is needed to let every source token read every potentially reachable vertex; the proof does not establish the same result for a causal mask. The construction also uses a Boolean threshold instead of standard normalization. With conventional dense matrix multiplication, each loop uses $O(n^3)$ arithmetic operations at width $2n$ and $O(n^2)$ state entries. Thus the logarithmic loop count is not a logarithmic total-work bound. Our explicit representation uses $\Theta(n)$ coordinates per token; this construction alone does not establish that such width is necessary for every possible representation.

\paragraph{Alternative input formats}
The theorem assumes an adjacency-column encoding. An edge-list encoding, such as that considered in \cite{zhu2025reasoning}, requires an additional construction that gathers edges into this representation. The cost and masking requirements of that conversion are not proved here and are not included in our loop bound.

\section{Evaluations}
\subsection{Evaluation Settings}
\label{sec:appendix_eval}
\paragraph{Base Model Evaluation Settings}
Table~\ref{tab:eval_base_settings} details the evaluation settings and frameworks used for the Ouro base models.

\begin{table}[htbp]
\centering
\caption{Evaluation settings and benchmark sources for base models.}
\label{tab:eval_base_settings}
\footnotesize
\begin{tabular}{lll}
\toprule
\textbf{Benchmark} & \textbf{Settings} & \textbf{Framework}\\
\midrule
\multicolumn{3}{l}{\textbf{General}} \\
\quad MMLU~\citep{hendrycks2020measuring} & logprobs, 5-shot & \texttt{lm-eval-harness}\\
\quad MMLU-Pro~\citep{wang2024mmlu} & strict match, 5-shot CoT & \texttt{lm-eval-harness}  \\
\quad BBH~\citep{suzgun2022challenging} & strict match, 3-shot CoT & \texttt{lm-eval-harness}\\
\quad ARC-C~\citep{clark2018think} & logprobs, 25-shot & \texttt{lm-eval-harness} \\
\quad HellaSwag~\citep{zellers2019hellaswag} & logprobs, 10-shot & \texttt{lm-eval-harness}\\
\quad Winogrande~\citep{sakaguchi2021winogrande} & logprobs, 5-shot & \texttt{lm-eval-harness}\\

\midrule
\multicolumn{3}{l}{\textbf{Math}} \\
\quad GSM8k~\citep{cobbe2021training} & strict match, 3-shot CoT & \texttt{lm-eval-harness} \\
\quad MATH500~\citep{hendrycks2021measuring} & strict match, 5-shot CoT & In-house \\
\midrule
\multicolumn{3}{l}{\textbf{Code}} \\
\quad HumanEval~\citep{chen2021codex}  & pass@1 & \texttt{evalplus} \\
\quad HumanEval+~\citep{evalplus} & pass@1 & \texttt{evalplus}\\
\quad MBPP~\citep{austin2021program} & pass@1 & \texttt{evalplus} \\
\quad MBPP+~\citep{evalplus} & pass@1 & \texttt{evalplus}\\
\bottomrule
\end{tabular}
\end{table}

\paragraph{Reasoning Model Evaluation Settings}
Table~\ref{tab:eval_reasoning_settings} details the evaluation settings and protocol used for the Ouro-Thinking reasoning models, as described in Section 5.2. All reasoning benchmarks utilized an in-house evaluation harness and an LLM-as-judge protocol with a fixed rubric.

\begin{table}[htbp]
\centering
\caption{Evaluation settings and protocol for reasoning models (Ouro-Thinking).}
\label{tab:eval_reasoning_settings}
\footnotesize
\begin{tabular}{lll}
\toprule
\textbf{Benchmark} & \textbf{Protocol} & \textbf{Decoding Settings}\\
\midrule
AIME 2024/2025 & In-house harness; LLM-as-judge & temp=1.0, top\_p=0.7 \\
OlympiadBench & In-house harness; LLM-as-judge & temp=1.0, top\_p=0.7 \\
GPQA & In-house harness; LLM-as-judge & temp=1.0, top\_p=0.7 \\
SuperGPQA & In-house harness; LLM-as-judge & temp=1.0, top\_p=0.7 \\
BeyondAIME & In-house harness; LLM-as-judge & temp=1.0, top\_p=0.7 \\
HLE & In-house harness; LLM-as-judge & temp=1.0, top\_p=0.7 \\
\bottomrule
\end{tabular}
\end{table}

\section{Scaling Law for \ut{}s}

To further explore the potential of \ut{}, we conduct a series of small-scale experiments to investigate the scalability and predictability inherent in \ut{}. Specifically, our work focuses on the following three research questions:

\begin{itemize}[itemsep=0.0pt,topsep=0pt,leftmargin=*]
    \item \textbf{RQ1}: What is the performance gap between standard models and \ut{}?
    \item \textbf{RQ2}: How do recurrent steps impact the total loss and step-wise loss in the context of \ut{}?
    \item \textbf{RQ3}: What is the inherent connection between total loss and step-wise loss?
\end{itemize}

\subsection{RQ1: What is the performance gap between standard models and \ut{}?}
\label{scaling_law_sub_1}

To understand the performance gap between standard models and \ut{}, we quantify this difference in terms of benchmark performance. We also observe how this gap varies with changes in recurrent step and model size to guide the further scaling and iteration of \ut{}.

\paragraph{Experimental Setup} We prepare five model sizes: 53M, 134M, 374M, 778M, and 1.36B. For recurrent steps, we prepare four different depths: 1, 2, 4, and 8. It is worth noting that for standard models, different recurrent steps effectively multiply the number of layers in the model. We evaluate the performance on the following benchmarks: ARC-Challenge~\citep{allenai:arc}, ARC-Easy~\citep{allenai:arc}, HellaSwag~\citep{zellers2019hellaswag}, LAMBADA~\citep{lambada}, OpenBookQA~\citep{mihaylov2018can}, and PIQA~\citep{bisk2020piqa}. In all sub-experiments, we train on 20B tokens using the FineWeb-Edu corpus~\citep{penedo2024fineweb}. We present the benchmark performance from the final step. For \ut{}, the recurrent step used for evaluating is the maximum recurrent step. 

\begin{figure}[htbp!]
    \centering
    \includegraphics[width=0.9\linewidth]{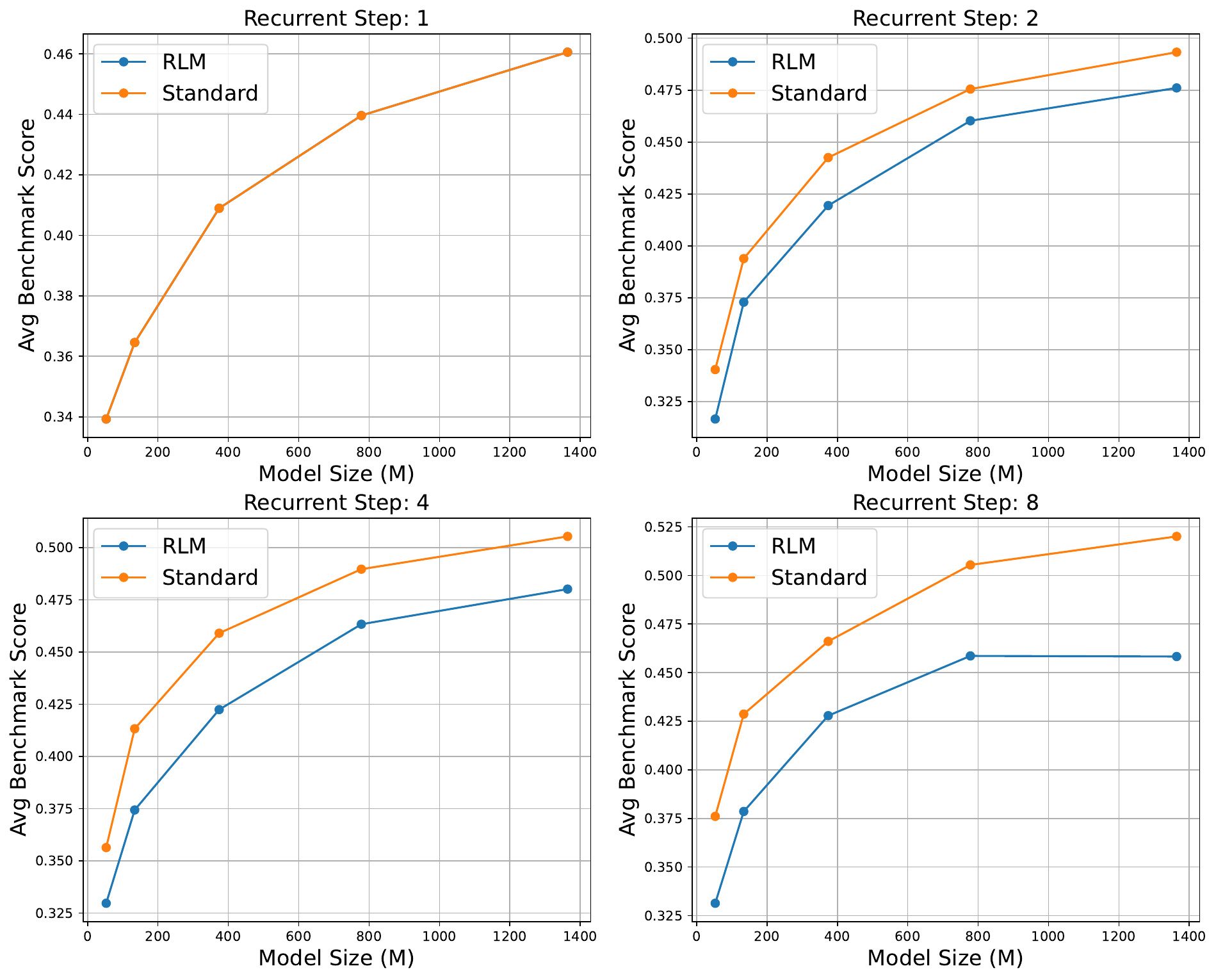}
    \caption{The average benchmark performance of \ut{} and Standard Transformer models under different recurrent steps as model size varies. With a recurrent step of 1 (\textit{top left}), both models have identical architectures, resulting in overlapping curves. Overall, as the model size increases, the benchmark performance improves. The average benchmark score demonstrates the average results of the six benchmarks.}
    \label{fig:scalinglaw_performance_vs_modelsize}
\end{figure}

\begin{figure}[htbp!]
    \centering
    \includegraphics[width=\linewidth]{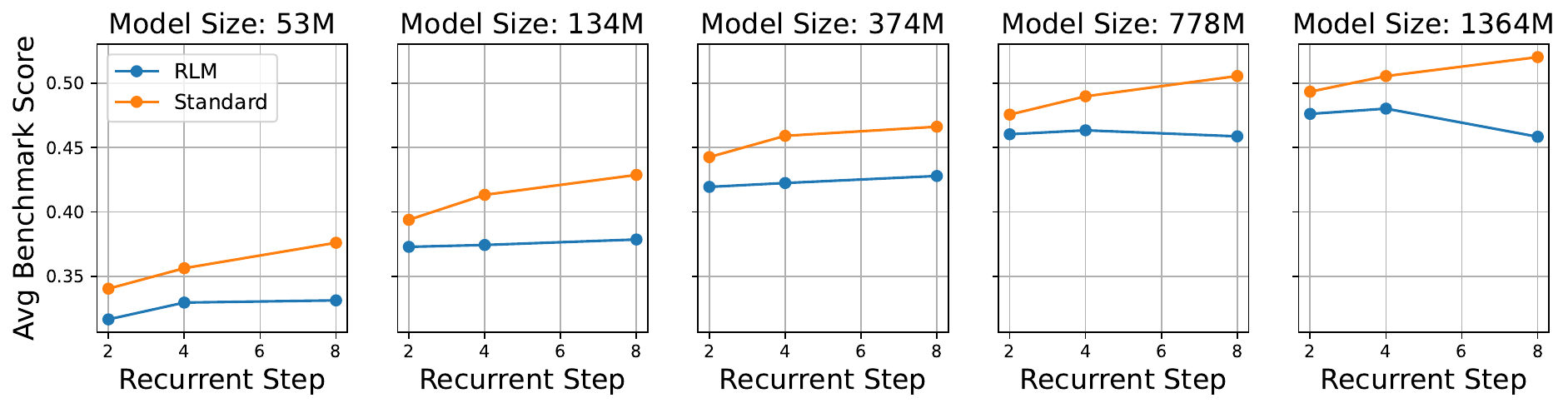}
    \caption{The average benchmark performance of \ut{} and Standard Transformer models under different model sizes as recurrent step varies. Except for the \ut{} at the model size of 778M and 1.364B, in all other cases, the benchmark performance of the model increases with the increase in recurrent steps.}
    \label{fig:scalinglaw_performance_vs_recurrentstep}
\end{figure}

By observing the trends in the curves, we derive the following observations:

1. Whether standard models or \ut{}, the model's performance improves with increasing model size and recurrent step. As shown in Figure~\ref{fig:scalinglaw_performance_vs_modelsize}, for all recurrent steps, the benchmark performance of both the \ut{} and Standard models increases as the model size grows, which aligns with the principle of LLMs: larger is better. As shown in Figure~\ref{fig:scalinglaw_performance_vs_recurrentstep}, except for the \ut{} at 778M and 1.364B, both the \ut{} and Standard models show that benchmark performance increases as the recurrent step increases. This indicates that latent reasoning is indeed useful for both the \ut{} and Standard Transformer.

2. Overall, the performance of the standard model exceeds that of \ut{} under the same conditions. This gap increases with the recurrent step and decreases with the model size. Observing Figure~\ref{fig:scalinglaw_performance_vs_modelsize} and Figure~\ref{fig:scalinglaw_performance_vs_recurrentstep}, it is clear that the benchmark performance of the Standard model is consistently higher than that of \ut{}, indicating that the Standard model has a scoring advantage without considering computational budget. Furthermore, we define the benchmark performance gap as the benchmark performance of the Standard model minus that of \ut{}, and this value is positive in all our experiments. As shown in Table~\ref{tab:scalinglaw_performancegap}, as the recurrent step increases, the benchmark performance gap also increases, suggesting that as the number of recurrences rises, the effect of models not sharing parameters gradually surpasses that of models sharing parameters. Besides, we find that the benchmark performance gap generally has a negative correlation with model size when the maximum recurrent step is relatively low, meaning that as the model size increases, the performance of \ut{} becomes closer to that of the Standard model, resulting in a smaller gap between the two. This trend is particularly consistent when the recurrent step is 4.

\begin{table}[htbp!]
    \centering
    \small
    \caption{The average benchmark performance gap between \ut{} and Standard models as the recurrent step varies at different model sizes. The gap is defined as (Standard model score - \ut{} score). As the recurrent step increases, the performance gap generally increases.}
    \label{tab:scalinglaw_performancegap}
    \begin{tabular}{@{}lrr@{}}
    \toprule
     & \multicolumn{2}{c}{\textbf{Average Performance Gap}} \\
    \cmidrule(lr){2-3}
    \textbf{Model Size} & \textbf{Step 2} & \textbf{Step 4} \\
    \midrule
    170M & 0.021 & 0.039 \\
    340M & 0.023 & 0.037 \\
    680M & 0.015 & 0.026 \\
    1.3B & 0.017 & 0.025 \\
    \bottomrule
    \end{tabular}
\end{table}

\subsection{RQ2: How do recurrent step impact the total loss and step-wise loss in the context of \ut{}?}
\label{scaling_law_sub_2}

In this subsection, we investigate the predictability and generalizability of \ut{} from the perspective of training loss, examining the impact of recurrent step on the trends in total loss and step-wise loss. The experiment is set up in complete consistency with Section~\ref{scaling_law_sub_1}, but we focus more on the total loss and step-wise loss during the training process. Here, step-wise loss refers to the loss of the same \ut{} at different recurrent step.

Here, we have following variables: model size $N$, training data size $D$, maximum recurrent step $T_m$, recurrent step $T$, total loss $L_t$, and step-wise loss $L_s$. Following Chinchilla~\citep{hoffmann2022training}, we first attempt to fit the relationship between $L_t$ and $N,D,T_m$ in the form of a power law:

$$
L_t = E + \frac{A}{(N+t_1)^\alpha} + \frac{B}{(D+t_2)^\beta} + \frac{C}{(T_m+t_3)^\gamma}
$$

\begin{figure}[htbp!]
    \centering
    \includegraphics[width=0.9\linewidth]{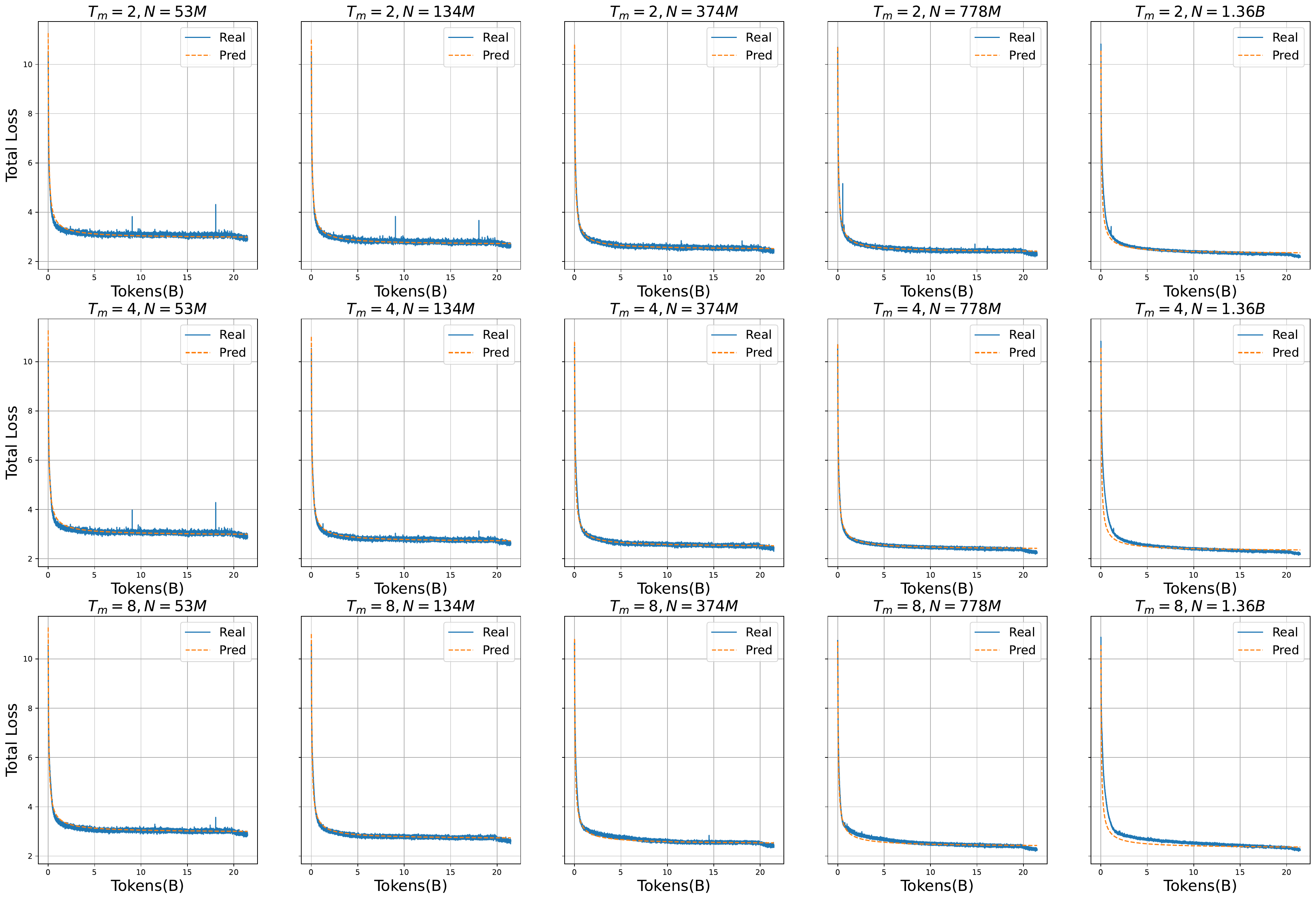}
    \caption{Illustration of the actual loss curve and the loss curve predicted by the scaling law. To demonstrate the predictability of \ut{}, we have used all data points for fitting, proving its predictability in terms of model size, training data size, and max recurrent step. The orange dashed line represents the prediction, while the blue solid line represents the actual loss.}
    \label{fig:scalinglaw_totalloss_scalinglaw_predictability}
\end{figure}

The purpose of $t_1$, $t_2$, and $t_3$ is to prevent the variables from exploding in value near zero, allowing the fitting curve to be smoother. We refer to above formula as the \textbf{Total Loss Scaling Law}. First, to validate the predictability of \ut{}, we fit all the data points, and the resulting curve is shown in Figure~\ref{fig:scalinglaw_totalloss_scalinglaw_predictability}. We find that the actual loss curve and the predicted loss curve are highly consistent, demonstrating the predictability of \ut{} in terms of model size, training data size, and max recurrent step. We quantify the consistency of the scaling law using the coefficient of determination $R^2$. For $R^2=1-\mathrm{SSE}/\mathrm{SST}$ on nonconstant targets, values closer to 1 indicate a better fit; $R^2=0$ matches the constant mean predictor, and a negative value indicates a worse fit than that baseline. The sign of $R^2$ does not indicate the direction of correlation. Fitting the Total Loss Scaling Law using all data points and calculating $R^2$ with all data points, we obtain an $R^2$ value of 0.9596. This confirms the strong correlation between total loss and model size, training data size, and max recurrent step, demonstrating the predictability of the Total Loss Scaling Law.

In addition to its predictability, we further explore the generalizability of the Total Loss Scaling Law. Predictability refers to the ability of the Scaling Law to fit all data points into a unified curve when all data points are available. Generalizability, on the other hand, indicates whether the Scaling Law can predict unseen data points when fitting is done with a subset of data points. For example, generalizability tests whether the performance of a 14B model can be predicted using the known performances of 1B and 7B models~\citep{que2024d}. To verify its generalizability across model size $N$, training data size $D$, and maximum recurrent step $T_m$, we have conducted related experiments, details can be found in Appendix~\ref{app:generalizabilty_exp_totalloss_scalinglaw}.

During the \ut{} training process, we compute the cross-entropy loss at each recurrent step, which we refer to as step-wise loss $L_s$. We aim to explore the relationship between step-wise loss $L_s$ and the current recurrent step $T$, model size $N$, and training data size $D$. Similarly, we can fit the scaling law between $L_s$ and $N,D,T$, with the formula as follows:

$$
L_s = E + \frac{A}{(N+t_1)^\alpha} + \frac{B}{(D+t_2)^\beta} + \frac{C}{(T+t_3)^\gamma}
$$

\begin{figure}[htbp!]
    \centering
    \includegraphics[width=0.9\linewidth]{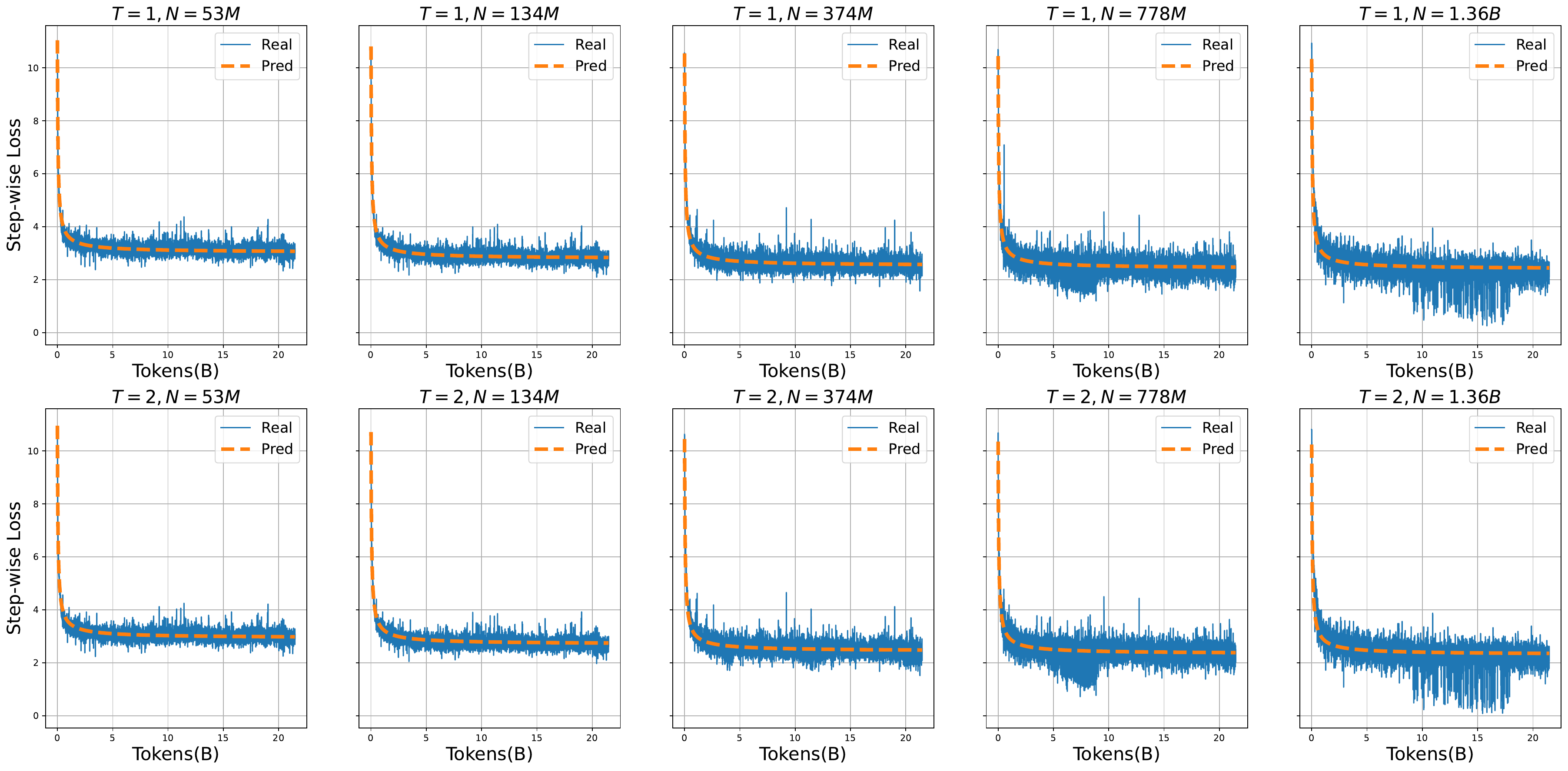}
    \caption{Illustration of the actual loss curve and the loss curve predicted by the Step-wise Loss Scaling Law when the maximum recurrent step is equal to 2.}
    \label{fig:scalinglaw_stepwiseloss_scalinglaw_predictability_maxrecurrentstep_2}
\end{figure}

\begin{figure}[htbp!]
    \centering
    \includegraphics[width=0.9\linewidth]{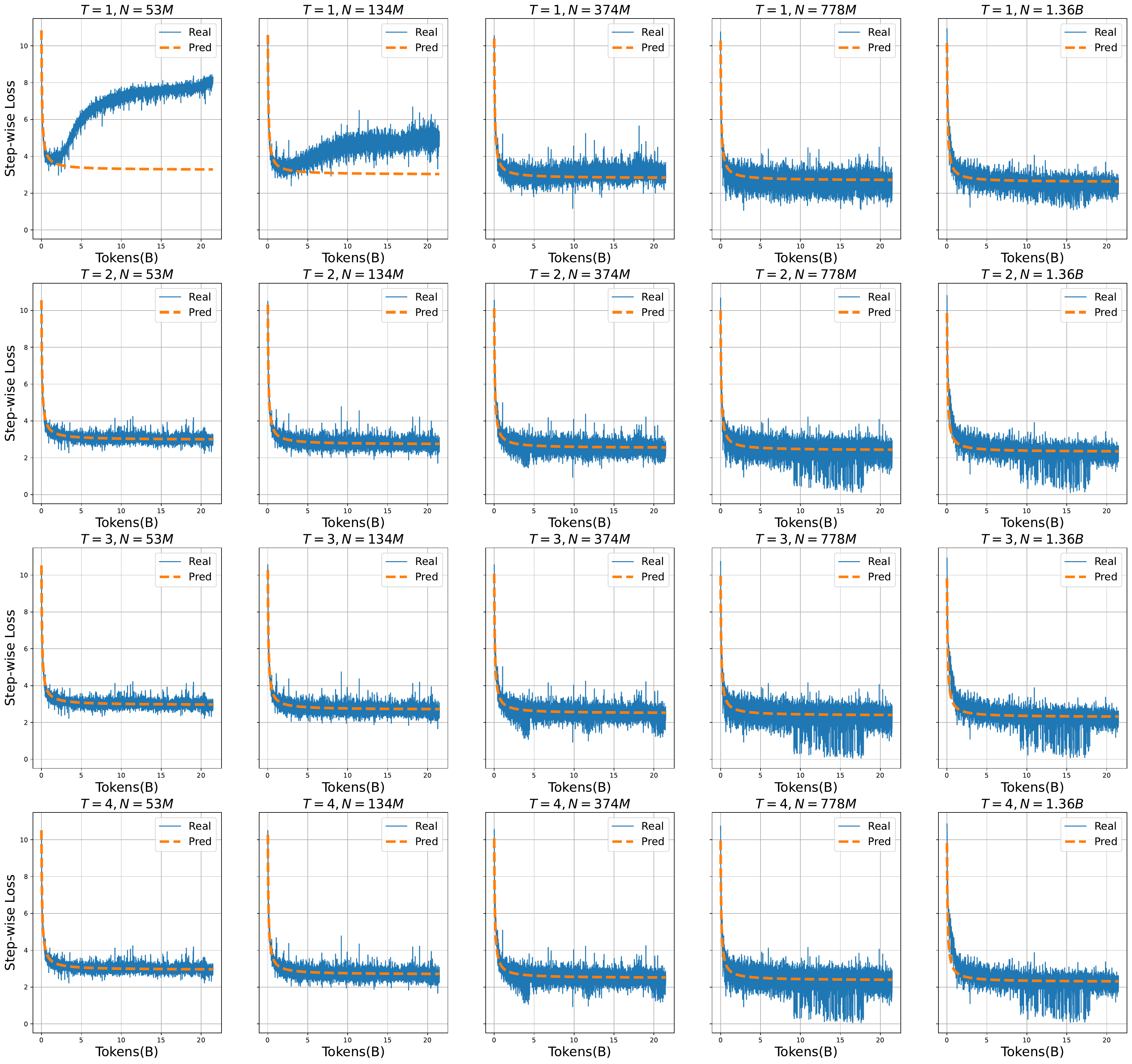}
    \caption{Illustration of the actual loss curve and the loss curve predicted by the Step-wise Loss Scaling Law when the maximum recurrent step is equal to 4.}
    \label{fig:scalinglaw_stepwiseloss_scalinglaw_predictability_maxrecurrentstep_4}
\end{figure}

\begin{figure}[htbp!]
    \centering
    \includegraphics[width=0.7\linewidth]{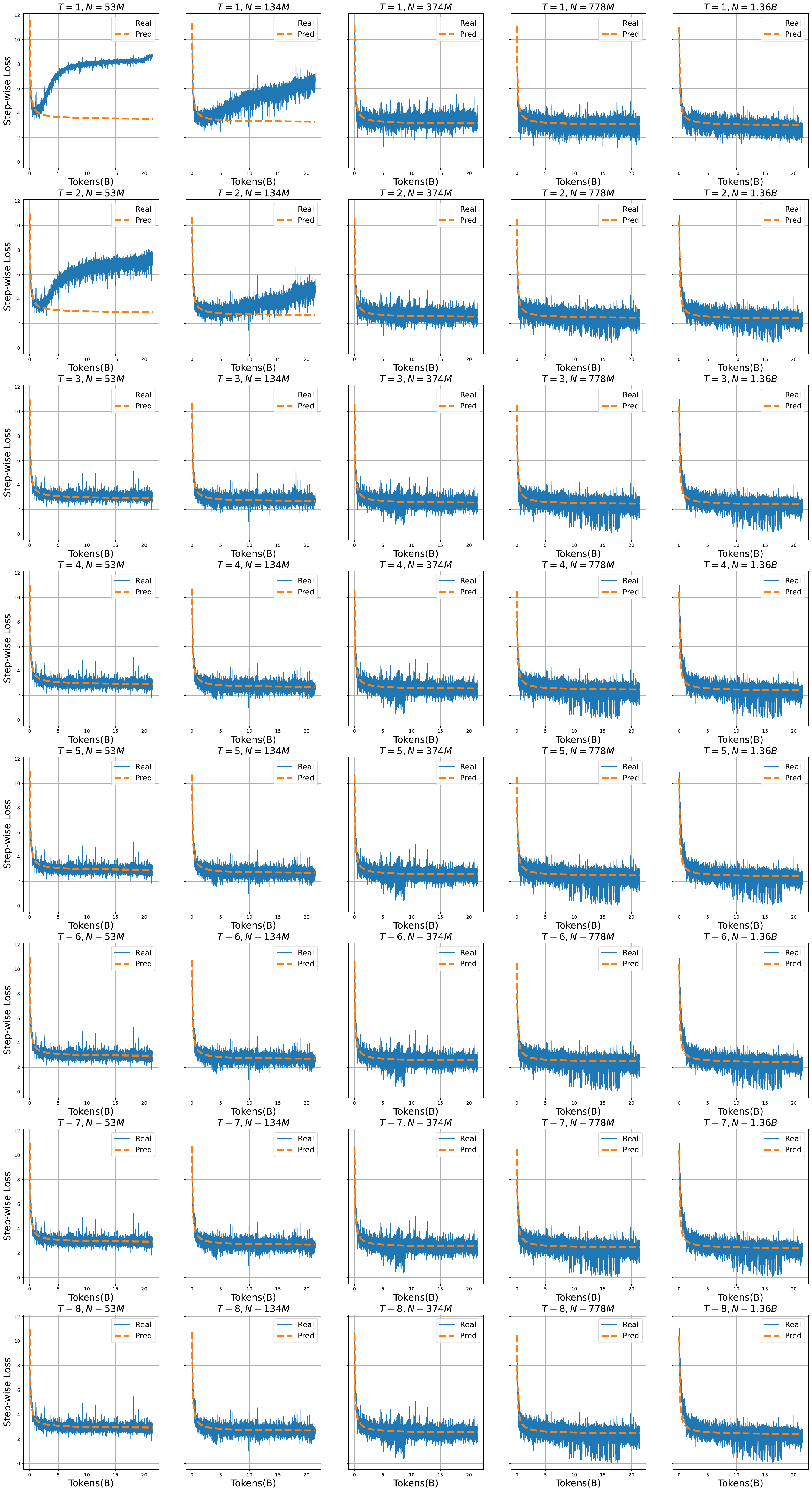}
    \caption{Illustration of the actual loss curve and the loss curve predicted by the Step-wise Loss Scaling Law when the maximum recurrent step is equal to 8.}
    \label{fig:scalinglaw_stepwiseloss_scalinglaw_predictability_maxrecurrentstep_8}
\end{figure}

We refer to the above formula as the \textbf{Step-wise Loss Scaling Law}. We also present the fitting effectiveness from the perspectives of predictability and generalizability. Regarding predictability, we fit all data points. 
Even with the same recurrent step, the loss curve can vary significantly across different maximum recurrent steps. To ensure the independence of the variable recurrent step $T$, we do not consider the maximum recurrent step in the Step-wise Loss Scaling Law formula and focus solely on the relationship between $L_s$ and $N,D,T$. 
Therefore, we have a total of three major experiments, each representing the fitting of the Step-wise Loss Scaling Law for maximum recurrent steps of 2, 4, and 8. The fitting results of the Step-wise Loss Scaling Law are shown in Figure~\ref{fig:scalinglaw_stepwiseloss_scalinglaw_predictability_maxrecurrentstep_2}, Figure~\ref{fig:scalinglaw_stepwiseloss_scalinglaw_predictability_maxrecurrentstep_4}, and Figure~\ref{fig:scalinglaw_stepwiseloss_scalinglaw_predictability_maxrecurrentstep_8}, which illustrate the trends of the actual and fitted curves for maximum recurrent steps of 2, 4, and 8, respectively. 
We find that in some cases in Figure~\ref{fig:scalinglaw_stepwiseloss_scalinglaw_predictability_maxrecurrentstep_4} and Figure~\ref{fig:scalinglaw_stepwiseloss_scalinglaw_predictability_maxrecurrentstep_8}, $L_s$ increases with the increase in $D$. We consider this a special case and will discuss it in detail in Section~\ref{sec:scalinglaw_rq3}; we will ignore these outlier data points during fitting. The $R^2$ for the three max recurrent steps are 0.8898, 0.8146, and 0.795, respectively. As the maximum recurrent step increases, the increase in the number of data points leads to lower $R^2$ values. The step-wise loss itself is less stable than the total loss, resulting in greater variability. Thus, the obtained $R^2$ values are not as high as those of the Total Loss Scaling Law. However, it is still evident that the scaling law is able to capture the overall trend of the curves, demonstrating the predictability of the Step-wise Loss Scaling Law.
The fitting parameter $\gamma$ of the Step-wise Loss Scaling Law is positive, indicating that $L_s$ decreases as the recurrent step increases. This aligns with our original intent in the design of the recurrence. Besides, we present the generalizability of the Step-wise Loss Scaling Law in Appendix~\ref{app:generalizabilty_exp_stepwiseloss_scalinglaw}.

In summary, both total loss and step-wise loss exhibit a strong correlation with $N,D,T/T_m$. The fitting results demonstrate the predictability and generalizability of the Scaling Law for \ut{}. In next section, we will explore the relationship between total loss and step-wise loss in greater depth.

\subsection{RQ3: What is the inherent connection between total loss and step-wise loss?}
\label{sec:scalinglaw_rq3}

We first review the training objectives of \ut{}:

$$
L_t = \sum_{T=1}^{T_{m}} q_\phi(z=T \mid x)\, L_{s}^{(T)} - \beta \cdot H(q_\phi(z \mid x))
$$

$L_s^{(T)}$ represents the step-wise loss at the recurrent step $T$. By analyzing the above form, we can see that the total loss consists of two components. The first part is the expected task loss, which is a weighted sum of the step-wise loss. The second part is entropy regularization, whose primary purpose is to ensure that the learned gating mechanism $q_\phi$ does not converge to a specific recurrent step.

In our extensive small-scale experiments, we have observed an interesting phenomenon: the exploitation of total loss for shallow step-wise loss. To be specific, as shown in Figure~\ref{fig:scalinglaw_stepwiseloss_scalinglaw_predictability_maxrecurrentstep_4} and Figure~\ref{fig:scalinglaw_stepwiseloss_scalinglaw_predictability_maxrecurrentstep_8}, when the model size is insufficient, the shallow step-wise loss increases with the growing amount of training data. This is an unusual phenomenon, typically, all step-wise losses should decrease as the amount of training data increases. We attempt to explain this phenomenon. In Section~\ref{scaling_law_sub_2}, it is mentioned that the step-wise loss decreases with an increasing recurrent step, indicating that deeper recurrent steps result in lower $L_s$. To minimize the expected task loss, the learned gating mechanism assigns more weight to deeper recurrent steps. However, entropy regularization ensures that the learned gating mechanism does not deviate too much from the prior distribution. When the model size is insufficient, the amount of information it can encode is limited. To further reduce the total loss, this results in an increase in shallow step-wise loss, which in turn allows the weights to favor higher recurrent steps to lower the total loss. 
Thus, to ensure that the trend of step-wise loss remains normal, a larger model size may be more effective for \ut{}.

As mentioned in Section~\ref{scaling_law_sub_2}, the scaling law for \ut{} is predictable and generalizable for both total loss and step-wise loss. We have:


$$
L_t = E_t + \frac{A_t}{(N+t_{1t})^\alpha} + \frac{B_t}{(D+t_{2t})^\beta} + \frac{C_t}{(T_m+t_{3t})^\gamma}
$$
$$
L_s^{(T)} = E_s + \frac{A_s}{(N+t_{1s})^\alpha} + \frac{B_s}{(D+t_{2s})^\beta} + \frac{C_s}{(T+t_{3s})^\gamma}
$$

The subscripts $s$ and $t$ represent the fitting parameters for the Step-wise Loss Scaling Law and the Total Loss Scaling Law, respectively. By substituting the Step-wise Loss Scaling Law into the training objectives, we have:

$$
L_t = \sum_{T=1}^{T_{m}} q_\phi(z=T \mid x)\, \left(E_s + \frac{A_s}{(N+t_{1s})^\alpha} + \frac{B_s}{(D+t_{2s})^\beta} + \frac{C_s}{(T+t_{3s})^\gamma}\right) - \beta \cdot H(q_\phi(z \mid x))
$$

For the first three terms in $L_s^{(T)}$, the sum of $q_\phi$ equals 1, allowing us to factor it out, which gives us:

$$
L_t = E_s + \frac{A_s}{(N+t_{1s})^\alpha} + \frac{B_s}{(D+t_{2s})^\beta} + \sum_{T=1}^{T_{m}} q_\phi(z=T \mid x)\, \frac{C_s}{(T+t_{3s})^\gamma} - \beta \cdot H(q_\phi(z \mid x))
$$

As the amount of training data increases, the learned gating mechanism $q_\phi$ stabilizes, and we observe that the value of the entropy regularization term becomes relatively low, accounting for approximately 1\% to 5\% of the total loss. If the form of $q_\phi$ gradually stabilizes, we treat it as a constant term, and the formula becomes:

$$
L_t = E_s + \frac{A_s}{(N+t_{1s})^\alpha} + \frac{B_s}{(D+t_{2s})^\beta} + E_{other}
$$

In Section~\ref{scaling_law_sub_2}, when considering the Step-wise Loss Scaling Law, we will fix the maximum recurrent step. Once we determine the model's maximum recurrent step $T_m$, the forms of the above formula and the Total Loss Scaling Law are completely consistent, indicating that there is a trend consistency in the scaling law between total loss and step-wise loss. 

\begin{figure}
    \centering
    \includegraphics[width=\linewidth]{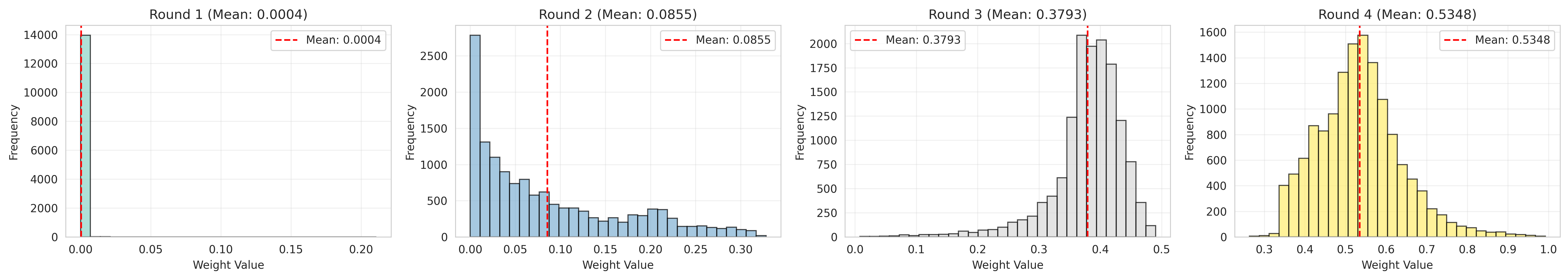}
    \caption{Distribution of the learned ponder weights ($q_\phi(z=t \mid x)$) for each recurrent step $t$ when the maximum recurrent step $T_m=4$. These weights were collected during inference on the MMLU benchmark.}
    \label{fig:round_wise_his}
\end{figure}
\begin{figure}[htbp!]
    \centering
    \includegraphics[width=0.9\linewidth]{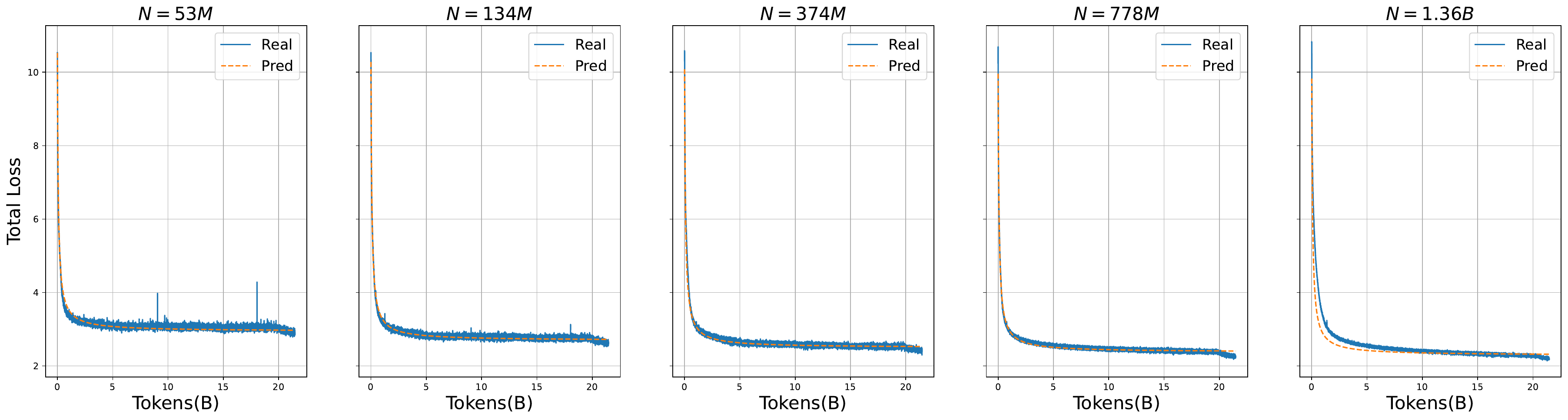}
    \caption{Illustration of the actual loss curve and the loss curve predicted by the estimated Scaling Law when the maximum recurrent step is equal to 4.}
    \label{fig:scalinglaw_estimate_scalinglaw}
\end{figure}

We further demonstrate this through practical experiments. We take the situation where the max recurrent step is equal to 4. First, we perform a standard fitting of the Step-wise Loss Scaling Law to obtain the fitting parameters $E_s, A_s, B_s, C_s,$ and so on. Next, we observe and record the distribution of $q_\phi$ for each recurrent step $t$ when the maximum recurrent step $T_m=4$, as shown in Figure~\ref{fig:round_wise_his}. For convenience, we take the average value of $q_\phi$ at different recurrent steps and treat it as a normal discrete distribution, resulting in the distribution \{0.0004, 0.0855, 0.3793, 0.5348\}. We then substitute this distribution and the $T$ values into the training objective, ignoring the entropy regularization term (after the training stabilizes, it becomes relatively low, for simplicity, we will just ignore it). This leads to a fitting formula, and upon substituting the actual fitting data points $N$ and $D$, the computed $R^2$ value is 0.961, with the fitting results illustrated in Figure~\ref{fig:scalinglaw_estimate_scalinglaw}. We can see that the fitting accuracy is high, and the predicted curve closely matches the actual curve, indicating that, under a relatively rough estimate, step-wise loss can be transformed into total loss, thus indirectly suggesting an intrinsic connection between the two.

\section{Details of the Scaling Law for \ut{}}
\subsection{Generalizability for the Total Loss Scaling Law}
\label{app:generalizabilty_exp_totalloss_scalinglaw}

To demonstrate the generalizability of the Total Loss Scaling Law across model size, training data, and maximum recurrent step, we have conducted relevant experiments. Our evaluation metric is the coefficient of determination $R^2$. To evaluate the fitting effectiveness of the Scaling Law, we calculate the coefficient of determination of all data points.

\begin{figure}[htbp!]
    \centering
    \includegraphics[width=0.9\linewidth]{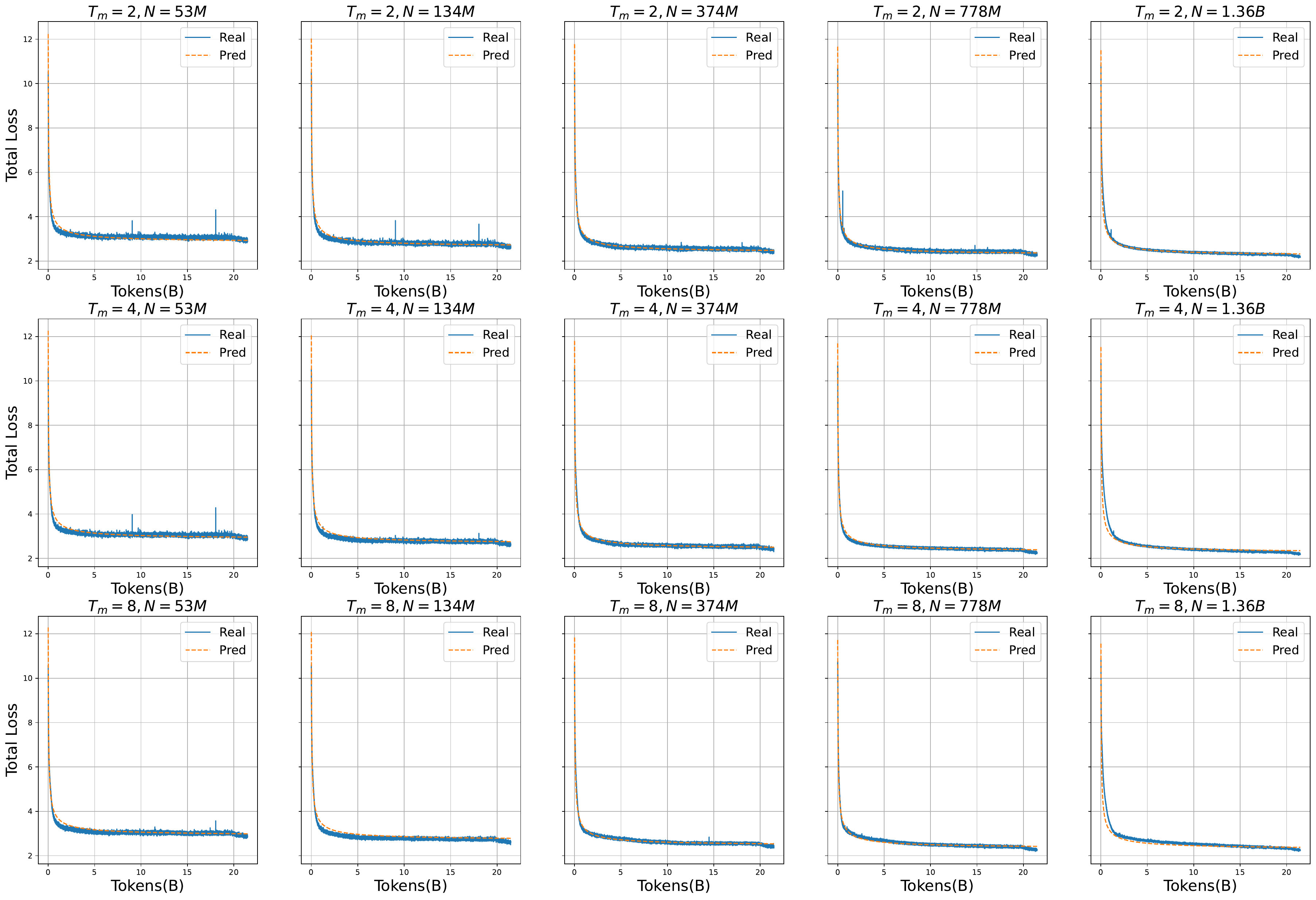}
    \caption{Illustration of model size generalizability for the Total Loss Scaling Law. The fitting data includes model sizes of 374M, 778M, and 1.364B. The predicted curves for the unseen model sizes of 53M and 134M closely align with the actual curves, demonstrating the generalizability of the Total Loss Scaling Law with respect to model size.}
    \label{fig:totalloss_scalinglaw_generalizability_modelsize}
\end{figure}

\textbf{Model Size Generalizability} For model size generalizability, our total data points include five different model sizes: 53M, 134M, 374M, 778M, and 1.364B. We select three model sizes as fitting data points, resulting in $\binom{5}{3}=10$ possible combinations. After fitting, the average $R^2$ across the 10 combinations is 0.9542, which is similar to the result obtained with the full data points, demonstrating the model size generalizability of the Total Loss Scaling Law. Figure~\ref{fig:totalloss_scalinglaw_generalizability_modelsize} illustrates an example.

\textbf{Training Data Generalizability} Regarding the training data size, we are primarily interested in whether the Scaling Law can predict model performance as training data increases. Therefore, we typically use the preceding data points to predict future data points. To align with this starting point, we have conducted three sets of experiments, using the current 25\%, 50\%, and 75\% of the data points as fitting data to predict the overall fitting performance.The $R^2$ values for using the first 25\%, 50\%, and 75\% of the data as fitting points are 0.9385, 0.9609, and 0.962, respectively. It is evident that as the number of data points increases, the consistency between the fitted curves and the actual curves improves. In other words, if you want to predict model performance at larger training sizes, collecting data points closer to those of larger model sizes will yield better prediction results.

\textbf{Max Recurrent Step Generalizability} We have conducted a total of three different maximum recurrent steps: 2, 4, and 8. To verify the generalizability with respect to maximum recurrent step, we select two of these as fitting data points and perform the fitting, followed by validation on the full data points and calculation of $R^2$. The average $R^2$ for the three sets of experiments is 0.9581, demonstrating the generalizability of the Total Loss Scaling Law with respect to maximum recurrent step.

\subsection{Generalizability for the Step-wise Loss Scaling Law}
\label{app:generalizabilty_exp_stepwiseloss_scalinglaw}

Following the same approach as in Section~\ref{app:generalizabilty_exp_totalloss_scalinglaw}, we seek to explore the performance of the Scaling Law on unseen data points, specifically regarding the generalizability of the Scaling Law. In this subsection, we explore the generalizability of the Step-wise Loss Scaling Law from three aspects: model size generalizability, training data generalizability, and recurrent step generalizability. The evaluation metric remains the coefficient of determination $R^2$. To evaluate performance on unseen data points, we will calculate the coefficient of determination using all data points, while fitting will only use a subset of the data points.

\begin{figure}[htbp!]
    \centering
    \includegraphics[width=0.9\linewidth]{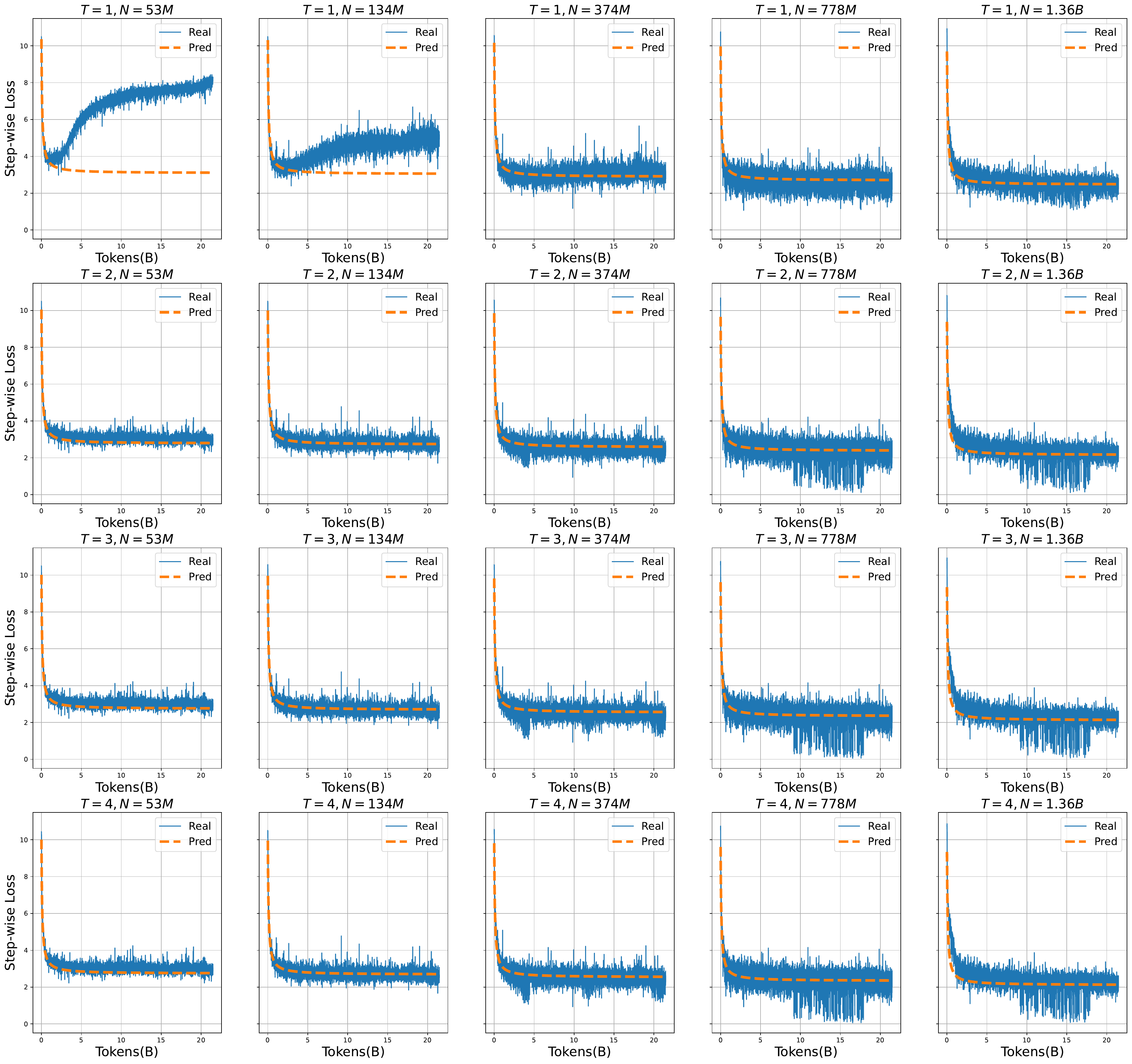}
    \caption{Illustration of model size generalizability for the Step-wise Loss Scaling Law. The fitting data comprises three medium model sizes: 134M, 374M, and 778M. To verify the fitting consistency of the model on unseen larger model size 1.364B and unseen smaller model size 53M, we can observe that the predicted curves reflect the trends of the actual data points, demonstrating the generalizability of the Step-wise Loss Scaling Law with respect to the model size.}
    \label{fig:stepwiseloss_scalinglaw_generalizability_modelsize}
\end{figure}

\textbf{Model Size Generalizability} 
The Scaling Law experiments include five different model sizes: 53M, 134M, 374M, 778M, and 1.364B. To verify the generalizability of model size, we select three of these as fitting data points. In each fitting experiment, the Scaling Law does not have access to the remaining two model size data points during fitting, ensuring the reasonableness and validity of the results through repeated experiments. 
Specifically, to save on resources, we have conducted experiments only for max recurrent step of 2 and 4, resulting in a total of $\binom{5}{3} \times 2 = 20$ small experiments.
The final experimental results show that for max recurrent step of 2, the average $R^2$ value is 0.8815, while for max recurrent step of 4, the average $R^2$ value is 0.797.
This difference is not significant compared to the $R^2$ values obtained from the full data points (0.8898 and 0.8146), demonstrating the generalizability of the Step-wise Loss Scaling Law with respect to model size. To illustrate the results more clearly, we show example fitting curves in Figure~\ref{fig:stepwiseloss_scalinglaw_generalizability_modelsize}. It is important to note that using only a subset of data points for fitting may lead to miscalculations of the Scaling Law on unseen data points. Due to the nature of the power law, if the values are too small, it may result in a very large computed value, causing inaccuracies. To ensure the validity of the fitting, we can attempt to adjust the initial fitting values or impose some constraints on the fitting algorithm. For convenience, we adjust the initial fitting values to make the fitting formula effective over a broader range of data points.

\textbf{Training Data Generalizability} 
Following Section~\ref{app:generalizabilty_exp_totalloss_scalinglaw}, to ensure the validity of the fitting, we have selected the first 25\%, 50\%, and 75\% of the data points for fitting. In the case of max recurrent step of 2, the $R^2$ values are 0.8686, 0.8882, and 0.8896, respectively. For max recurrent step of 4, the $R^2$ values are 0.793, 0.813, and 0.8142. 
It can be observed that as the number of fitting data points increases, the fitting accuracy improves. This aligns with the intuition that fitting with more data points generally yields better results. 
Additionally, these results are similar to those obtained from fitting with the full data points (0.8898 and 0.8146), demonstrating the generalizability of the Step-wise Loss Scaling Law with respect to training data.

\begin{figure}[htbp!]
    \centering
    \includegraphics[width=0.9\linewidth]{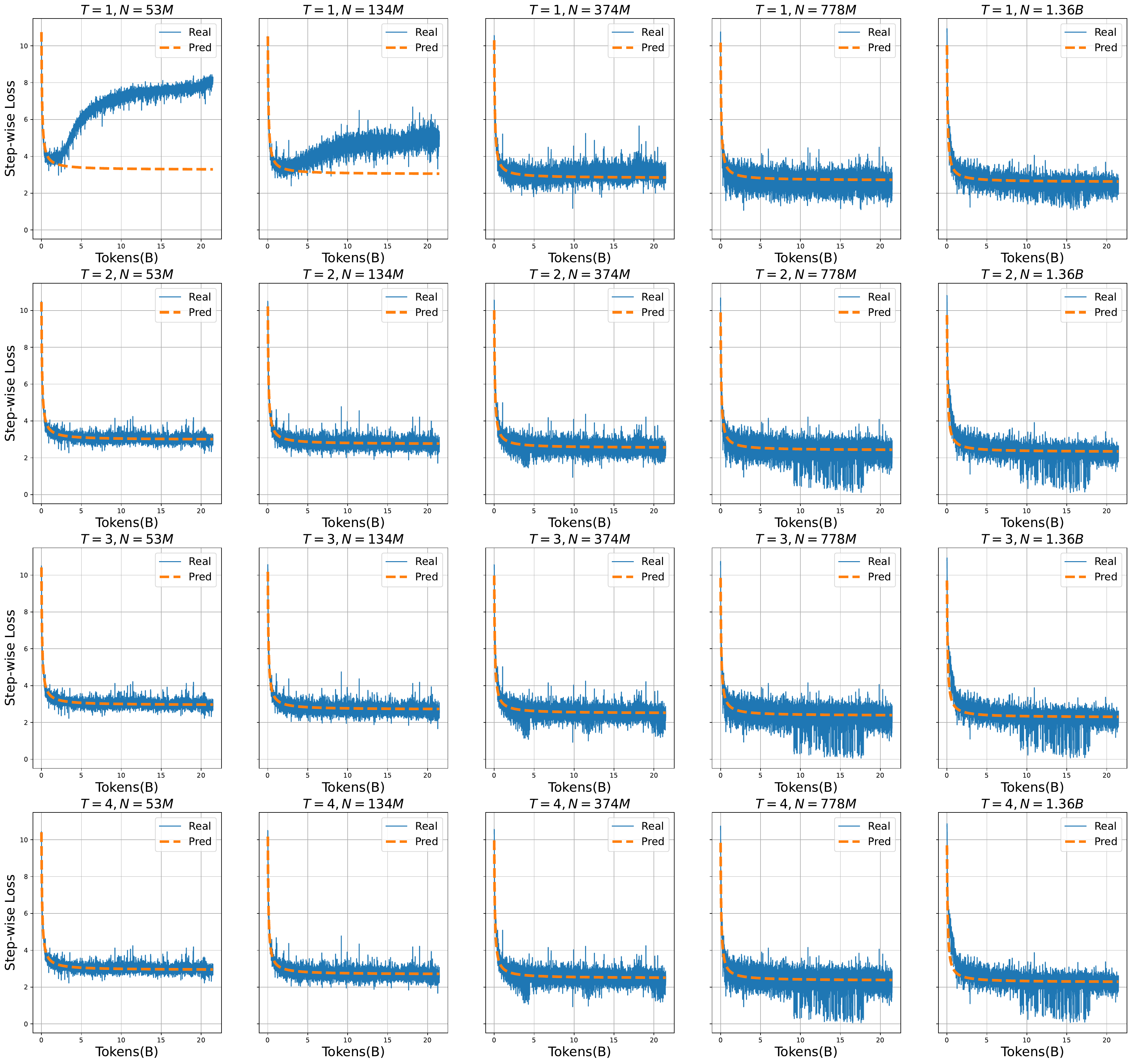}
    \caption{Illustration of recurrent step generalizability for the Step-wise Loss Scaling Law. The fitting data includes three different recurrent steps: recurrent step = 1, 2, and 3. At the unseen data points of recurrent step = 4, the predicted curve closely matches the actual curve, demonstrating the generalizability of the Step-wise Loss Scaling Law with respect to recurrent step.}
    \label{fig:stepwiseloss_scalinglaw_generalizability_recurrentstep}
\end{figure}

\textbf{Recurrent Step Generalizability} In the case of max recurrent step equal to 2, there are only two recurrent step values, making it unreasonable to conduct generalizability experiments. Therefore, we choose to perform experiments with max recurrent step equal to 4. In this situation, we have four different recurrent step values: 1, 2, 3, and 4. We randomly select three of these as fitting data points, resulting in a total of $\binom{4}{3}=4$ experiments. The average $R^2$ value obtained from these four experiments is 0.8118, which is similar to the $R^2$ value of 0.8146 obtained from the full data points, demonstrating the generalizability of the Step-wise Loss Scaling Law with respect to recurrent step. Figure~\ref{fig:stepwiseloss_scalinglaw_generalizability_recurrentstep} presents a specific example, showing a high degree of consistency between the fitted curve and the actual curve.

\end{document}